\documentclass[11pt,a4]{article}

\RequirePackage{xr-hyper}[6.00]

\RequirePackage{hyperref}[6.83]
\hypersetup{
  frenchlinks=false,
  pdfborder={0 0 0},
  naturalnames=false,
  hypertexnames=false,
  breaklinks,
  colorlinks=true,
    linkcolor=black,
    citecolor=black,
    filecolor=black,
    urlcolor=black,
}

\usepackage{amssymb, amsthm, amsmath, amsbsy, latexsym,bbm,url}

\usepackage{mathrsfs} 
\usepackage{multirow}
\usepackage[affil-it]{authblk}
\usepackage{placeins}
\usepackage{fullpage}
\usepackage{tabularx}
\usepackage[all]{hypcap}
\usepackage{pifont}
\usepackage{bigints}
\usepackage[caption=false,font=footnotesize]{subfig}
\usepackage{symfdpreamble}
\usepackage{cleveref}
\usepackage{xr-hyper}

\usepackage{booktabs}
\usepackage{threeparttable}

\usepackage[toc,page]{appendix}

\usepackage{tikz}
\usetikzlibrary{spy,positioning,calc}
\usepackage{pgfplots}
\pgfplotsset{compat=1.11}

\newcommand{\rev}[1]{#1}

\ifpdf
\hypersetup{
  pdftitle={Edge, Ridge, and Blob Detection with Symmetric Molecules},
  pdfauthor={R. Reisenhofer, and Emily J. King}
}
\fi

\begin{document}
	\date{}
	\title{Edge, Ridge, and Blob Detection with Symmetric Molecules}
	\author[1]{Rafael Reisenhofer\thanks{rafael.reisenhofer@univie.ac.at}}
	\author[2]{Emily J. King}
	\affil[1]{University of Vienna, Faculty of Mathematics}
	\affil[2]{University of Bremen, AG Computational Data Analysis}	
  \maketitle
\begin{abstract}
We present a novel approach to the detection and characterization of edges, ridges, and blobs in two-dimensional images which exploits the symmetry properties of directionally sensitive analyzing functions in multiscale systems that are constructed in the framework of $\alpha$-molecules. The proposed feature detectors are inspired by the notion of phase congruency, stable in the presence of noise, and by definition invariant to changes in contrast. We also show how the behavior of coefficients corresponding to differently scaled and oriented analyzing functions can be used to obtain a comprehensive characterization of the geometry of features in terms of local tangent directions, widths, and heights. The accuracy and robustness of the proposed measures are validated and compared to various state-of-the-art algorithms in extensive numerical experiments in which we consider sets of clean and distorted synthetic images that are associated with reliable ground truths. To further demonstrate the applicability, we show how the proposed ridge measure can be used to detect and characterize blood vessels in digital retinal images and how the proposed blob measure can be applied to automatically count the number of cell colonies in a Petri dish.
\end{abstract}
\section{Introduction}
The correct localization of the significant structures in an image as well as the precise characterization of their geometry are two \rev{eminent} tasks in digital image processing with an overwhelming number of applications. Since the advent of digital image processing, a large body of research has been devoted to the development and analysis of algorithms for the detection and characterization of features such as edges, ridges, and well-defined two-dimensional shapes. Surprisingly, in many practical situations, even highly optimized implementations of popular methods such as the Canny edge detector \cite{Can1986}, approaches that are based on directionally sensitive filters \cite{jacob2004design}, or even multiscale systems of anisotropic functions \cite{YLEK2009} are not always capable of reliably identifying the features in question. This is often the case for images which are heavily distorted by noise, in which different features are strongly overlapping, or where the geometry of the considered features is characterized by a high variation and irregularities such as corner points. It is also worth noting that feature detection in general is a task where, despite recent advances, computers are still often outperformed by humans.

A main difficulty researchers face when developing methods for feature detection is to identify universal and computationally tractable properties that are characteristic of or even unique to points at which a certain feature is localized. In the case of edge detection, the first and most basic observation was that edges are typically associated with changes in contrast and can thus be identified by considering points with large image gradients. This approach led to the development of early edge enhancement filters such as the Roberts filter \cite{roberts1963machine}, the Prewitt filter \cite{prewitt1970object}, and the Sobel filter \cite{SF68}, which are still widely applied today. However, any procedure merely considering the magnitude of gradient filters as an indicator is highly sensitive to noise and variations in image illumination. Two famous and influential descriptors which already made a big step towards capturing the structural nature of edges were proposed in the 1980's by Marr and Hildreth \cite{marr1980theory} and Canny \cite{Can1986}. Marr and Hildreth observed that edges coincide with zero-crossings in the second directional derivative of the image intensities while Canny developed an algorithm for edge detection which identifies points at which the image gradient reaches a local maximum. The Canny edge detector is also based on a gradient filter, and the Marr-Hildreth-operator applies a discrete Laplace filter. However, both methods are not just evaluating whether a magnitude exceeds a certain threshold but consider local structural properties by testing for zero-crossings and local maxima. 

A dimensionless approach to identifying features such as edges and ridges may be found in the so-called local energy model, which was proposed in the late 1980's and postulates that the location of features coincides with points in an image for which the Fourier components are maximally ``in phase'' \cite{MoOw1987,MRBO1986,VeOw1990}. \rev{Dimensionless here means that the values yielded by the measure are {not} describing a physical quantity such as image intensities.} To precisely calculate the degree of phase congruency, it is necessary to optimize a certain value over inputs from a continuous interval for every point in an image. This computationally highly inefficient procedure was later significantly simplified by Kovesi, who showed that an equivalent measure can be formulated in terms of convolutions with differently scaled and oriented complex-valued wavelet filters \cite{Kov1999,Kov2000}. The resulting phase congruency measure yields a contrast-invariant feature detector that simultaneously detects edges and ridges and can be efficiently implemented to process digital images. It was one of the inspirations for the measures for edge, ridge, and blob detection presented here and will be explicitly defined later.%It has furthermore strongly inspired the measures for edge, ridge, and blob detection presented here and will therefore be introduced in greater detail later in this work.

In recent years, multiscale systems of analyzing functions developed in the field of applied harmonic analysis have become increasingly successful in efficiently representing, approximating, and analyzing two-dimensional digital images. In particular, systems that are based on anisotropic scaling such as curvelets \cite{candes2000curvelets,candes2004new}, shearlets \cite{labate2005sparse,kutyniok2012shearlets}, or contourlets \cite{DoV03} yield provably in-a-sense optimal error decay rates when approximating certain classes of digital images that are governed by curvilinear and anisotropic structures. These constructions have been considered in a wide range of applications in digital image processing including edge and feature detection \cite{YLEK2009,kutyniok2017classificationofedges,duval2017scale,KaLa16}.  Grohs et al.\ recently introduced the so-called $\alpha$-molecules framework, which allows for a unified treatment of the analysis of most multiscale systems developed in the field of applied harmonic analysis \cite{grohs2016alpha}. In the present work, we derive measures for the detection of edges, ridges, and blobs that are inspired by 
%the local energy model and 
the notion of phase congruency but realized by making full use of the flexibility of modern construction principles for anisotropic analyzing functions provided within the framework of $\alpha$-molecules.

Saying that the Fourier components are in phase at the location of an edge basically means that locally the considered function is purely defined by odd-symmetric sine components while all even-symmetric cosine components vanish. In the first step, we will derive a novel edge measure that utilizes this observation by testing for the local symmetry properties of a function by considering differently scaled and oriented odd- and even-symmetric analyzing functions constructed in the $\alpha$-molecule framework. This measure can then be generalized to detect ridges by interchanging the roles of the odd- and even-symmetric molecules and taking into account the width of the feature. We further show how, with a few modifications regarding the measure and the construction of analyzing functions, similar principles can be applied to the case of blob detection. Each analyzing function in a multiscale system considered in this work is associated with a specific pair of scaling and orientation parameters and thus conveys not only information about the location of features but also about their geometry in terms of feature width and orientation. Finally, we demonstrate how this information can be utilized to obtain measures for local tangent directions, widths, and heights, that yield detailed information about the geometry of an image.
\subsection{Contributions}
Earlier variants of the edge measure derived in Section~\ref{sec:edgemeasure} and the  ridge measure derived in Section~\ref{sec:ridgemeasure} that are based on complex-valued shearlets have already been published previously by the authors of the present work and their co-authors \cite{KRKLLH,Rei14,reisenhofer2016shearlet}. The main contribution of this work is to provide a comprehensive derivation and description of the proposed measures and bring them into the framework of $\alpha$-molecules which provides enough flexibility to allow for a unified treatment of the cases of edge, ridge, and blob detection. In particular, we demonstrate in Theorem~\ref{thm:smorder} that by considering tensor products of derivatives of the one-dimensional Gaussian and their Hilbert transforms, one can obtain a large class of well-behaved $\alpha$-molecules that also provide the necessary symmetry properties. We further show in Section~\ref{sec:blobmeasure} that local symmetry properties may be used in a related manner to yield a blob measure. We specifically derive functions that estimate the local tangent orientations of edge contours and ridge centerlines, local diameters of ridges and blobs, and the local contrast of a feature; these are summarized in Section~\ref{sec:orientationmeasure}.

To evaluate the strengths and weaknesses of the proposed measures in comparison with other state-of-the-art methods for feature detection, we performed extensive numerical experiments with respect to synthetic images with reliable ground truths, which may be found in Section~\ref{sec:syntheval}. This section also contains a figure of merit to measure success of edge detection that is less forgiving of false positives than Pratt's figure of merit ($\PFOM$) \cite{AbPr79}. To further demonstrate the applicability of the proposed framework, we consider the application of the ridge measure in the context of retinal image analysis in Section~\ref{sec:retinal} and show how the proposed blob measure can be applied for automatically counting the number of cell colonies in a Petri dish in Section~\ref{sec:cellcol}.

We have also developed a  \matlab{} toolbox that implements all of the proposed measures for two-dimensional digital images. The implementation strategies are briefly explained in Section~\ref{sec:digitalimplementation} of the Supplemental Materials, and the toolbox can be downloaded from \url{http://www.math.uni-bremen.de/cda/software.html}. 
\section{Related Work}
\label{sec:relatedwork}
We begin by giving a short review of the most important tools and concepts that will later be used to derive the respective edge, ridge, and blob measures.
\subsection{\texorpdfstring{$\alpha$}{Alpha}-Molecules}
\label{sec:alphamolecules}
Active research in the fields of time-frequency analysis and applied harmonic analysis led to the development of a wide variety of function systems and techniques that yield efficient multiscale representations for different classes of functions and signals. Building on results and ideas from classical Fourier analysis and the work of pioneers like Alfred Haar \cite{Haar10}, prominent examples include wavelets \cite{Dau92} but also systems based on anisotropic scaling such as ridgelets \cite{candes1998ridgelets,candes1999ridgelets}, curvelets \cite{candes2000curvelets,candes2004new}, shearlets \cite{labate2005sparse,kutyniok2012shearlets}, and contourlets \cite{DoV03}, which \rev{were developed} to address weaknesses of the traditional wavelet transform when dealing with anisotropic features such as edges or ridges that can appear in two- or higher-dimensional signals.

In spite of various differences in their construction and their approximation behavior for different classes of signals, many of the aforementioned function systems are constructed by applying translation, scaling, and rotation or shear operators to 
%a possibly infinite number of so-called 
generator functions, which typically need to satisfy a certain admissibility condition. In \cite{grohs2016alpha}, so-called $\alpha$-molecules were introduced to provide a unifying framework that includes different constructions of multiscale representations in order to enable the transfer of results from one theory to another, to allow for a categorization of multiscale representations with respect to the approximation behavior and also to facilitate the construction of novel systems. %The partitions of the Fourier domain induced by three different multiscale representation systems within the $\alpha$-molecule framework are schematically depicted in Figure~\ref{fig:freqpartitions}.

Systems of $\alpha$-molecules are constructed by translating, scaling, and rotating members of a possibly infinite set of generator functions, where the latter two operations are performed by applying scaling matrices
\begin{equation}
\ScM{s}{\alpha} = \begin{pmatrix}s & 0 \\0 & s^\alpha\end{pmatrix}, \quad s > 0,\; \alpha \in [0,1],
\label{eq:scalingmatrix}
\end{equation}
and rotation matrices
\begin{equation}
\label{eq:rotationmatrix}
\RoM{\theta} = \begin{pmatrix}\cos(\theta) & -\sin(\theta)\\ \sin(\theta) & \cos(\theta)\end{pmatrix},\quad \theta \in \bT,
\end{equation} to the argument of a generator function, where $\bT= \left[-\frac{\pi}{2},\frac{\pi}{2}\right)$. Note that the parameter $\alpha$ in \eqref{eq:scalingmatrix} controls the degree of anisotropy and interpolates between isotropic scaling ($\alpha = 1$), which is used in the definition of wavelet-based systems, and fully anisotropic scaling ($\alpha = 0$), as applied in the definition of ridgelets.

For a fixed parameter $\alpha \in [0,1]$, a single $\alpha$-molecule is thus defined by its corresponding generator and a point $(s,\theta,y)$ in the parameter space
\begin{equation}
\bP:= \bR_{>0} \times \bT \times \bR^2,
\label{eq:amparameterspace}
\end{equation}
where $s$ is a scaling, $\theta$ a rotation, $y$ a translation parameter, and $\bR_{>0}$ denotes the positive real numbers.

Let $\Lambda$ be an index set and $\alpha \in [0,1]$ be fixed. A family of functions $\left(m_\lambda\right)_{\lambda\in\Lambda}$ is called a system of $\alpha$-molecules of order $(L,M,N_1,N_2)$ with $L,M,N_1,N_2 \in \bNz \cup \{\infty\}$, if each molecule can be written as 
\begin{equation}
m_\lambda(x) = s_\lambda^{(1+\alpha)/2}g^{(\lambda)}\left(A_{s_\lambda,\alpha}R_{\theta_\lambda}(x-y_\lambda)\right),
\label{eq:defalphamolecule1}
\end{equation}
where $(s_\lambda,\theta_\lambda,y_\lambda) \in \bP$ is a point in the $\alpha$-molecule parameter space and the corresponding generator $g^{(\lambda)} \in L^2(\bR^2)$ satisfies
\begin{equation}
\abs{\partial^\rho \widehat{g^{(\lambda)}}(\xi)} \lesssim \min\left\{1, s_\lambda^{-1} + \abs{\xi_1}  + s_\lambda^{-(1-\alpha)}\abs{\xi_2}\right\}^M \cdot (1 + \abs{\xi}^2)^{\frac{-N_1}{2}} \cdot (1 + (\xi_2)^2)^{\frac{-N_2}{2}}
\label{eq:defalphamolecule2}
\end{equation}
for all $\abs{\rho} \leq L$, where $\lesssim$ means that there exists a constant $C > 0$ such that the left hand side is smaller than the right hand side times $C$ for all points in the considered domain. \rev{If one or more of the parameters $L$, $M$, $N_1$, and $N_2$ equal $\infty$, {then \eqref{eq:defalphamolecule2} is interpreted to hold for arbitrarily large values of the respective parameter(s).}}

The authors of \cite{grohs2016alpha} aimed to provide a maximum amount of flexibility by allowing each molecule of an $\alpha$-molecule system to be based on a different generator in order to allow shears to be replaced by rotations. In practice, however, systems of $\alpha$-molecules are often constructed from only a small number of generators. For instance, wavelet orthonormal bases of $L^2(\bR^2)$ are typically based on four generators, while so-called cone-adapted shearlet systems (see, e.g., \cite{kutyniok2012shearlets}) only require three different generators. Note that in the latter case, due to the application of shear matrices instead of rotation matrices, the elements of a cone-adapted shearlet system can not directly be expressed in the form \eqref{eq:defalphamolecule1}. However, it can be shown that certain cone-adapted discrete shearlet systems are equivalent to an $\alpha$-molecule system that is based on an infinite number of generators, all of which satisfy Condition~\eqref{eq:defalphamolecule2} \cite{grohs2016alpha}.
%\begin{figure}[t!]
%	\centering
%	\subfloat[]{\label{fig:freqpartitions1}\includegraphics[width=0.3\textwidth]{figs/waveletpartition_gray.png}}\myhspace\subfloat[]{\label{fig:freqpartitions2}\includegraphics[width=0.3\textwidth]{figs/curveletpartition_gray.png}}\myhspace\subfloat[]{\label{fig:freqpartitions3}\includegraphics[width=0.3\textwidth]{figs/shearletpartition_gray.png}}
%	\caption{Schematic depiction of partitions of the Fourier domain induced by (a): wavelets ($\alpha = 1$), (b): curvelets ($\alpha = 1/2$) and (c): shearlets ($\alpha = 1/2$). In each image, the areas drawn in the same shade of gray indicate the frequency support of $\alpha$ molecules associated with the same generator and the same scaling and rotation parameters. \todo{check image rights/remake them}}
%	\label{fig:freqpartitions}
%\end{figure}

Condition~\eqref{eq:defalphamolecule2} requires the generators of an $\alpha$-molecule to satisfy certain time-frequency localization properties, where the parameter $L$ describes the degree of spatial localization, $M$ is the number of directional vanishing moments and $N_1$ and $N_2$ describe the smoothness of a generator.

The transform of a function $f\in L^2(\bR^2)$ with respect to an $\alpha$-molecule system $\left(m_\lambda\right)_{\lambda\in\Lambda} \subset L^2(\bR^2)$ can be defined via the $L^2$-inner product:
\begin{equation}
(\AMT f)(g^{(\lambda)},s_\lambda,\theta_\lambda,y_\lambda) = \ip{f}{m_\lambda} = s_\lambda^{(1+\alpha)/2}\int\limits_{\bR^2}f(x)\overline{g^{(\lambda)}\left(A_{s_\lambda,\alpha}R_{\theta_\lambda}(x-y_\lambda)\right)}\mathrm{d}x.
\end{equation}
In the remainder of this paper, we will often refer to the values $\ip{f}{m_\lambda}$ as the \emph{coefficients} of an $\alpha$-molecule-based transform of a signal $f$.

\subsection{Edge and Ridge Detection Via Phase Congruency}
Two classes of feature detection algorithms served as the inspiration for the methods in this paper: phase congruency and shearlet-based methods. For a more detailed exposition about the development of feature detection based on phase congruency, see \cite[Section 4.2.2]{Rei18}.  The key idea underlying using phase congruency to detect features is that the inner products of even- or odd-symmetric functions centered at  even- or odd-symmetric features have certain traits, like \rev{having similar values, or having similarly large values.} The original phase congruency measures used complex exponentials as the analyzing functions ~\cite{AdBe1985,MRBO1986,MoOw1987,VeOw1990,Kov1999}. 
However, such an approach attempts to define a local property 
%The apparent issue behind computing \eqref{eq:phasecongruency} is that phase congruency is a local property but 
based on the coefficients of the discrete Fourier transform, which -- when viewed independently -- only provide information about \emph{which} frequency components are present in a signal but not \emph{where} they occur. A possible solution was given by Kovesi, who proposed extracting the local phase congruency by using sets of differently scaled complex-valued wavelets \cite{Kov1999}. Similar to the use of sines and cosines in the local energy approach, the real and imaginary parts of these wavelets formed Hilbert transform pairs, with the real part being even- and the imaginary part being odd-symmetric. By considering multiple scalings, Kovesi was able to combine phase information from different parts of the frequency domain and to eventually derive a dimensionless and easy-to-compute complex wavelet-based phase congruency measure for a \rev{one-dimensional} function $f\in L^2\rev{(\bR)}$ and a location $x$, namely
\begin{equation}
\label{eq:phasecongruencywavelets}
\operatorname{PC}_\psic(f,x) = \frac{\abs{\sum\limits_{a \in A}\iptwo{f}{\psic_{a,x}}}}{\sum\limits_{a\in A}\abs{\iptwo{f}{\psic_{a,x}}}+\epsilon},
\end{equation}
where $\psic \in L^2(\bR)$ is a complex-valued wavelet with certain symmetry properties, $A\subset \bR_{>0}$ a set of scaling parameters, $\psic_{a,x} = a^{-\frac{1}{2}}\psi\left(\frac{\cdot - x}{a}\right)$ (i.e. $\psic$ centered at $x$ and scaled by $a$) and $\epsilon > 0$ prevents division by zero. %Analogous to \eqref{eq:phasecongruency}, this measure is defined for any square integrable function $f$ and each point $x$ in its domain. 
\rev{The quantity $\operatorname{PC}_\psic(f,x)$} approaches $1$ if the phases of the local frequency components are in congruency and approaches $0$ if they are widely spread. More details about the generalization of this measure to two dimensions and its implementation may be found here \cite{Kov1999,Kov2000,KovONLINE}.

The measure \eqref{eq:phasecongruencywavelets} can effortlessly be computed and is by construction contrast-invariant. Features such as edges and ridges are picked up solely by the traces their structure leaves in the phases of the frequency representation of a signal rather than the local magnitude of contrast. Furthermore, as Kovesi points out in his original work, it is an interesting property of the phase congruency measure that it does not distinguish edges and ridges.

While contrast-invariance is indeed a highly desirable feature in many edge and ridge detection tasks, localizing a property of the frequency components in the time representation of a signal comes at a price. Indeed, it was already observed in \cite{storath2013amplitudeandsign} and \cite{Rei14} that replacing isotropic wavelets with anisotropically scaled analyzing functions in \eqref{eq:phasecongruencywavelets} deteriorates the detection of features in images rather than improving it. This should come as a surprise, as one would assume that anisotropically scaled elements would be better suited for detecting typically anisotropic features such as edges or ridges. % than their isotropically scaled counterparts. 
Finding a way to bring together the intuitions behind the phase congruency measure $\operatorname{PC}_\psic$ and modern constructions of anisotropic analyzing functions was in fact one of the main factors motivating the present work.
\subsection{Derivative of Gaussian Wavelets}
\label{sec:dogandhilbert}
In order to detect features by testing for local symmetry properties, we require even- and odd-symmetric real-valued functions with good localization in both the time and the frequency domain. Throughout this paper, we call a function $f$ \emph{even-symmetric}, if it satisfies $f(x) = f(-x)$ for all $x$, \emph{odd-symmetric}, if it holds for all $x$ that $f(x) = -f(-x)$, and \emph{symmetric} if one of the above is true.

In the remainder of this work, we will consider two simple tools to construct one-di\-men\-sion\-al even- and odd-symmetric generators with these properties, namely taking derivatives of the Gaussian as well as applying the so-called Hilbert transform. We note that such functions have already appeared in the literature in other multi-scale edge and ridge detection algorithms~\cite{MaHw92,ST02,KaLa16}. A comprehensive discussion of the harmonic analysis used in this section may be found in~\cite{benedetto1996harmonic}. Note that throughout this paper, we denote the Fourier transform of a function $f\in L^1(\rev{\bR^d})$ with $\hat{f}({\xi}) = \rev{\int_{\bR}f(x) e^{-2\pi \iu \ip{{x}}{{\xi}}_{\ell^2}}\mathrm{d}{x}}$ and consider the usual extension to $L^2(\bR^d)$. For $k\in \bNz$, let us denote the $k$-th derivative of the unnormalized Gaussian by
\begin{equation}
G_k(x) = (-1)^kH_k(x)e^{-x^2} =  (-1)^kH_k(x) G_0(x),
\label{eq:dogtimedomain}
\end{equation}
where $H_k$ is the $k$-th Hermite polynomial (see, e.g., \cite{szego1939orthogonal}). Given standard results from harmonic analysis, it follows that
\begin{equation}
\widehat{G_k}(\xi) =  (2\pi\iu\xi)^k \sqrt{\pi} e^{-\pi^2\xi^2}.
\label{eq:dogfreqdomain}
\end{equation}
%It is further well known that for a differentiable function $f$ with $f,f'\in L^1(\bR)$, its derivative is given in the Fourier domain by
%\begin{equation}
%\widehat{f'}(\xi) = 2\pi\iu\xi\hat{f}(\xi),
%\end{equation}
%and that the Gaussian $G_0(\sqrt{\pi}x) = e^{-\pi x^2}$ is a fixed point of the Fourier transform. $G_k$ can thus be expressed in the Fourier domain by
%\begin{equation}
%\widehat{G_k}(\xi) =  (2\pi\iu\xi)^k \sqrt{\pi} e^{-\pi^2\xi^2}.
%\label{eq:dogfreqdomain}
%\end{equation}
\rev{Due to $G_0$ being even-symmetric, the $k$-th derivative $G_k$ is odd-symmetric if $k$ is odd and even-symmetric if $k$ is even. Consequently, $\widehat{G_k}$ is purely imaginary for odd $k$ and purely real if $k$ is even.} Furthermore, the Gaussian lies in the Schwartz space of rapidly decreasing functions. That is, the Gaussian is infinitely differentiable and all of its derivatives decay faster than the inverse of any polynomial. \rev{With the zero-mean property $\widehat{G_k}(0) = \int_{\bR}G_k(x)\mathrm{d} x = 0$, this implies that $G_k$ satisfies the wavelet admissibility condition (see, e.g., \cite{Dau92}) for any integer $k>0$. Note that, due to the zero-mean property, the functions $G_k$ are orthogonal to constant functions and the associated wavelet coefficients are hence invariant under the addition of constants.}

The Hilbert transform is a bounded linear operator on $L^2(\bR)$ and defined for functions $f\in L^2(\bR)$ by
%\begin{equation}
%(\HT f)(x) = \frac{1}{\pi}\int_\bR \frac{f(t)}{t - x} \mathrm{d}t,
%\label{eq:hilberttransformtimedomain}
%\end{equation}
%or equivalently in the Fourier domain via
\begin{equation}
\widehat{\HT f}(\xi) = -\iu \sgn (\xi)\hat{f}(\xi),
\label{eq:hilberttransformfreqdomain}
\end{equation}
where $\sgn$ denotes the sign function.
%
%\begin{figure}[t!]
%	\centering
%	\includegraphics[width= 0.35\textwidth]{figs/dog1.png} \hspace{1cm} \includegraphics[width= 0.35\textwidth]{figs/dog2.png}
%	\caption{The first and the second derivative of the Gaussian plotted alongside their respective Hilbert transforms.}
%	\label{fig:dogwavelets}
%\end{figure}
It is easy to see from \eqref{eq:hilberttransformfreqdomain} that if $\hat{f}$ is purely real, $\widehat{\HT f}$ is purely imaginary and vice versa. This implies that for any even-symmetric function $f$, its Hilbert transform $\HT f$ is odd-symmetric, while any odd-symmetric function will become even-symmetric under the Hilbert transform. Further, the Hilbert transform leaves the magnitude of the Fourier transform invariant. \rev{In particular, this implies that the zero-mean property as well as the wavelet admissibility condition are preserved by the Hilbert transform} and that for any function $f\in L^2(\bR)$, it holds that $\|f\|_2 =\|\HT  f\|_2$. Note that the latter does not hold for the $L^1$-norm, \rev{where for a Schwartz function $f$, $\HT f \in L^1$ if and only if it has the zero-mean property}. Two examples of even- and odd-symmetric wavelets based on derivatives of the Gaussian are plotted along with their respective Hilbert transforms in the Supplementary Materials (see Figures~\ref{fig:dog1},~and~\ref{fig:dog2}).
\subsection{Maximum Point Estimation From Discrete Samples}
\label{sec:maxpointest}
Each two-dimensional analyzing function within a system of $\alpha$-molecules is associated with a certain scaling and rotation parameter. These parameters can be used to obtain first estimates of the local tangent direction and width of a feature by considering the most significant analyzing function (i.e. the member of a system of functions yielding the largest coefficient). To obtain more precise measurements, we consider a refinement procedure which allows us to estimate the maximum point of a function defined in the continuum from only a few discrete samples.

Let $f\in C^2(\bR)$, $\{x_n\}_{n=1}^N \subset \bR$ be a strictly increasing sequence of $N\in\bN$ sampling points (i.e. $x_n < x_{n+1}$ for all $1 \leq n \leq N - 1$) and denote
\begin{equation}
n^* = \argmax\limits_{n\leq N}f(x_n).
\end{equation}
If $1<n^*<N$, the function $f$ has at least one local maximum point in the interval $[x_{n^*-1},x_{n^*+1}]$, which is in real-world applications often taken as a best guess for the global maximum point of $f$. One method of refining the estimate $x_{n^*}$ for a local maximum point is to assume that $f$ can be approximated on $[x_{n^*-1},x_{n^*+1}]$ by the unique parabola fit through the points $(x_{n^*-1},f(x_{n^*-1}))$,$(x_{n^*},f(x_{n^*}))$, and $(x_{n^*+1},f(x_{n^*+1}))$, that is
\begin{equation}
f(x) = c_2(x-x_{n^*})^2 + c_1(x-x_{n^*}) + c_0, \quad \textrm{for $x \in \{ x_{n^*-1}, x_{n^*}, x_{n^*+1}\}$}.
\label{eq:maximumestparabola}
\end{equation}
%for $x \in [x_{n^*-1}, x_{n^*+1}]$, where the parameters $c_2, c_1, c_0 \in \bR$ are chosen as
%\begin{align}
%c_2 &= \frac{d_+(f(x_{n^*-1})-f(x_{n^*}))-d_-(f(x_{n^*+1})-f(x_{n^*}))}{d_+((d_-)^2-d_+d_-)},\label{eq:parabolaa}\\ 
%c_1 &= \frac{(d_-)^2(f(x_{n^*+1}) - f(x_{n^*})) - (d_+)^2(f(x_{n^*-1})-f(x_{n^*}))}{d_-(d_+d_--(d_+)^2)} \label{eq:parabolab}\\
%c_0 &= f(x_{n^*}) \label{eq:parabolac},
%\end{align}
%with $d_- = x_{n^*-1}-x_{n^*}$, and $d_+ = x_{n^*+1}-x_{n^*}$. 
The maximum point of the parabola \eqref{eq:maximumestparabola}, denoted as $x^*$, can now be computed by
\begin{equation}
x^* = x_{n^*} - \frac{c_1}{2c_2}.
\label{eq:maximumestfinal}
\end{equation}
%Note that in the case $d_- = -1$ and $d_+ = 1$, Equations~\eqref{eq:parabolaa}~to~\eqref{eq:parabolab} simplify significantly. This simplified case appears in \cite{YLEK2009}, where they use it to refine orientation estimation in a certain shearlet-based algorithm. 
This approach appears in \cite{YLEK2009}, where \rev{it is used} to refine orientation estimation in a certain shearlet-based algorithm. A short review of this and other methods for estimating maximal points from discrete samples in the context of edge and line detection may be found in \cite{Bai03}.
\begin{figure}[t!]
	\centering
	\subfloat[Exact $\argmax$ = $1.058$]{\label{fig:maxpointestimation1}\includegraphics[width=0.3\textwidth]{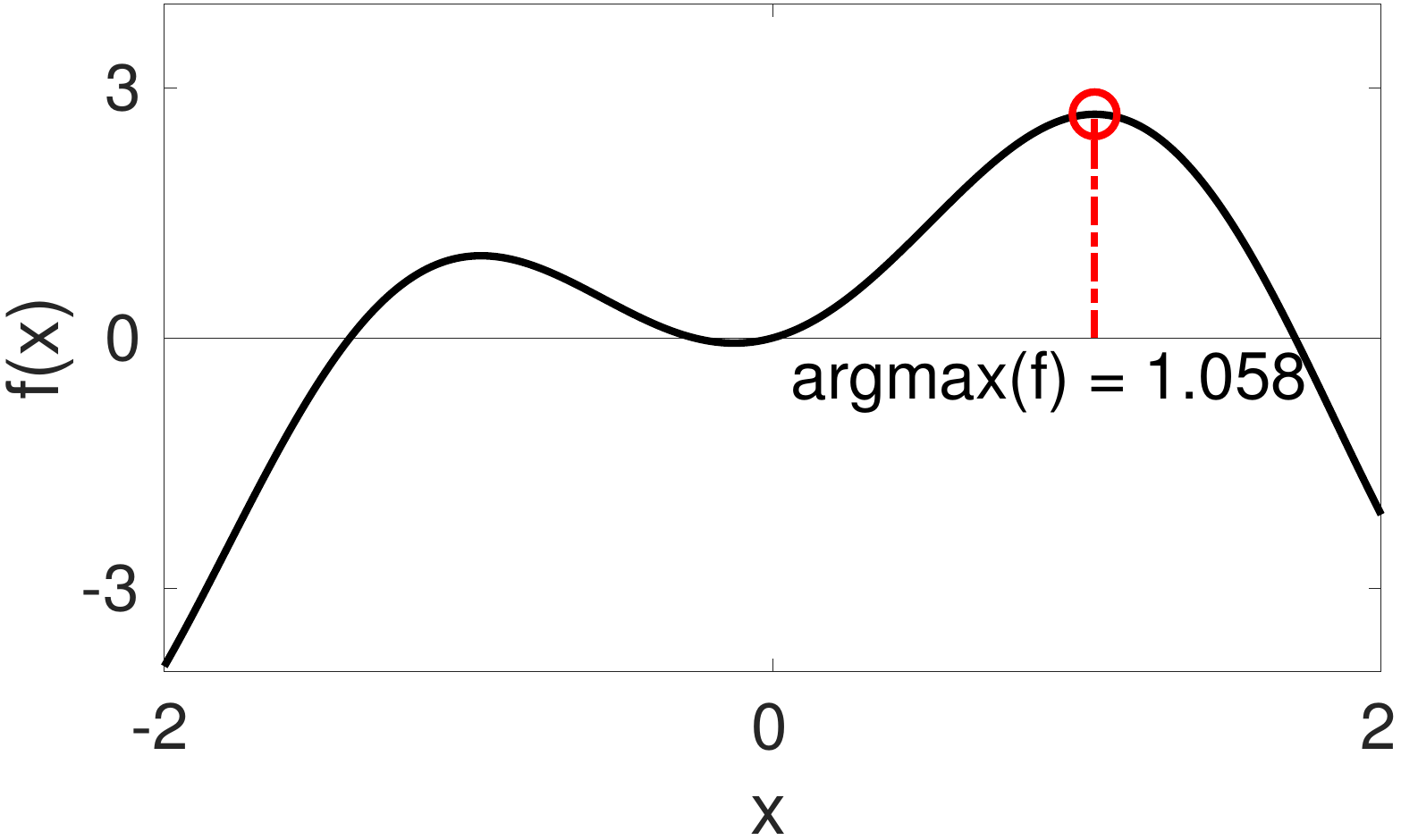}}\myhspace\subfloat[Estimated $\argmax$ = $0.96064$]{\label{fig:maxpointestimation2}\includegraphics[width=0.3\textwidth]{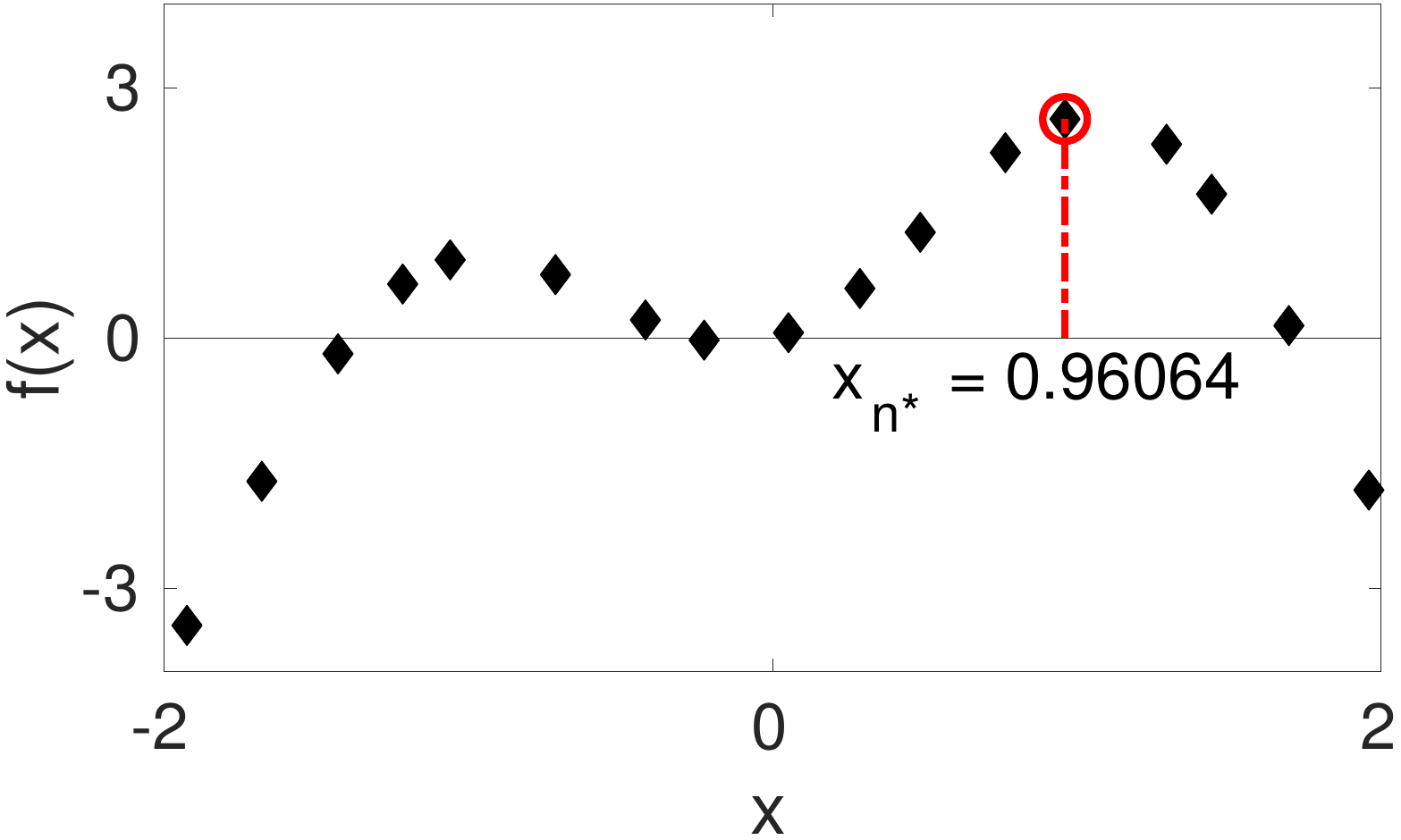}}\myhspace\subfloat[Refined $\argmax$ = $1.0466$]{\label{fig:maxpointestimation3}\includegraphics[width=0.3\textwidth]{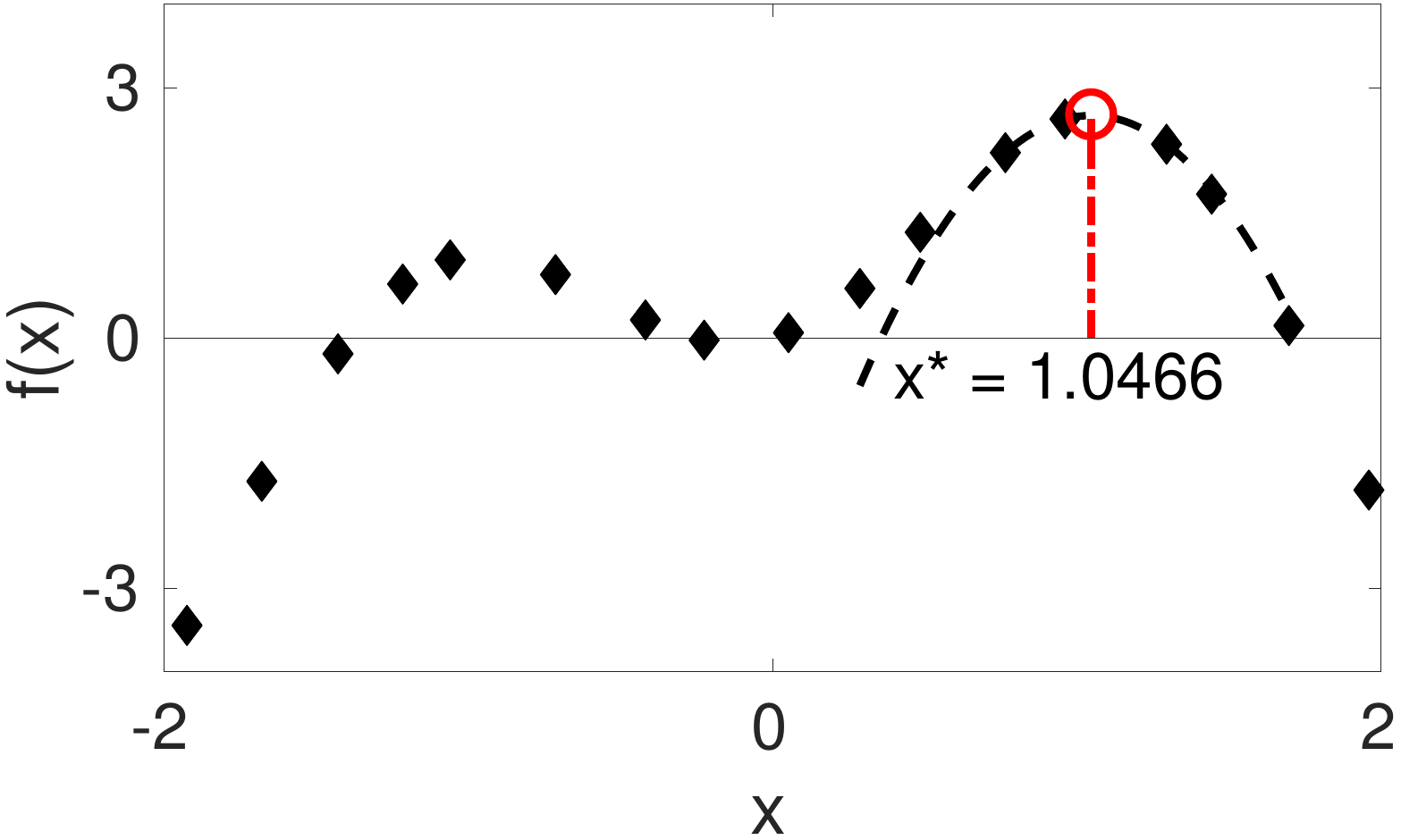}}
	\caption{(a): A $C^2$-function and its maximum point in the interval $[-2,2]$. (b): An estimate $x_{n^*}$ of the maximum point obtained from sampling the function at $17$ different points. (c): A refined estimate $x^*$ of the maximum point obtained from fitting a parabola through $x_{n^*}$ and its nearest neighbors.}
	\label{fig:maxpointestimation}
\end{figure}

Figure~\ref{fig:maxpointestimation} illustrates how \eqref{eq:maximumestfinal} can be used to yield refined estimates of local maxima from a finite number of samples. In this paper, we will use \eqref{eq:maximumestfinal} to improve on estimates of local tangent orientations and widths of features from finite samples of coefficients associated with differently scaled and oriented $\alpha$-molecules.
\section{Symmetric Molecule-Based Feature Detection}
\label{sec:symfd}
In this section, we will \rev{estimate} the likelihood of a certain feature being centered at a point $y\in \bR^2$ in the image domain for a given image $f\in L^2(\bR^2)$ by considering the transform of $f$ with respect to systems of even- and odd-symmetric $\alpha$-molecules. \rev{We will furthermore} demonstrate how additional information about the detected features, namely the local tangent orientations of edges and ridges as well as local widths (diameters) of ridges and blobs and heights (contrasts) can be extracted. Formally, we will define three mappings that serve as feature detectors:
\begin{align}
\text{an edge measure}\quad\EM(f,y)&\colon L^2(\bR^2)\times \bR^2 \to [0,1],\label{eq:measures1}\\
\text{a ridge measure}\quad\RM(f,y)&\colon L^2(\bR^2)\times \bR^2 \to [0,1],\label{eq:measures2}\\
\text{and a blob measure}\quad\BM(f,y)&\colon L^2(\bR^2)\times \bR^2 \to [0,1].\label{eq:measures3}
\end{align}
Furthermore, we will define functions that extract local properties of a feature centered at a given point in the image domain, namely
\begin{align}
\text{tangent orientation measures}&&\quad\OMall(f,y)&\colon L^2(\bR^2)\times \bR^2 \to \left[-\frac{\pi}{2},\frac{\pi}{2}\right),  &&x\in\{\EM,\RM\},\label{eq:measures4}\\
\text{local width measures}&&\quad\WMall(f,y)&\colon L^2(\bR^2)\times \bR^2 \to \bR_{>0},\quad  &&x\in\{\RM,\BM\},\label{eq:measures5}\\
\text{and local height measures}&&\quad\HMall(f,y)&\colon L^2(\bR^2)\times \bR^2 \to \bR,\quad  &&x\in\{\EM,\RM, \BM\}.\label{eq:measures6}
\end{align}
\rev{The measures} $\EM(f, y)$, and $\HME(f, y)$ will be defined in Section~\ref{sec:edgemeasure}\rev{;} $\RM(f, y)$, $\WMR(f, y)$, and $\HMR(f, y)$ in Section~\ref{sec:ridgemeasure}\rev{;} $\BM(f, y)$, $\WMB(f, y)$, and $\HMB(f, y)$ in Section~\ref{sec:blobmeasure}\rev{;} and $\OME(f, y)$, and $\OMR(f, y)$ in Section~\ref{sec:orientationmeasure}.

\begin{figure}[t!]
	\centering
	\subfloat[Ideal 1D edge]{\label{fig:modeledge1d}\includegraphics[width= 0.3\textwidth]{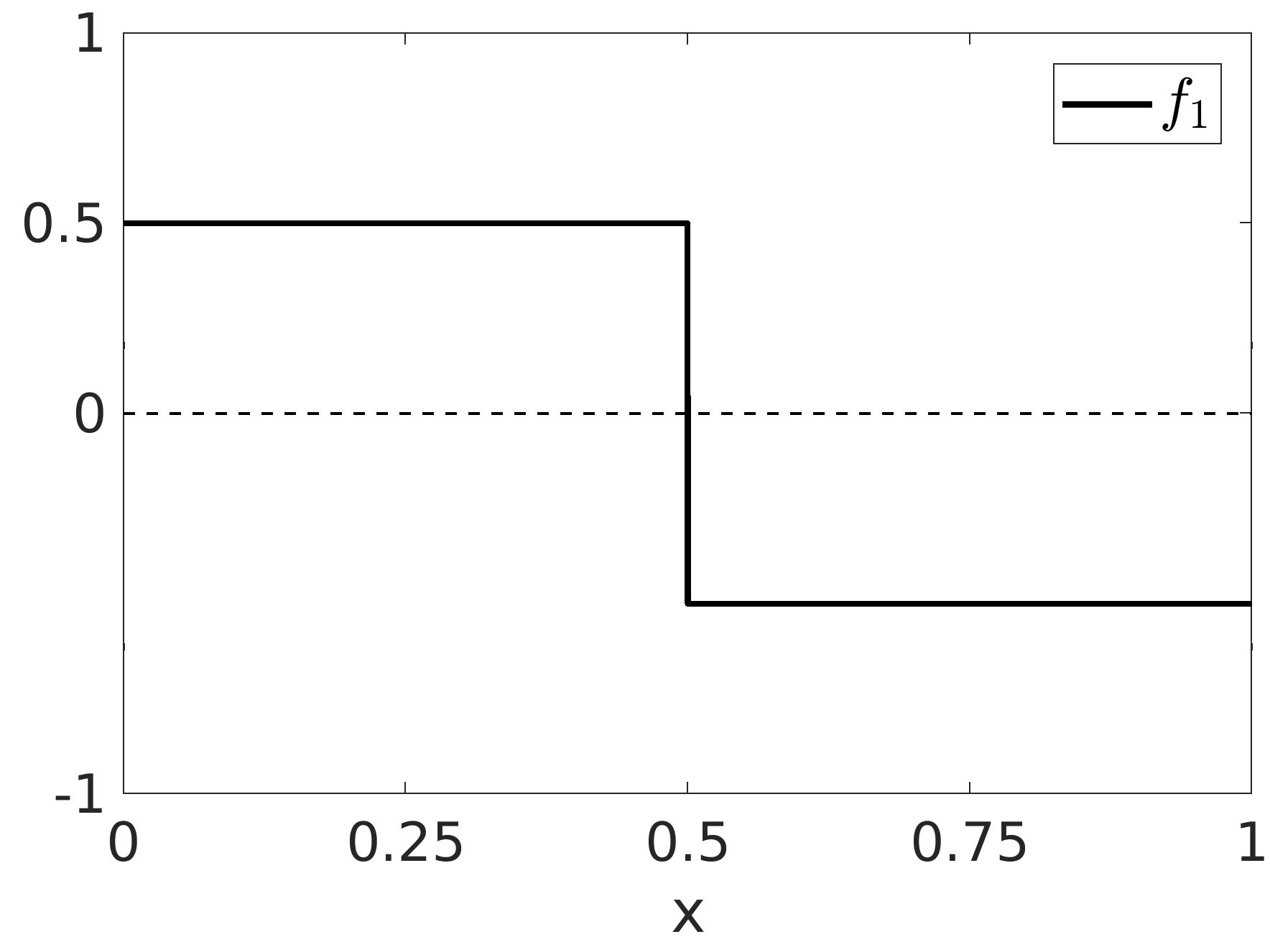}}\myhspace \subfloat[Ideal 1D ridge/blob]{\label{fig:modelridge1d}\includegraphics[width= 0.3\textwidth]{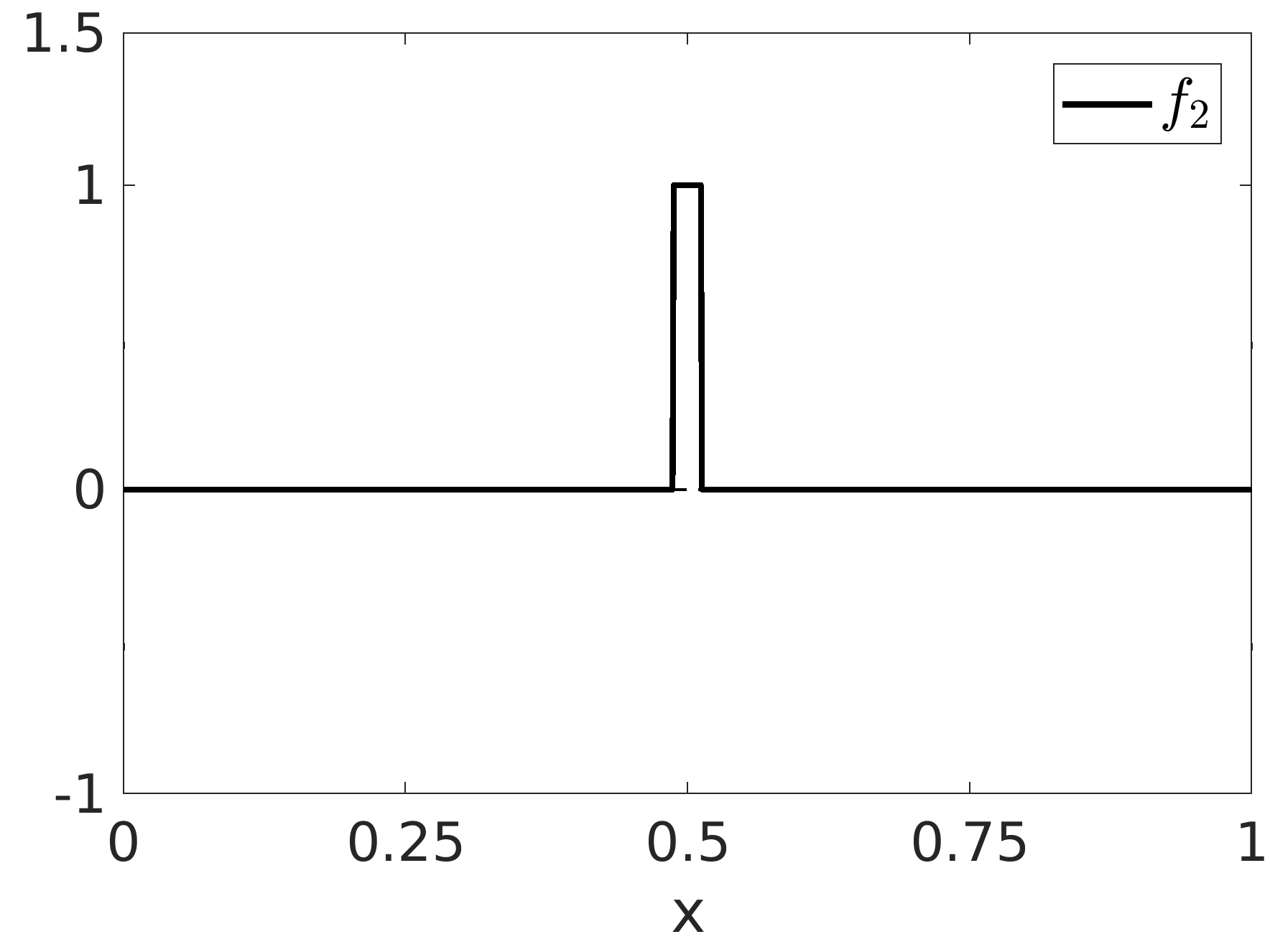}}\myhspace \subfloat[]{\label{fig:dog1scales1d}\includegraphics[width= 0.3\textwidth]{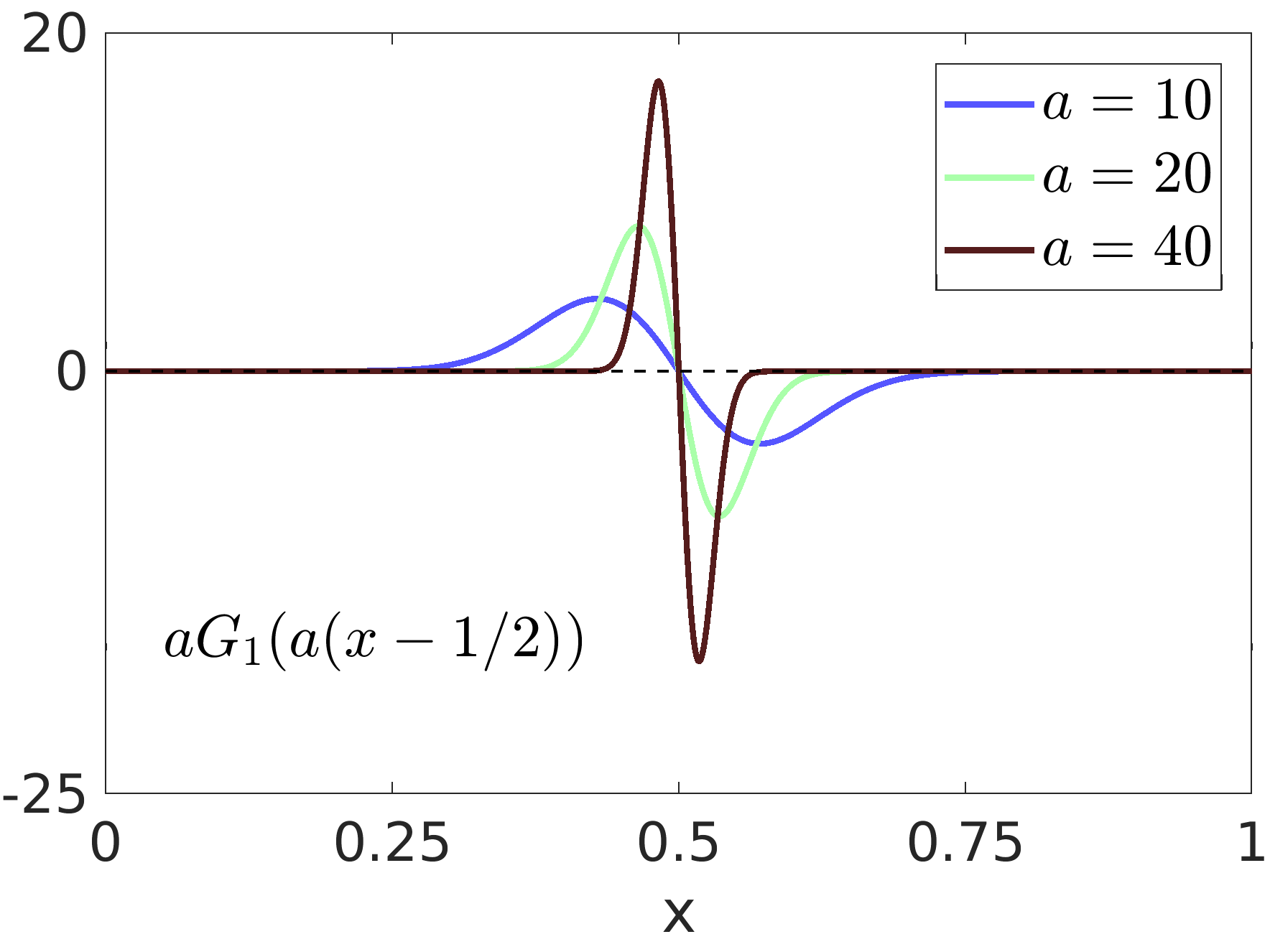}}\\ \subfloat[]{\label{fig:dog2scales1d}\includegraphics[width= 0.3\textwidth]{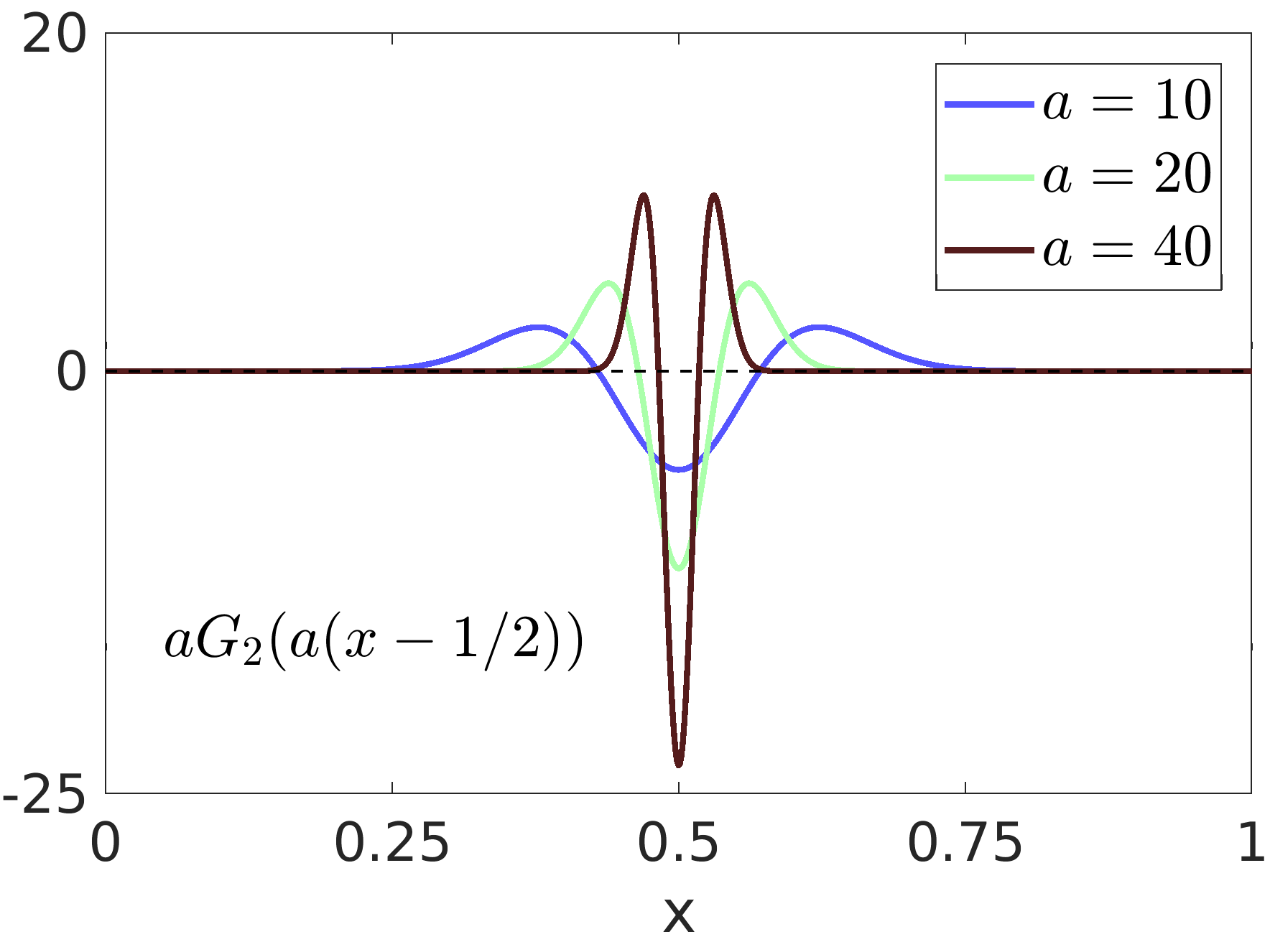}}\myhspace \subfloat[]{\label{fig:dog1coeffsedge1d}\includegraphics[width= 0.3\textwidth]{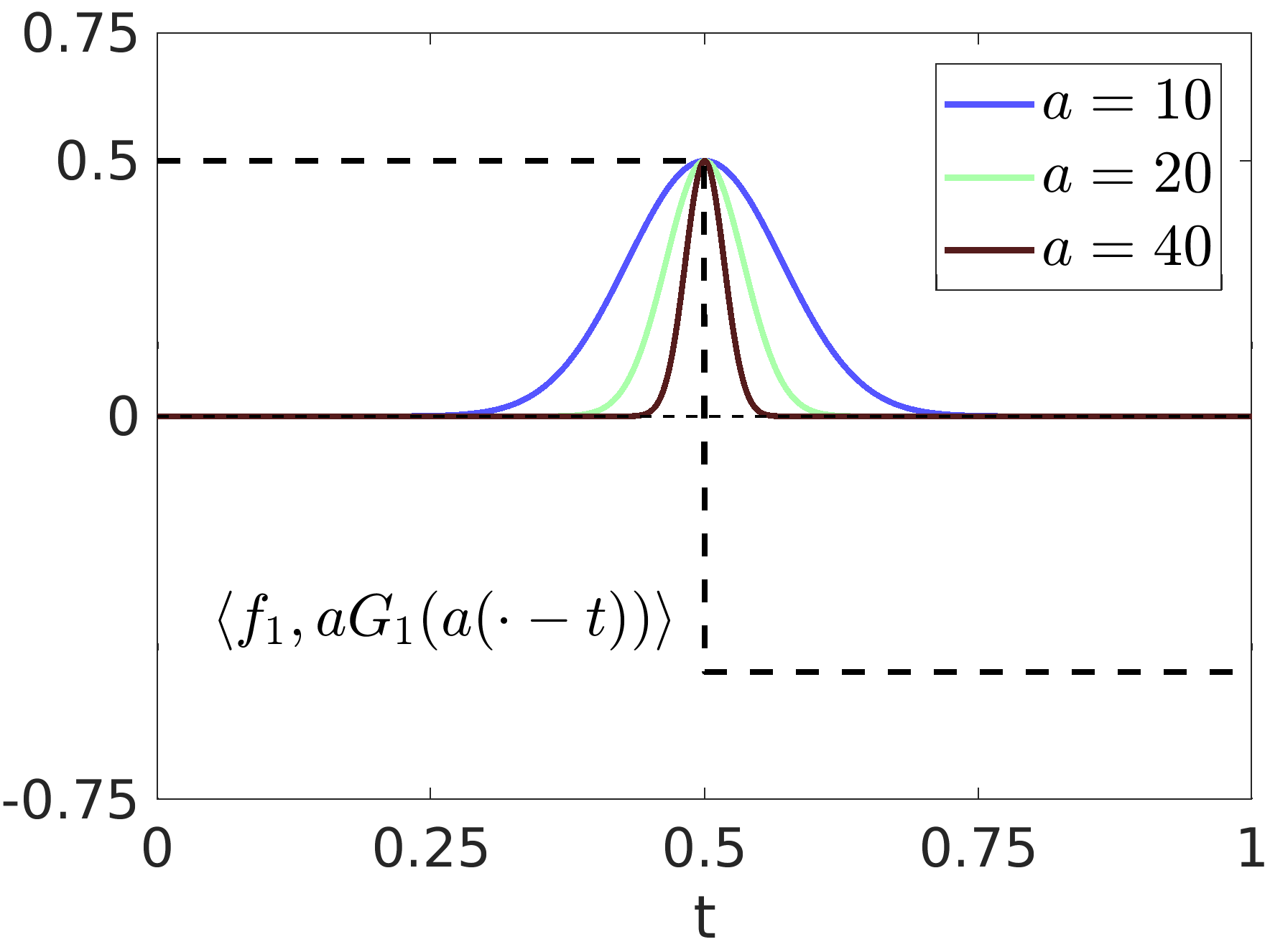}}\myhspace \subfloat[]{\label{fig:dog2coeffsedge1d}\includegraphics[width= 0.3\textwidth]{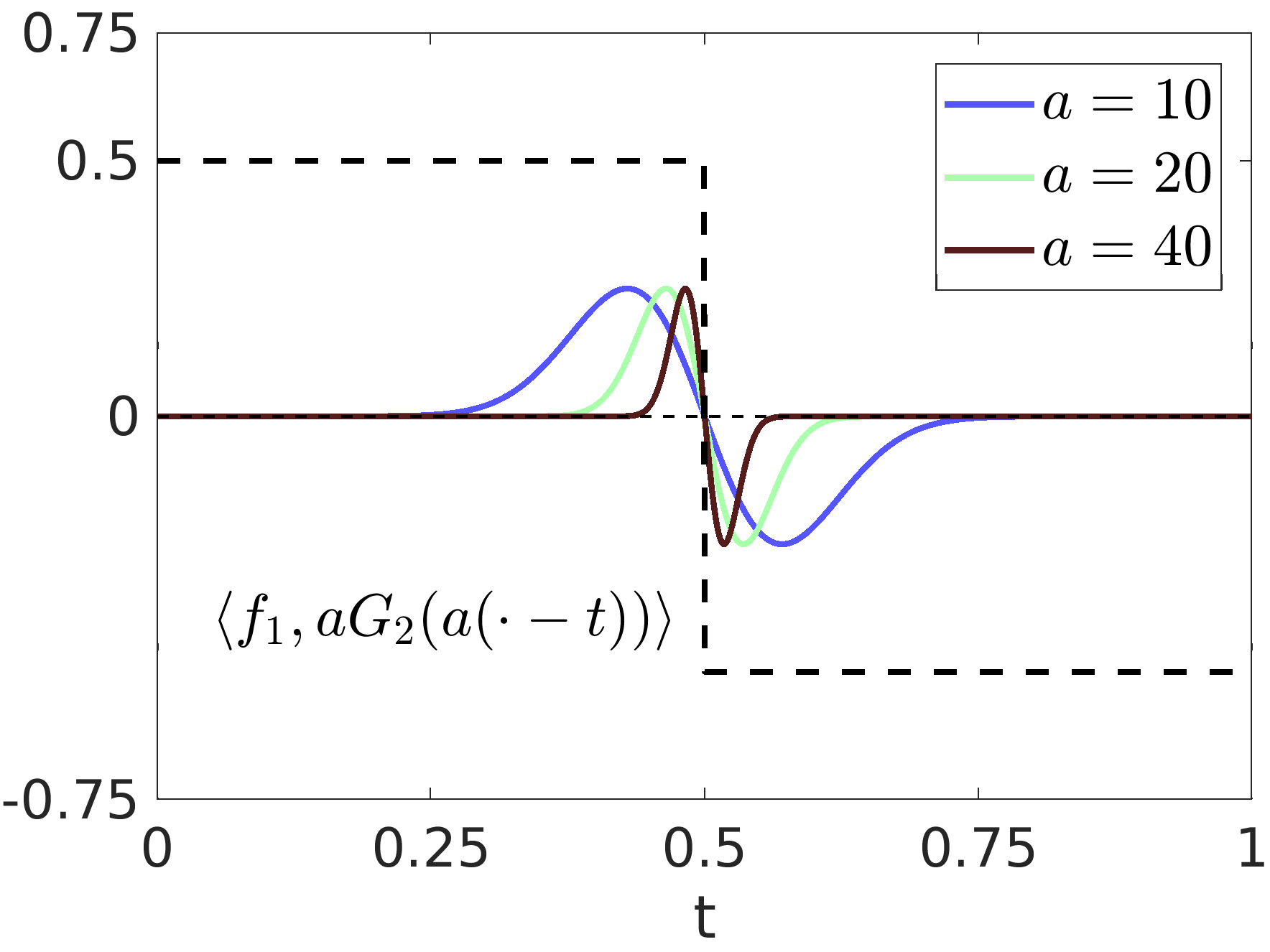}} %\subfloat[]{\label{fig:dog1coeffsridge1d}\includegraphics[width= 0.24\textwidth]{figs/dog1coeffsridge1d.png}}\myhspace %\subfloat[]{\label{fig:dog2coeffsridge1d}\includegraphics[width= 0.24\textwidth]{figs/dog2coeffsridge1d.png}}
	\caption{(a): An idealized one-dimensional edge. (b): An idealized one-dimensional ridge/blob. (c): Three different $L^1$-normalized dilates of the first derivative of the Gaussian centered at $1/2$. (d): Three different $L^1$-normalized dilates of the second derivative of the Gaussian centered at $1/2$. (e): Inner products of the dilates of the first derivative of the Gaussian with an ideal edge plotted as a function of the shift parameter $t$. (f): Inner products of the dilates of the second derivative of the Gaussian with the ideal edge.}
	\label{fig:modeledge1dridgeandbehavior}
\end{figure}

Before turning to the derivation of the measures in \eqref{eq:measures1}~to~\eqref{eq:measures6}, we would like to explain the basic principle of symmetric molecule-based feature detection in the one-dimensional setting, which we emphasize is distinct from phase congruency, although it is inspired by it. Figure~\ref{fig:modeledge1d} depicts what we consider an ideal one-dimensional edge, namely a step discontinuity preceded and followed by two distinct constant functions. An ideal ridge, which could be described as a short constant function interrupting a constant baseline, is shown in Figure~\ref{fig:modelridge1d}. The features plotted in Figures~\ref{fig:modeledge1d}~and~\ref{fig:modelridge1d} are both centered at the point $1/2$ and the main question we aim to answer is how this point could be discriminated from its neighborhood and, in general, from other points not lying in the center of a ridge, an edge, or another kind of feature. Note that the notions of ridges and blobs coincide in the one-dimensional setting.

One way of looking at this problem is to observe that the function plotted in Figure~\ref{fig:modeledge1d} is odd-symmetric around $1/2$, while the function shown in Figure~\ref{fig:modelridge1d} is even-symmetric around $1/2$. 
%This is consistent with the phase congruency-based approach to feature detection, in which edges are characterized by phase congruency with respect to the angle $\pi/2$ and ridges are characterized by phase congruency with respect to the angle $0$ (cf. \eqref{eq:phasecongruencywavelets}). In the phase congruency framework, this simply means that in the vicinity of an edge, a function can purely be represented by odd-symmetric sine components, while in the vicinity of a ridge, a function solely consists of even-symmetric cosine components. 
For edges, we can additionally observe a certain kind of self-similarity. When restricting the function plotted in Figure~\ref{fig:modeledge1d} to any neighborhood around the point $1/2$ and then scaling it back to the full interval $[0,1]$, the function will remain unchanged. Ridges on the other hand are not defined by a single step-discontinuity but by two step-discontinuities which are in close proximity of each other. Any ridge can thus be associated with a width (i.e. the distance between its step-discontinuities) and is not invariant under the restricting and rescaling operation described above. In the one-dimensional setting, we will only deal with the case of edge detection but this issue will be taken into consideration when defining the two-dimensional ridge measure in Section~\ref{sec:ridgemeasure}.

To summarize: In order to detect edges, we aim to test each point in the domain of a function for the scale-invariant symmetry and self-similarity properties described in the preceding paragraph. If a point is at the location of an edge, restrictions of the analyzed function to reasonably small neighborhoods around this point should look roughly the same and, in particular, be odd-symmetric when centered at the analyzed point. A simple and efficient way of testing for these properties is to consider the coefficients associated with differently scaled even- and odd-symmetric wavelets that are centered around the point of interest. %Again, this approach is consistent with \eqref{eq:phasecongruencywavelets}, where complex-valued wavelets whose real part is even- and whose imaginary part is odd-symmetric are applied to approximate the phase congruency measure \eqref{eq:phasecongruency}.

Figures~\ref{fig:dog1scales1d}~and~\ref{fig:dog2scales1d} show three differently scaled versions of the odd-symmetric first and the even-symmetric second derivative of the Gaussian (cf.\ \eqref{eq:dogtimedomain}~and~\eqref{eq:dogfreqdomain}), all of which are $L^1$-normalized to one. Figures~\ref{fig:dog1coeffsedge1d}~and~\ref{fig:dog2coeffsedge1d} depict the corresponding inner products with the idealized one-dimensional edge as a function of the translation parameter $t$. For the remainder of this paper, we shall call inner products of a function $f$ with an even-symmetric wavelet or $\alpha$-molecule the \textit{even-symmetric coefficients} of $f$ and the inner products with an odd-symmetric wavelet or $\alpha$-molecule the \textit{odd-symmetric coefficients}. Independently of scale, the even-symmetric coefficients shown in Figure~\ref{fig:dog2coeffsedge1d} go to zero at the location of the step-discontinuity, as the even-symmetric wavelets are orthogonal to the locally odd-symmetric function centered at the point $1/2$. Furthermore, the odd-symmetric coefficients all peak in the point $1/2$ with a maximum value of approximately $1/2$. This is due to the fact that the jump size of the singularity is one and because the first derivative of the Gaussian is positive on $\bR_{<0}$. Consequently, the integral of the $L^1$-normalized first derivative of the Gaussian over $\bR_{<0}$ is $\frac{1}{2}$. 

In summary, Figure~\ref{fig:modeledge1dridgeandbehavior} illustrates that at the location of an ideal edge, $L^1$-normalized symmetric wavelets show scale invariant behavior in the sense that odd-symmetric coefficients remain constant and locally maximal (with respect to translation) under scaling, while even-symmetric coefficients remain fixed at zero. In particular, this behavior is invariant of the jump size of the singularity and thus unaffected by changes in contrast.

To translate the above observations into an edge measure for functions in $L^2(\bR)$, we first use the derivatives of the Gaussian and the Hilbert transform defined in Section~\ref{sec:dogandhilbert} to define sets of odd- and even-symmetric $L^1$-normalized wavelets:
\begin{align*}
\Psie &= \left\{\frac{G_{2k}}{\|G_{2k}\|_1} \colon k\in \bN \right\} \cup \left\{\frac{\HT G_{2k-1}}{\|\HT G_{2k-1}\|_1} \colon k\in \bN \right\} \subset L^2(\bR) \cap L^1(\bR),\\
\Psio &= \left\{\frac{G_{2k-1}}{\|G_{2k-1}\|_1} \colon k\in \bN \right\} \cup \left\{\frac{\HT G_{2k}}{\|\HT G_{2k}\|_1} \colon k\in \bN \right\} \subset L^2(\bR) \cap L^1(\bR).
\end{align*}
\rev{Note that $\|\HT G_{k}\|_1 < \infty$ for all $k\in\bN$ is a consequence of $G_{k}$ being a Schwartz function and the zero-mean property $\widehat{G_{k}}(0) = 0$. To see this, one can split the integral of the Fourier representation $\HT G_{k}(x) = \int_{\bR}\widehat{\HT G_{k}}(\xi)e^{2\pi\iu x \xi}\mathrm{d}\xi$ at the origin and apply integration by parts to show that $\HT G_{k}$ is dominated by $Cx^{-2}$ for some $C>0$ and hence in $L^1(\bR)$. }

Let $\psio \in \Psio$, $\psie \in \Psie$, \rev{$a>1$} be fixed, $J = \{j_n\}_{n=1}^{N_J} \subset\bZ$ be an increasing sequence of $N_J \in \bN$ scaling parameters and denote $L^1$-normalized dilates and shifts of the generating wavelets $\psio$ and $\psie$ with
\begin{equation}
\psio_{j,y}(x) = a^{j}\psio(a^j(x - y)), \quad \psie_{j,y}(x) = a^{j}\psie(a^j(x - y)),
\end{equation}
for scaling parameters $j\in J$ and translation parameters $y\in\bR$. Note that $f \mapsto a^jf(a^j\cdot)$ defines an isometric mapping of $L^1$. As already discussed earlier in the case of the first derivative of the Gaussian, the odd-symmetric coefficient at the location of an ideal edge is, independently of scaling, fully determined by the jump size of the discontinuity and the integral of the generating wavelet $\psio$ over $\bR_{\leq 0}$. To be precise, the value
\begin{equation}
\Kpsio = \int_{-\infty}^0 \psio(x) \mathrm{d}x,
\label{eq:kpsio}
\end{equation}
is equal to the odd-symmetric coefficients at the location of an ideal edge with jump size one. With a soft-thresholding parameter $\beta > 0$ that corresponds to the minimal jump size that is required for the detection of an edge and a fixed scaling offset for the even-symmetric wavelets $\je\in \bR_{>0}$, a one-dimensional edge measure for a function $f\in L^2(\bR)$ and $y\in \bR$ is given by
\begin{equation}	
\EMDT(f,y) = \frac{\abs{\sum\limits_{j\in J}\ip{f}{\psio_{j,y}}} - \sum\limits_{j \in J}\abs{\ip{f}{\psie_{j-\je,y}}} - \beta N_J\Kpsio}{N_J\max\limits_{j\in J}\left\{\abs{\ip{f}{\psio_{j,y}}}\right\}+\epsilon},
\label{eq:emdt}
\end{equation}
where $\epsilon > 0$ prevents division by zero. The measure $\EMDT$ is bounded above by one but to ensure a mapping to $[0,1]$, we finally set
\begin{equation}
\EMD(f,y) = \max\{0,\EMDT(f,y)\}.
\label{eq:emd}
\end{equation}
For example, if a function $g$ has an ideal edge at a point $y^*$ with jump size $s$, then all odd-symmetric coefficients will virtually\footnote{Note that the derivatives of the Gaussian are not compactly supported in the time domain and that an ideal edge that separates two different constant functions covering the whole real line is not in $L^2(\bR)$ (cf. Figure~\ref{fig:modeledge1d}).} be equal to $\Kpsio$ times the jump size $s$, and all even-symmetric coefficients will be zero. That is, when disregarding $\epsilon$, the measure $\EMDT(g,y^*)$ reduces to
\begin{equation}
\EMDT(g,y^*) = \frac{N_J s\Kpsio - \beta N_J\Kpsio}{ N_J s\Kpsio} = 1 - \frac{\beta}{s},
\label{eq:betaisminimal}
\end{equation}
illustrating that the jump size of edges that can be detected by the measure $\EMD$ is indeed bounded below by the parameter $\beta$.
\begin{figure}[t!]
	\centering
	\subfloat[Input signal]{\label{fig:edgemeasure1d1}\includegraphics[width=0.3\textwidth]{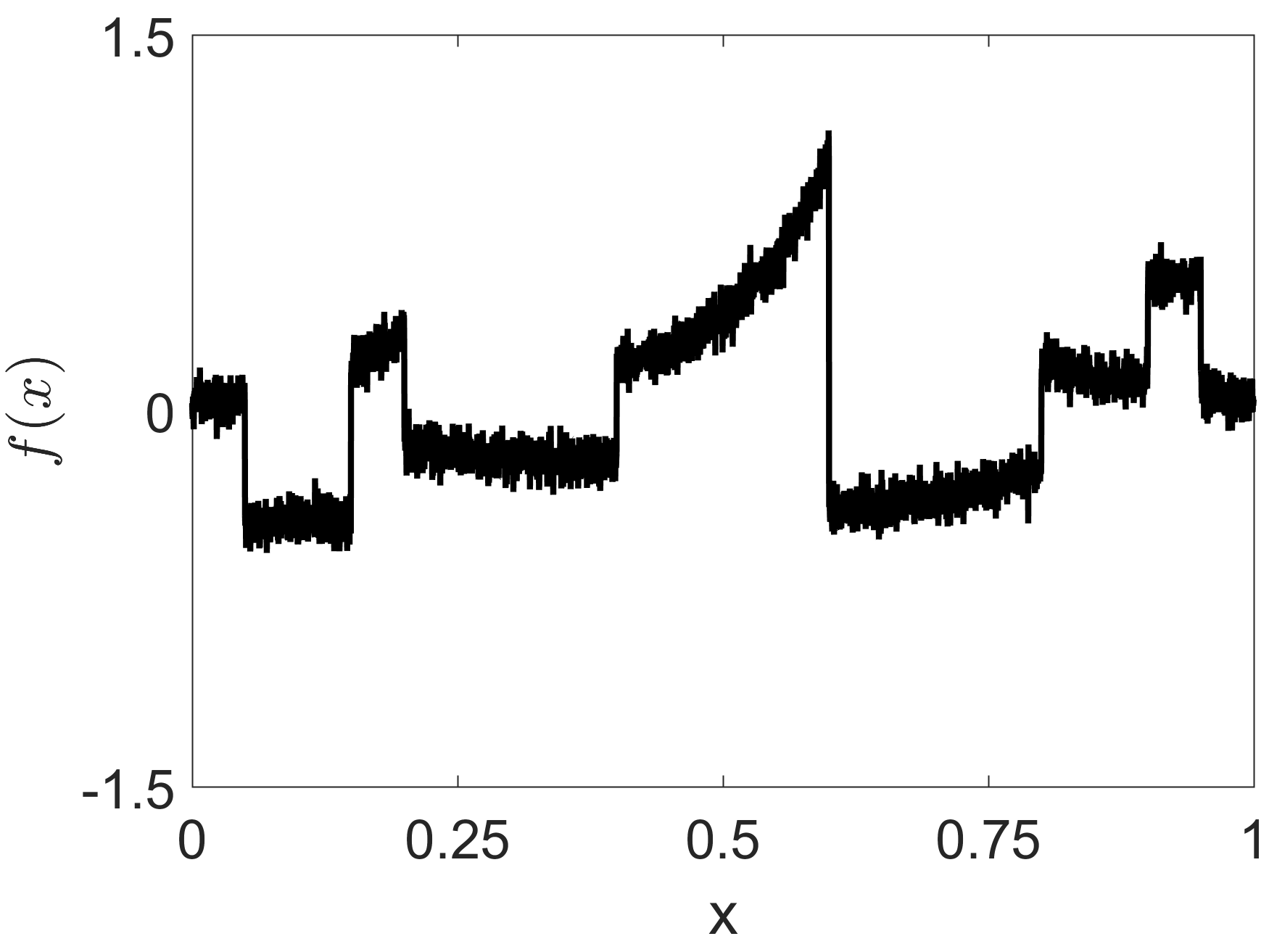}}\myhspace\subfloat[1D edge measure with $\je= 0$]{\label{fig:edgemeasure1d2}\includegraphics[width=0.3\textwidth]{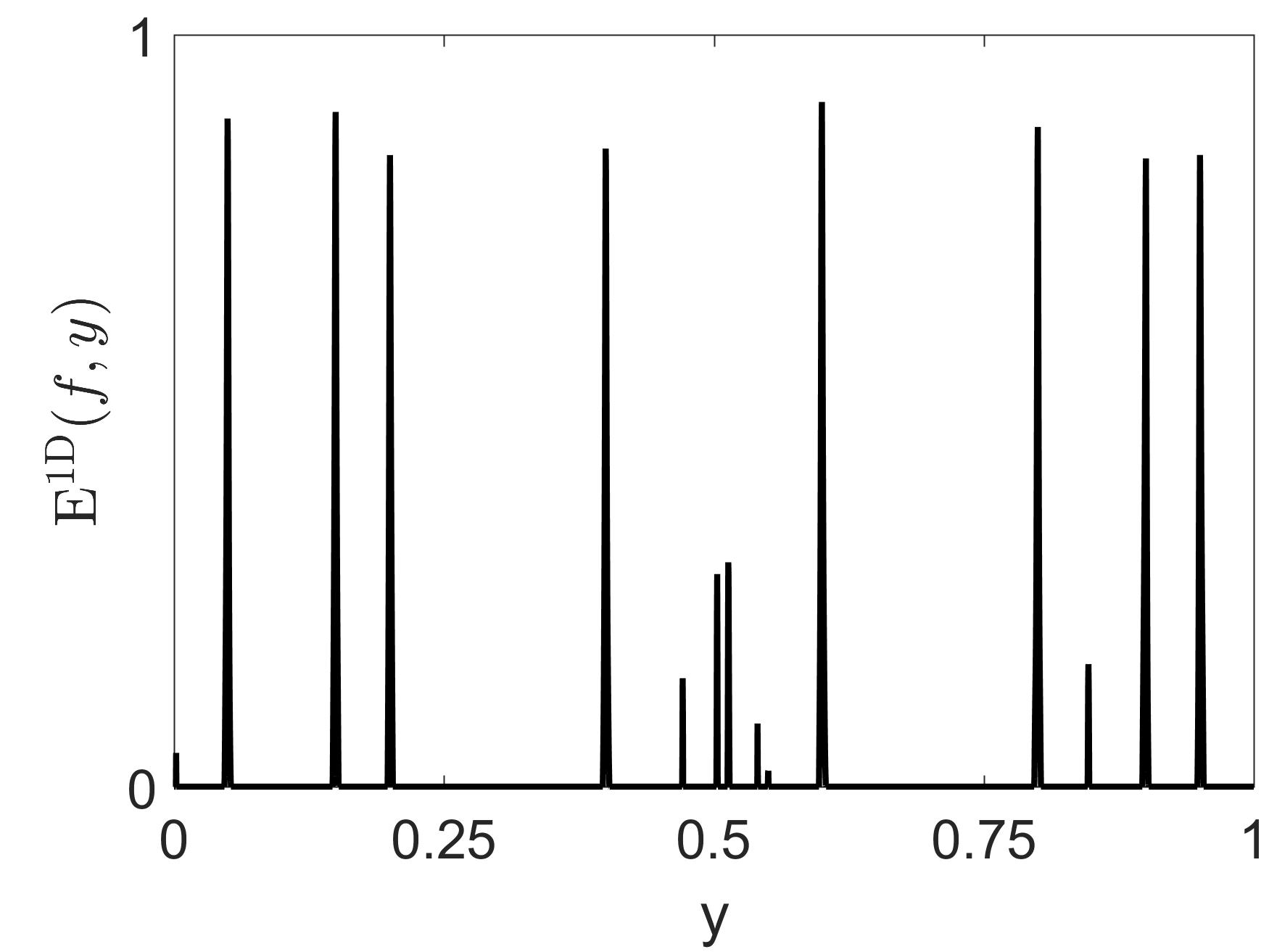}}\myhspace\subfloat[1D edge measure with $\je=2$]{\label{fig:edgemeasure1d3}\includegraphics[width=0.3\textwidth]{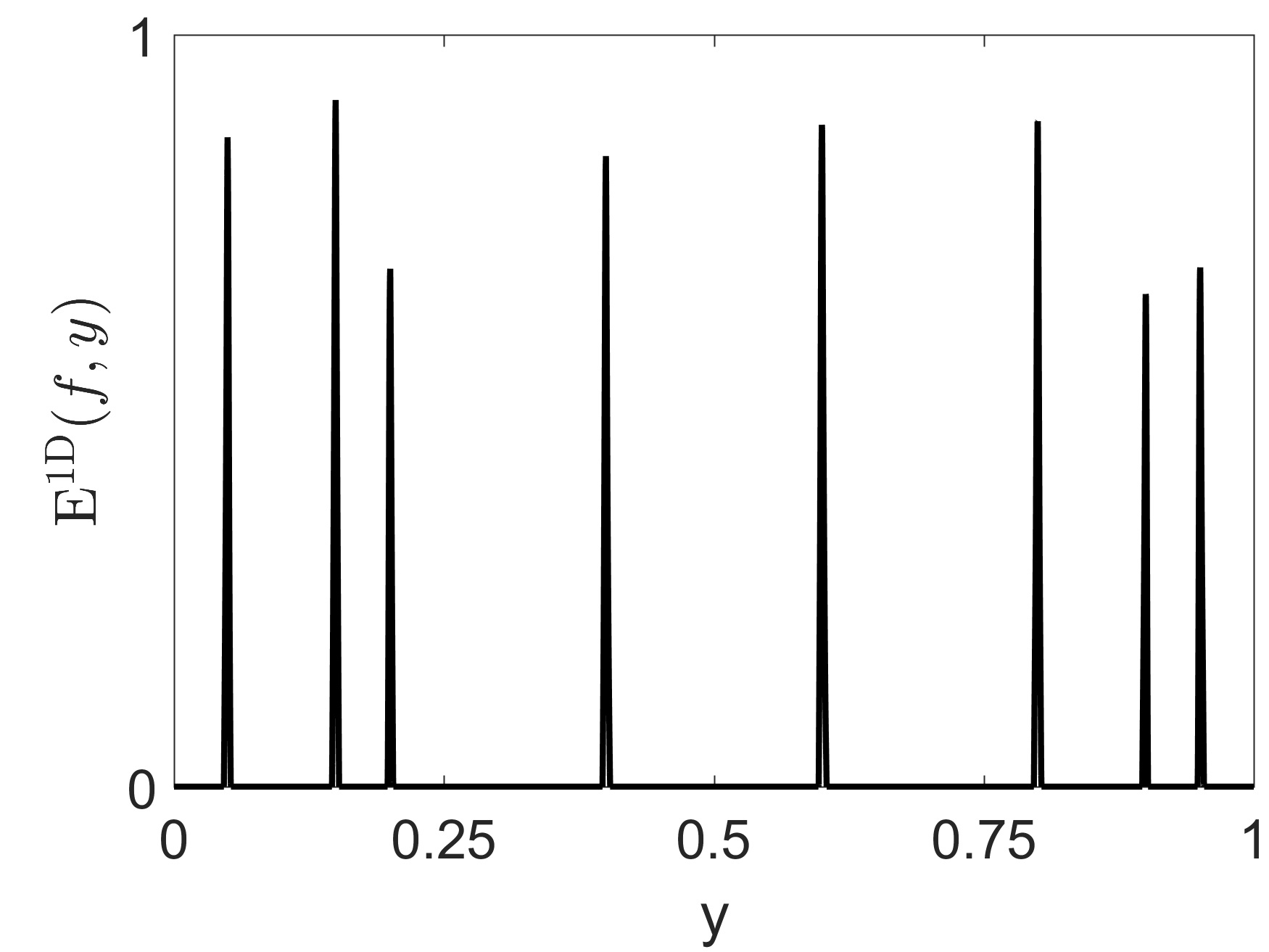}}
	\caption{(a): A piecewise polynomial function perturbed with additive Gaussian white noise (computed using \cite{donoho2006wavelab}) . (b): The one-dimensional edge measure \eqref{eq:emd} with parameters $\psio = \frac{G_1}{\|G_1\|_1}$, $\psie = \frac{\HT G_1}{\|\HT G_1\|_1}$, $a=2$, $N_J = 4$, $\beta = 0.03$ and $\je = 0$. (c): The one-dimensional edge measure \eqref{eq:emd} with a scaling offset $\je = 2$ (other parameters are the same as in \ref{fig:edgemeasure1d2}).}
	\label{fig:edgemeasure1d}
\end{figure}

The term $\beta\Kpsio$ in \eqref{eq:emdt} can also be seen as a soft-thresholding parameter that is applied to all odd-symmetric coefficients and thereby implicitly denoises the analyzed signal $f$. In practice, the parameter $\beta$ is typically chosen as a function of the expected level of noise present in a given signal or image.

\rev{Apart from} the lower bound on $s$ introduced by $\beta$, the measure $\EMD$ is by construction contrast-invariant. Due to normalization with respect to the odd-symmetric coefficient with the greatest magnitude, $\EMD$ is independent of the jump size of the discontinuity associated with an edge. In particular, regardless of the jump size, $\EMD$ will be close to one at the location of an ideal edge and only diminish at a point $y$ if the analyzed function has even-symmetric components around $y$, or if the odd-symmetric coefficients at $y$ have a strong variance with respect to scaling. $\EMD$ thus provides a continuous measure of the structural similarity of the neighborhood of a given point with an ideal edge, which sets it categorically apart from most gradient-based approaches to edge detection that are measuring jump sizes.

Broadly speaking, the odd-symmetric coefficients in \eqref{eq:emdt} serve as evidence in favor of -- while the even-symmetric coefficients serve as evidence against -- the presence of an edge at a given location. To suppress the emergence of false positives in the extended neighborhood of an actual edge, it has proven useful to consider even-symmetric wavelets that have lower frequencies than their odd-symmetric counterparts, which can be achieved by setting the scale offset parameter $\je$ to any real value greater than $0$. \rev{By doing so, the even-symmetric coefficients have a slower decay on the left and right side of an edge and can thereby better suppress the faster decaying odd-symmetric coefficients in the neighborhood of an edge.} The effect of the scale offset parameter is illustrated in Figure~\ref{fig:edgemeasure1d}, which visualizes the one-dimensional edge measure $\EMD$ computed on a piecewise polynomial signal with additive Gaussian noise. The odd- and even-symmetric wavelets were chosen to be the first derivative of the Gaussian and its Hilbert transform, that is $\psio = \frac{G_1}{\|G_1\|_1}$ and $\psie = \frac{\HT G_1}{\|\HT G_1\|_1}$. \rev{All numerical results in this manuscript were in fact obtained by considering even- and odd-symmetric generators that form Hilbert transform pairs, that is, up to normalization $\psio = \HT \psie$. Such pairs of wavelets also play an important role in the dual-tree complex wavelet transform \cite{selesnick2005dualtree}. One advantage of considering Hilbert transform pairs is that the respective functions \emph{only} differ with respect to their symmetry properties in the sense that they differ by a $90^\circ$ phase shift but have the same magnitude response in the Fourier domain.} The other parameters were selected as follows: $a=2$, $N_J = 4$, $\beta = 0.03$, $\je = 0$ for the plot in Figure~\ref{fig:edgemeasure1d2} and $\je = 2$ for the plot in Figure~\ref{fig:edgemeasure1d3}.

The main difference between edges and ridges is that edges are locally odd-symmetric features while ridges are locally even-symmetric features. Thus, to obtain a one-dimensional ridge measure, it almost suffices to simply switch the roles of the even- and odd-symmetric wavelets in \eqref{eq:emdt}. However, due to the fact that ridges, in contrast to edges, are associated with widths, \rev{the} even-symmetric coefficients are not scaling invariant at the location of an ideal ridge as it is the case for odd-symmetric coefficients at the location of an ideal edge. As the one-dimensional edge measure was mainly derived to illustrate the general concept behind this work, a ridge measure will only be rigorously defined for the two-dimensional case in Section~\ref{sec:ridgemeasure}.
\begin{figure}[t!]
	\centering
%	\subfloat[]{\label{fig:genodd}\includegraphics[width=0.3\textwidth]{figs/genodd.png}}\myhspace\subfloat[]{\label{fig:geneven}\includegraphics[width=0.3\textwidth]{figs/geneven.png}}\myhspace\subfloat[]{\label{fig:geneventilde}\includegraphics[width=0.3\textwidth]{figs/geneventilde.png}}\\
	\subfloat[2D edge]{\label{fig:idealedge2d}\includegraphics[width=0.3\textwidth]{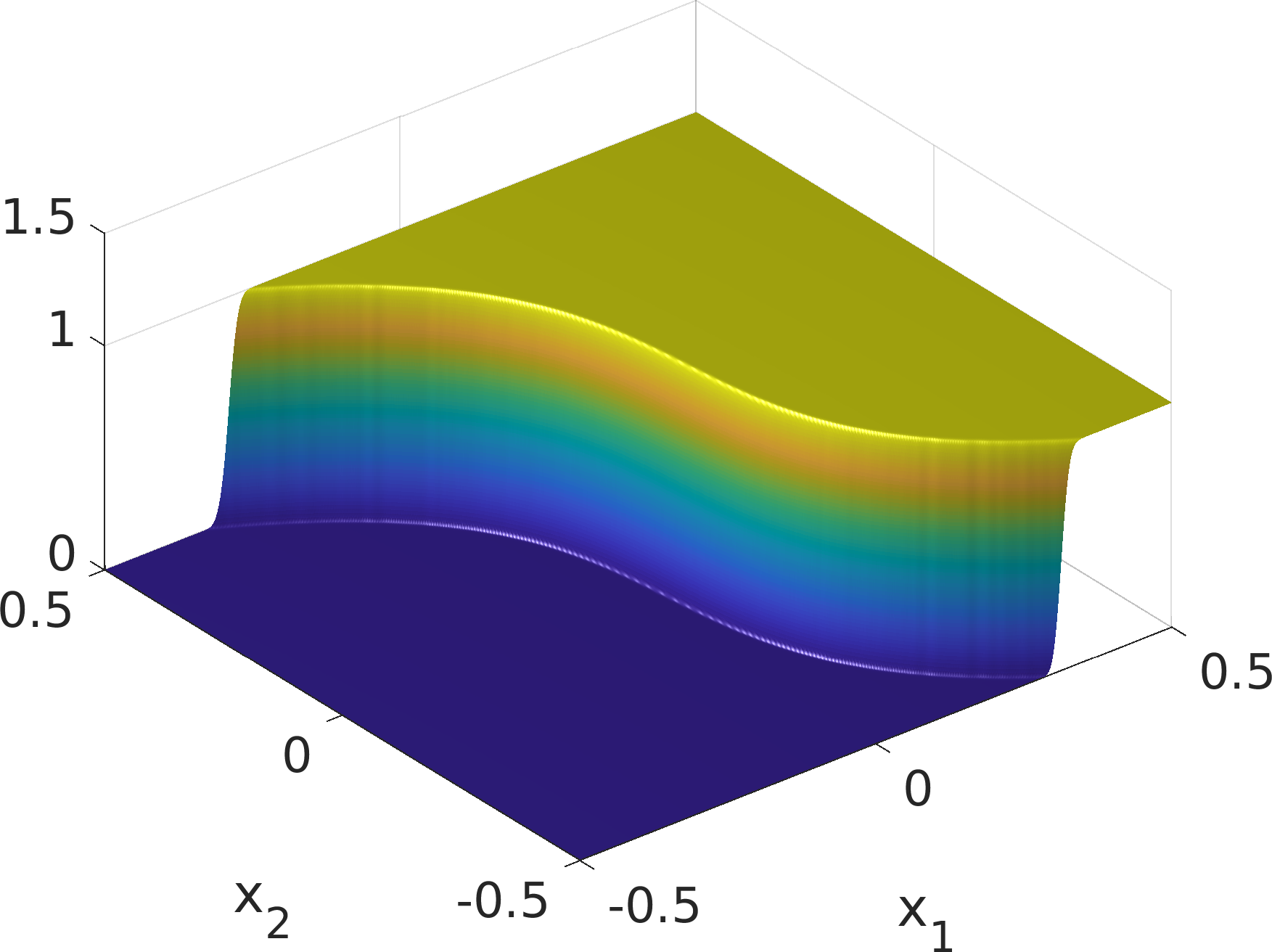}}\myhspace\subfloat[2D ridge]{\label{fig:idealridge2d}\includegraphics[width=0.3\textwidth]{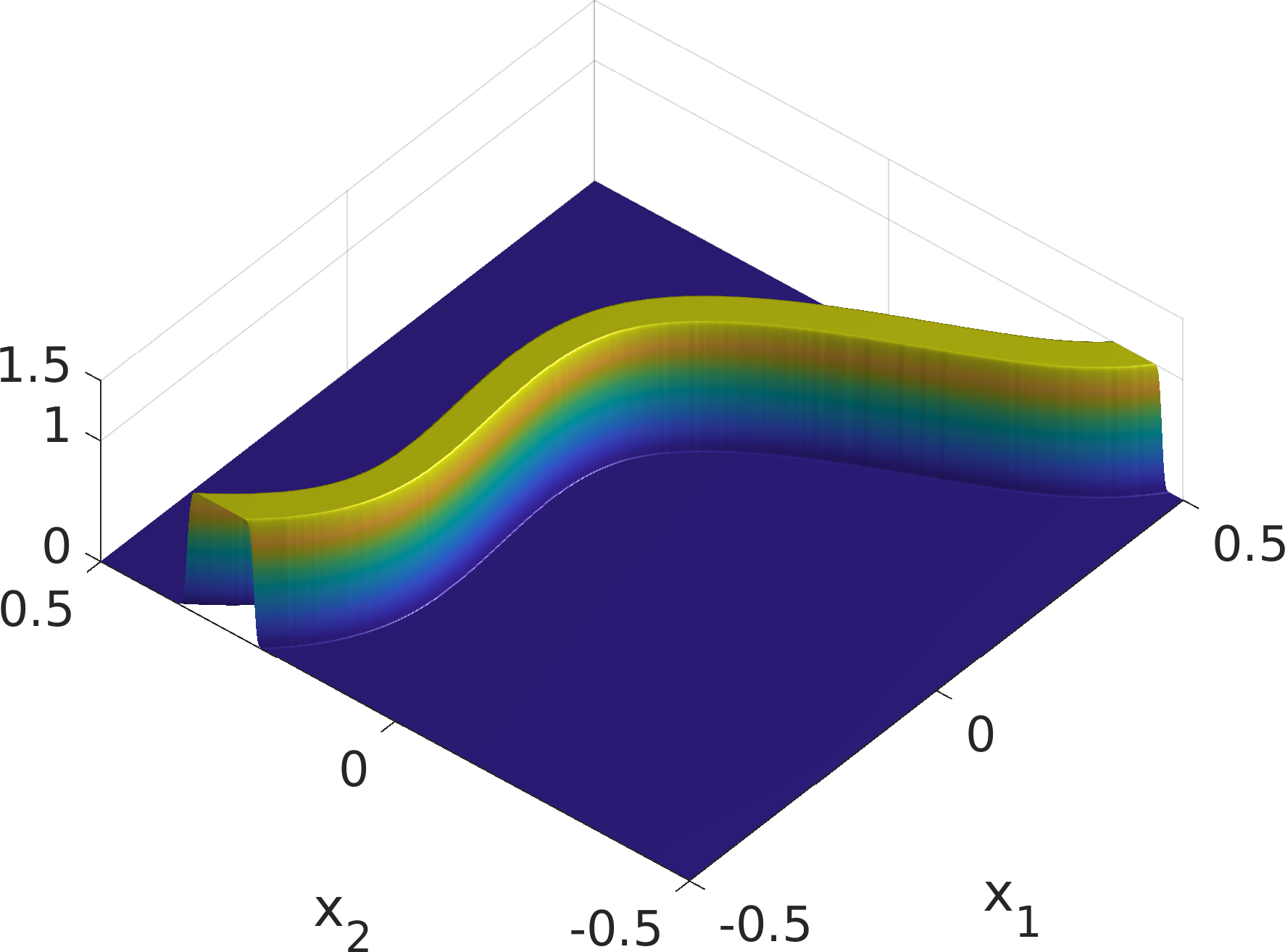}}\myhspace\subfloat[2D blobs]{\label{fig:idealblob2d}\includegraphics[width=0.3\textwidth]{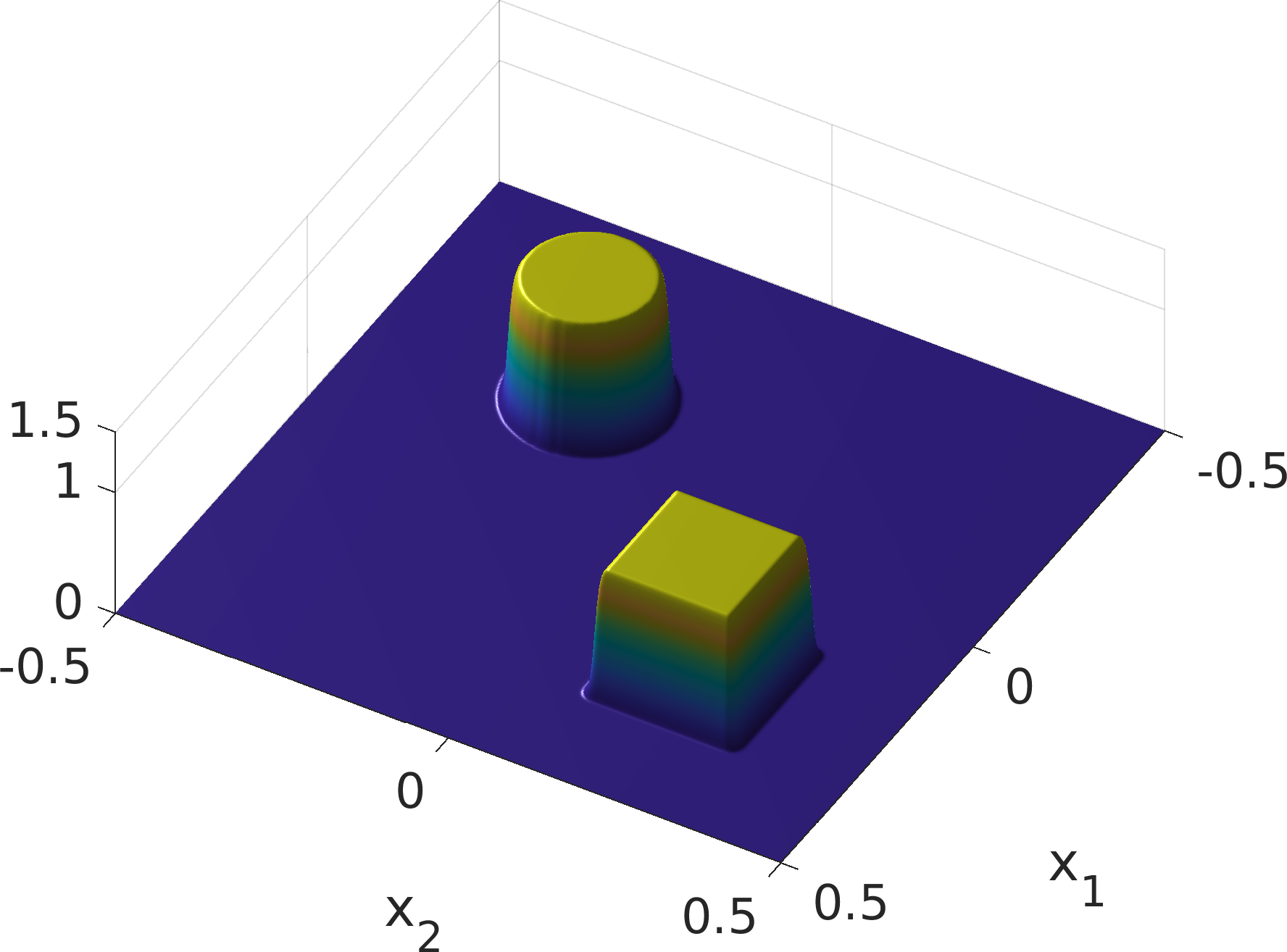}}
	\caption{Examples of ideal features in the two-dimensional setting.}%Examples for $L^1$-normalized two-dimensional symmetric generators. (a): Odd-symmetric generator given by a separable product of $G_1$ and $G_0$. (b): Even-symmetric generator based on $G_2$ and $G_0$. (c): Even-symmetric generator given by the tensor product of $G_2$ with itself. (d): Example of an ideal edge. (e): Example of an ideal ridge. (f): Examples of ideal blobs.}
	\label{fig:2dfeatures}
\end{figure}
\subsection{Symmetric \texorpdfstring{$\alpha$}{Alpha}-Molecules for Edge, Ridge, and Blob Detection in 2D}
We now generalize the one-dimensional edge measure $\EMD$ to edges, ridges, and blobs in the two-dimensional setting (cf. Figure~\ref{fig:2dfeatures}). With small adjustments, the symmetry and self-similarity properties that characterize edges and ridges in the one-dimensional setting can also be utilized to extract features from two-dimensional signals. A function describing a two-dimensional image is locally odd-symmetric at a point lying on an edge with respect to the tangent of the edge contour. On the other hand, the tangents of the contour of a ridge define axes of symmetry associated with even-symmetry while a function is locally point symmetric at the center of a blob. \rev{In order to detect two-dimensional edges and ridges, we will consider generating functions in $L^2(\bR^2)\cap L^1(\bR^2)$ that are defined as tensor products of the Gaussian $G_0$ with one-dimensional odd- or even-symmetric wavelets from the sets $\Psie$ and $\Psio$. This way, we will be able to test for, in a sense, one-dimensional symmetry properties that are present in two-dimensional functions along specific directions. In the case of blob detection, we will consider tensor products of two even-symmetric wavelets.} To this end, we define the sets
\begin{align}
\Psiee &= \left\{c_1c_2\pi^{-\frac{1}{2}}\psie(c_1x_1)G_0(c_2x_2) \colon \psie\in \Psie,  c_1,c_2\in\bR_{>0}\right\}\subset \LTLO,\label{eq:geno}\\
\Psioo &= \left\{c_1c_2\pi^{-\frac{1}{2}}\psio(c_1x_1)G_0(c_2x_2) \colon \psio\in \Psio,  c_1,c_2\in\bR_{>0}\right\}\subset \LTLO\label{eq:gene}
\end{align}
that contain two-dimensional wavelet generators that are based on even- and odd-symmetric one-dimensional wavelets and the Gaussian $G_0$, while the generators in 
\begin{equation}
\Psieet = \left\{c_1c_2\psie(c_1x_1)\psie(c_2x_2) \colon \psie\in \Psie,  c_1,c_2\in\bR_{>0}\right\}\subset \LTLO\label{eq:genetilde}
\end{equation}
are even-symmetric and have vanishing moments in both dimensions. The parameters $c_1$ and $c_2$ can be used to adjust the scale of the even- and odd-symmetric generators as well as their aspect ratio. Examples of two-dimensional symmetric generators contained in \cref{eq:geno,eq:gene,eq:genetilde} are depicted in Figures~\ref{fig:genodd2d}~to~\ref{fig:geneventilde2d}.

Analogous to the one-dimensional case, we require sets of differently scaled analyzing functions to test for scale-invariant local symmetry properties. However, the local symmetry at the location of a two-dimensional feature is typically only given with respect to a specific symmetry axis. Thus, the analysis of two-dimensional signals cannot solely be based on differently scaled wavelet-like functions but also requires a means of changing their preferred orientation. A natural way of obtaining systems of dilations, rotations, and shifts of a single two-dimensional generator is to consider the $\alpha$-molecule framework \cite{grohs2016alpha} briefly described in Section~\ref{sec:alphamolecules}. 
%\begin{defn}
Let $\alpha \in [0,1]$ be fixed, $g\in\Psiee\cup\Psioo\cup\Psieet$ be a symmetric generator, \rev{$a>1$}, $J\subset \bZ$ be a set of scaling parameters, and $\Theta \subset \Tor$ be a set of \rev{orientation} parameters, then the corresponding system of $L^1$-normalized symmetric molecules (noting the $L^1$-isometric dilation) is given by
\begin{equation}
\SM(g,\alpha,a,J,\Theta) = \left\{m_{j,\theta,y}(x) = a^{j(1+\alpha)}g(\ScM{a^j}{\alpha}\RoM{\theta}(x - y))\colon j \in J, \theta \in \Theta, y \in \bR^2 \right\}.
\label{eq:symmetricmoleculesystem}
\end{equation}
If further for all $\abs{\rho} \leq \rev{L}$ with $\rev{L}, M, N_1, N_2 \in \bNz \cup \{\infty\}$
	\begin{equation}
	\abs{\partial^\rho \widehat{g}(\xi)} \lesssim \min\left\{1, a^{-j} + \abs{\xi_1}  + a^{-j(1-\alpha)}\abs{\xi_2}\right\}^M \cdot (1 + \abs{\xi}^2)^{\frac{-N_1}{2}} \cdot (1 + (\xi_2)^2)^{\frac{-N_2}{2}}
	\label{eq:amorder1}
	\end{equation}
holds with the implicit constants independent of the choice of parameters $j\in J$, then we call the associated set $\SM(g,\alpha,a,J,\Theta)$ a system of $L^1$-normalized symmetric $\alpha$-molecules of order $(\rev{L},M,N_1,N_2)$.

The parameter $\alpha$ describes the degree of anisotropy introduced to a system by the scaling matrix $\ScM{a^j}{\alpha}$ and should be chosen as a function of the expected level of noise and the regularity of the structures that are to be detected. If a signal was perturbed with severe noise but the contours of its edges or ridges are smooth, introducing a high degree of anisotropic scaling by choosing $\alpha$ close to zero can significantly improve the detection performance. However, setting $\alpha$ close to zero can lead to problems at locations where the contour of a feature is irregular and can thus not be well approximated by a linear function. \rev{We emphasize} that the elements in \eqref{eq:symmetricmoleculesystem} are $L^1$-normalized, while classical systems of $\alpha$-molecules are normalized with respect to the $L^2$-norm.

The following proposition specifies the order of symmetric molecule systems obtained from generators defined by \cref{eq:gene,eq:geno,eq:genetilde} in terms of a localization parameter $L$, a vanishing moments parameter $M$ and smoothness parameters $N_1$ and $N_2$ (cf. \eqref{eq:defalphamolecule2}).
\begin{thm}
	\label{thm:smorder}
	Let $k\in\bN$ and denote four types of generators by
	\begin{align}
	g_k^{(1)}(x) &= G_k(x_1)G_0(x_2),\quad g_k^{(2)}(x) = \HT G_k(x_1)G_0(x_2), \\
	g_k^{(3)}(x) &= G_k(x_1)G_k(x_2),\quad  g_k^{(4)}(x) = \HT G_k(x_1)\HT G_k(x_2).
	\end{align}
	Let $\alpha\in [0,1]$, \rev{$a>1$}, $J\subset\bZ$ be a set of scaling parameters, and $\Theta \subset [-\frac{\pi}{2},\frac{\pi}{2})$ a set of rotation parameters then it holds that $\SM(\|g_k^{(i)}\|_1^{-1}g_k^{(i)},\alpha,a,J,\Theta)$ is of order \rev{$(L,\max\{0,k - L\},\infty,\infty)$, where $L \in \bNz \cup \{\infty\}$ for $i \in \{1,3\}$, and $L \in \{0,\ldots,k-1\}$ for $i \in \{2,4\}$}.
%	Let $\alpha\in [0,1]$, \rev{$a>1$}, $J\subset\bZ$ be a set of scaling parameters and $\Theta \subset [-\frac{\pi}{2},\frac{\pi}{2})$ a set of rotation parameters then it holds for all $\rho \in \bNz^2$, $M \leq k$ and $N_1,N_2 \in \bNz$ that
%	\begin{equation}
%	\abs{\partial^\rho \widehat{g_k^{(1)}}(\xi)} \lesssim \min\left\{1, a^{-j} + \abs{\xi_1}  + a^{-j(1-\alpha)}\abs{\xi_2}\right\}^M \cdot (1 + \abs{\xi}^2)^{\frac{-N_1}{2}} \cdot (1 + (\xi_2)^2)^{\frac{-N_2}{2}},
%	\label{eq:amorder1}
%	\end{equation}
%	where the implicit constants are independent of the choice of parameters $j\in J$ and we call the associated set $\SM(\|g_k^{(1)}\|_1^{-1}g_k^{(1)},\alpha,a,J,\Theta)$ a system of $L^1$-normalized symmetric $\alpha$-molecules of order $(\infty,k,\infty,\infty)$. Further, with $\alpha$, $a$, $J$ and $\Theta$ as above, $\SM(\|g_k^{(3)}\|_1^{-1}g_k^{(3)},\alpha,a,J,\Theta)$ is also a system of $L^1$-normalized $\alpha$-molecules of order $(\infty,k,\infty,\infty)$, while the systems \linebreak $\SM(\|g_k^{(2)}\|_1^{-1}g_k^{(2)},\alpha,a,J,\Theta)$ and $\SM(\|g_k^{(4)}\|_1^{-1}g_k^{(4)},\alpha,a,J,\Theta)$ are of order $(k-1,k,\infty,\infty)$.
	\begin{proof}
		We need to verify that~\eqref{eq:amorder1} holds. \rev{Let $\rho \in \bNz^2$ and observe that
		\begin{equation}
		\abs{\partial^{\rho_1}_1 (\xi_1)^k e^{-(\xi_1)^2}} = \abs{P(\xi_1)e^{-(\xi_1)^2}} \lesssim \left(\abs{\xi_1}^{k + \rho_1} + \abs{\xi_1}^{\max\{0,k-\rho_1\}}\right) e^{-(\xi_1)^2},
			\end{equation}
			where $P(\xi_1)$ is a polynomial of degree $k + \rho_1$ whose growth is asymptotically bounded by $\abs{\xi_1}^{k + \rho_1}$ and whose decay towards zero is asymptotically bounded by $\abs{\xi_1}^{\max\{0,k-\rho_1\}}$.} Let $N_1,N_2\in\bNz$. We use that $e^{-x^2}\lesssim(1 + \abs{x})^{-N}$ for all $N\in\bNz$ and write
		\begin{align}
		\abs{\partial^\rho \widehat{g_k^{(1)}}(\xi)} &= \abs{\partial^\rho\left((2\pi\iu\xi_1)^k \pi e^{-\pi^2(\xi_1)^2} e^{-\pi^2(\xi_2)^2}\right)}\\
		&=\abs{(2\iu)^k\pi^{k+1}\left(\partial_1^{\rho_1}(\xi_1)^k e^{-\pi^2(\xi_1)^2}\right)\left(\partial_2^{\rho_2}e^{-\pi^2(\xi_2)^2}\right)}\\
		&\lesssim \rev{\left(\abs{\xi_1}^{k+\rho_1} + \abs{\xi_1}^{\max\{0,k-\rho_1\}}\right)}e^{-\pi^2(\xi_1)^2}\cdot\rev{\left(1 + \abs{\xi_2}^{\rho_2}\right)} e^{-\pi^2(\xi_2)^2}\\
		&\lesssim \rev{\left(\abs{\xi_1}^{k+\rho_1} + \abs{\xi_1}^{\max\{0,k-\rho_1\}}\right)\left(1 + \abs{\xi_2}\right)^{\rho_2}}(1+\abs{\xi_1})^{-(N_1+k+\rho_1)}(1 + \abs{\xi_2})^{-(N_1 + N_2 +\rho_2)}\\
		&= \rev{\frac{1 + \abs{\xi_1}^{k+\rho_1 - \max\{0,k-\rho_1\}}}{(1+\abs{\xi_1})^{k + \rho_1 - \max\{0,k-\rho_1\}}}\frac{\abs{\xi_1}^{\max\{0,k-\rho_1\}}}{(1+\abs{\xi_1})^{\max\{0,k-\rho_1\}}}}(1+\abs{\xi_1})^{-N_1}(1+\abs{\xi_2})^{-(N_1+N_2)}\\
		&\leq \min\left\{1,\rev{\abs{\xi_1}^{\max\{0,k-\rho_1\}}}\right\}(1+\abs{\xi_1})^{-N_1}(1+\abs{\xi_2})^{-(N_1+N_2)}\\
		&\lesssim \min\left\{1, a^{-j} + \abs{\xi_1}  + a^{-j(1-\alpha)}\abs{\xi_2}\right\}^{\rev{\max\{0,k-\rho_1\}}}(1 + \abs{\xi}^2)^{\frac{-N_1}{2}} \cdot (1 + (\xi_2)^2)^{\frac{-N_2}{2}},
		\end{align}
		independently of the choice of parameters \rev{$a>1$}, $j\in \bZ$ and $\alpha\in [0,1]$. This shows \eqref{eq:amorder1} and a similar argument leads to the same properties for the generator $g_k^{(3)}$.
		
		For $g_k^{(2)}$, we consider $\abs{\rho} < k$, $N_1,N_2\in\bNz$ and write
		\begin{align}
		\abs{\partial^\rho \widehat{g_k^{(2)}}(\xi)} &= \abs{\partial^\rho\left(-\iu \sgn(\xi_1)(2\pi\iu\xi_1)^k \pi e^{-\pi^2(\xi_1)^2} e^{-\pi^2(\xi_2)^2}\right)}\\
		&=\abs{(2\iu)^k\pi^{k+1}\left(\partial_1^{\rho_1}\abs{\xi_1}(\xi_1)^{k-1} e^{-\pi^2(\xi_1)^2}\right)\left(\partial_2^{\rho_2}e^{-\pi^2(\xi_2)^2}\right)},
		\end{align}
		where $\partial_1^{\rho_1}\abs{\xi_1}(\xi_1)^{k-1} e^{-\pi^2(\xi_1)^2}$ only exists for $\rho_1 < k$. By repeating the same steps as in the case of $\widehat{g_k^{(1)}}$, we can conclude that 
		\begin{equation}
		\abs{\partial^\rho \widehat{g_k^{(2)}}(\xi)}\lesssim \min\left\{1, a^{-j} + \abs{\xi_1}  + a^{-j(1-\alpha)}\abs{\xi_2}\right\}^{\rev{\max\{0,k-\rho_1\}}}(1 + \abs{\xi}^2)^{\frac{-N_1}{2}} \cdot (1 + (\xi_2)^2)^{\frac{-N_2}{2}},
		\end{equation}
		independently of the choice of parameters \rev{$a>1$}, $j\in \bZ$ and $\alpha\in [0,1]$. The same argument can be used to show that $\SM(\|g_k^{(4)}\|_1^{-1}g_k^{(4)},\alpha,a,J,\Theta)$ is of order \rev{$(L,\max\{0,k-L\},\infty,\infty)$ for $L \in \{0,\ldots,k-1\}$}.
	\end{proof}
\end{thm}
\rev{Theorem}~\ref{thm:smorder} illustrates that symmetric molecule systems based on derivatives of the Gaussian are well-localized and smooth in the time domain while the number of vanishing moments increases with the number of derivatives. However, \rev{Theorem}~\ref{thm:smorder} also shows that changing the symmetry properties of a generator by applying the Hilbert transform reduces smoothness in the Fourier domain and thus localization in the time domain.
\subsection{Edge Detection}
\label{sec:edgemeasure}
In the neighborhood of a point $y$ that lies on an ideal edge (cf. Figure~\ref{fig:idealedge2d}), the image is locally odd-symmetric with respect to a symmetry axis defined by the tangent of the edge contour. The measure $\EMD$ can thus be generalized to yield a two-dimensional edge measure by restricting the analysis to symmetric molecules whose direction of vanishing moments is orthogonal to the tangent of the edge contour going through $y$. Let $\geno\in\Psioo$, $\gene\in\Psiee$ be a pair of odd- and even-symmetric generators, $\alpha\in [0,1]$, \rev{$a>1$}, $J = \{j_n\}_{n=1}^{N_J} \subset\bZ$ be an increasing sequence of $N_J \in \bN$ scaling parameters, and $\Theta \subset \bT$ be a set of orientation parameters. For the remainder of this section, we will denote elements from the symmetric molecule system $\SM(\geno,\alpha,a,J,\Theta)$ with $\mo_{j,\theta,y}$ and elements from the system $\SM(\gene,\alpha,a,J,\Theta)$ with $\me_{j,\theta,y}$ (cf. \eqref{eq:symmetricmoleculesystem}). For an image defined by a two-dimensional function $f\in L^2(\bR^2)$, we first denote the most significant scaling and orientation parameters at a point $y\in \bR^2$ by 
\begin{equation}
(j^*(y),\theta^*(y)) = \argmax\limits_{(j,\theta)\in J\times \Theta}\abs{\ip{f}{\mo_{j,\theta,y}}}.
\end{equation}
Then, with a soft-thresholding parameter $\beta > 0$ and a \rev{possibly non-integer} scaling offset parameter $\je\in \bR_{>0}$, a two-dimensional edge measure is given by
\begin{equation}	
\EMT(f,y) = \frac{\abs{\sum\limits_{j\in J}\ip{f}{\mo_{j,\theta^*(y),y}}} - \sum\limits_{j \in J}\abs{\ip{f}{\me_{j-\je,\theta^*(y),y}}} - \beta N_J\Kpsio}{N_J\abs{\ip{f}{\mo_{j^*(y),\theta^*(y),y}}}+\epsilon},
\label{eq:edgemeasure2d1}
\end{equation}
where $\epsilon> 0$ prevents division by zero and $\Kpsio$ (as defined in \eqref{eq:kpsio} \rev{for the one-dimensional wavelet $\psio$ appearing in the definition of $\geno$ \eqref{eq:geno}}) corresponds to the odd-symmetric coefficient at the location of an ideal edge with jump-size one. As in the one-dimensional case, we finally set
\begin{equation}  
\label{eq:edgemeasure2d2}
\EM(f,y)  = \max\{0,\EMT(f,y) \},
\end{equation}
to ensure that the measure maps to $[0,1]$.

As we already pointed out during the derivation of the one-dimensional edge measure $\EMD(f, y)$ earlier in this section (cf.~\eqref{eq:emdt}), the measure~\eqref{eq:edgemeasure2d1} is partly based on the observation that the coefficients obtained from an odd-symmetric and $L^1$-normalized analyzing function are invariant to scaling if it is spatially and directionally aligned with an ideal edge. In this regard, we would like to emphasize an interesting connection to other shearlet-based approaches to edge detection, which utilize the special decay behavior of $L^2$-normalized shearlets. It was shown in~\cite{guo2009characterization} that the coefficents of band-limited shearlets at a point that lies on a smooth boundary curve, and whose local tangent direction is aligned with the considered shearlet, decay with $O(s^{-3/4})$ for $s\to\infty$, where $s$ denotes the scaling parameter (cf. \eqref{eq:defalphamolecule1}). Similar results have also been shown in the case of compactly supported shearlet generators~\cite{kutyniok2017classificationofedges}. This is related to the measures proposed here in the sense that, when considering a two-dimensional generator $\psi \in L^1(\bR^2)\cap L^2(\bR^2)$ and parabolic scaling (i.e., $\alpha = 1/2$), it holds that
\begin{equation}
\label{eq:relationl1l2}
\frac{\normLL{\psi\left(\ScM{s}{1/2}\;\cdot\right)}}{\normL{\psi\left(\ScM{s}{1/2}\;\cdot\right)}} = \frac{s^{-\frac{3}{4}}\normLL{\psi}}{s^{-\frac{3}{2}}\normL{\psi}} = s^{\frac{3}{4}}\frac{\normLL{\psi}}{\normL{\psi}},\quad s>0.
\end{equation}
In other words, considering $L^1$- instead of $L^2$-normalization precisely neutralizes the decay of shearlet coefficients at points that lie on smooth boundary curves.

Further note that, since $\normL{\mo} = 1$ for all $\mo \in \SM(\geno, \alpha, a, J, \Theta)$, the \rev{coefficient associated with the} most significant scale parameter $j^*(y)$ at a point $y$ can directly be related to the contrast (i.e, the height) of an edge going through $y$. By considering $\Kpsio$ (cf. \eqref{eq:kpsio}), which denotes the value of an odd-symmetric coefficient at the location of an ideal edge with jump size one, we define the local height measure for edges as
\begin{equation}
\label{eq:heightmeausreedge}
\HME(f, y) = \frac{\ip{f}{\mo_{j^*(y),\theta^*(y),y}}}{\Kpsio}.
\end{equation}
\subsection{Ridge Detection}
\label{sec:ridgemeasure}
\begin{figure}[t!]
	\centering
	\subfloat[]{\label{fig:ridgeproblem1}\includegraphics[width=0.3\textwidth]{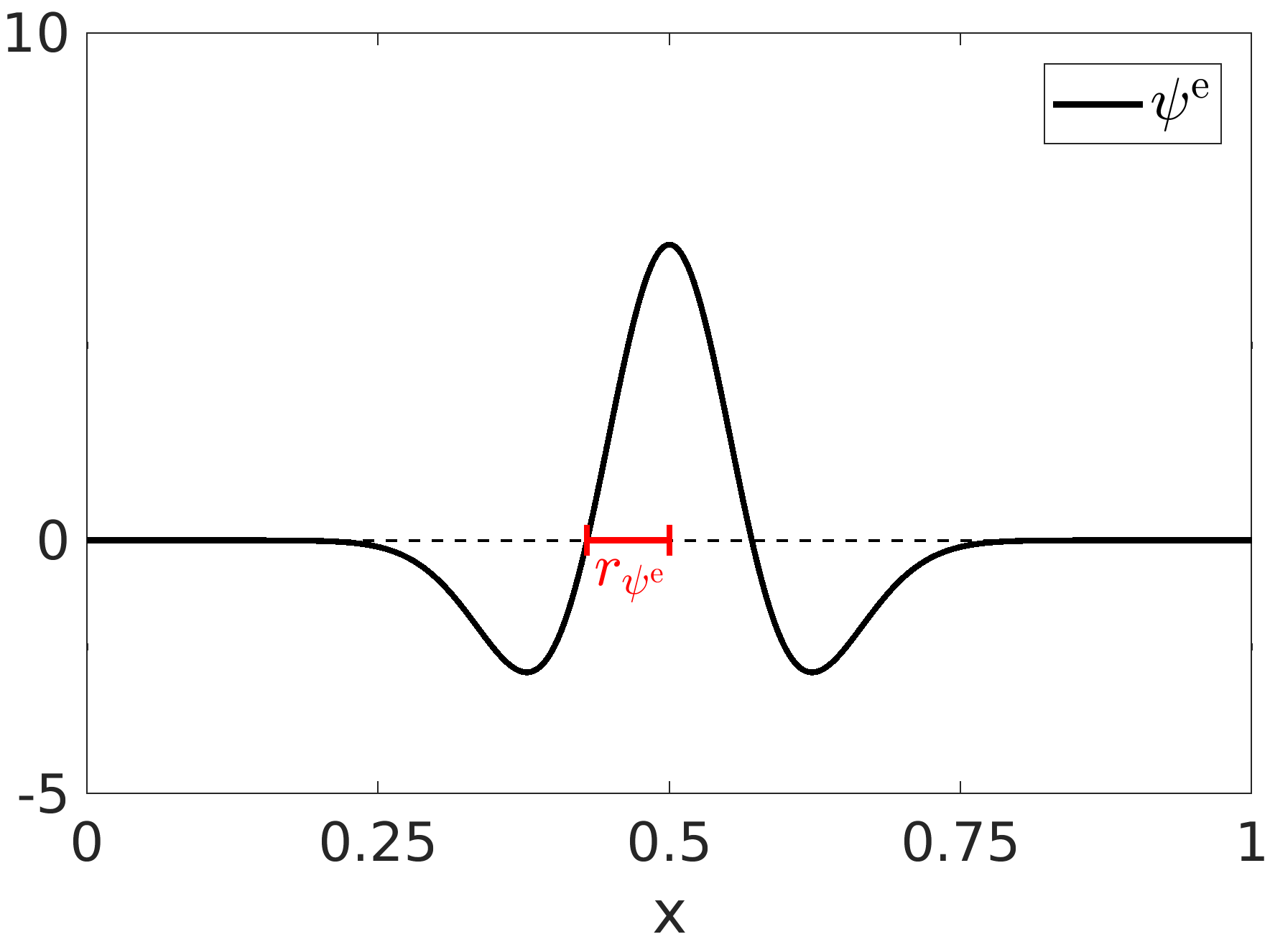}}\myhspace\subfloat[]{\label{fig:ridgeproblem2}\includegraphics[width=0.3\textwidth]{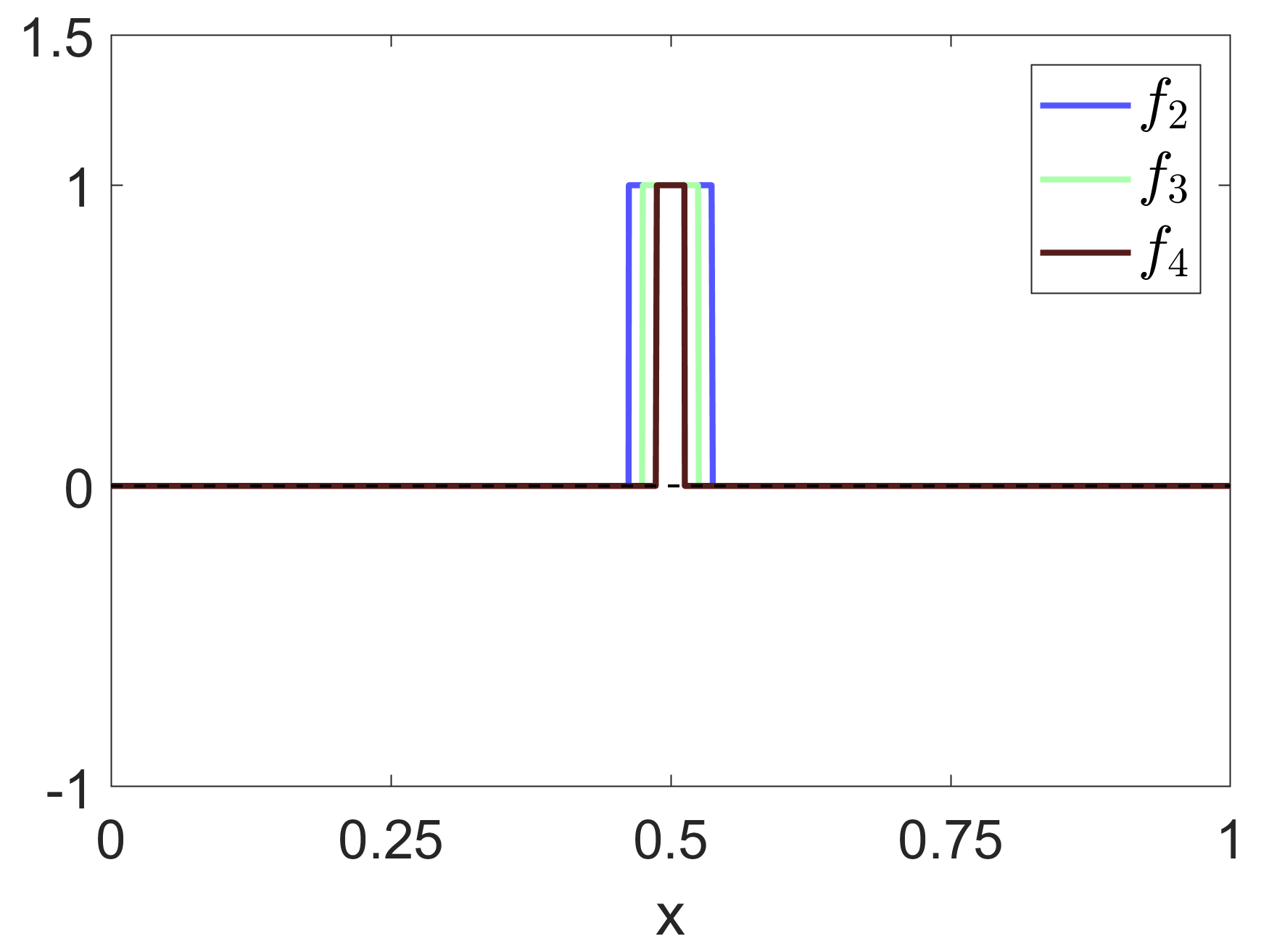}}\myhspace\subfloat[]{\label{fig:ridgeproblem3}\includegraphics[width=0.3\textwidth]{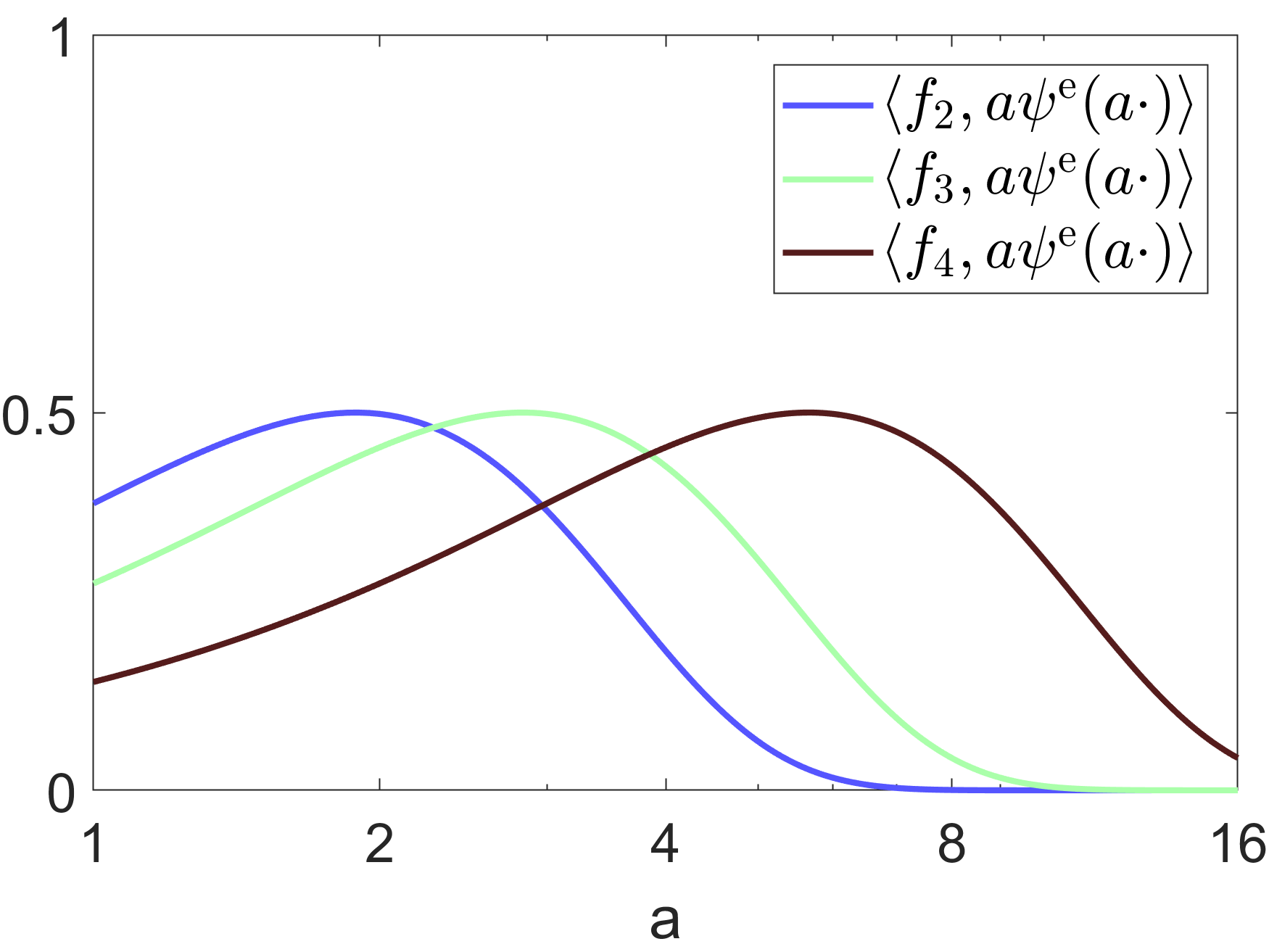}}
	\caption{(a): The radius $r_{\psie}$ of a one-dimensional even-symmetric wavelet, as defined in \eqref{eq:genradius}. (b): Three ideal one-dimensional ridges with different widths. (c): The inner products of $\psie$ with the ideal ridges plotted as a function of the scaling parameter $a$.}
	\label{fig:ridgeproblem}
\end{figure}

For ridge detection, we can exploit the fact that in the neighborhood of a point that lies on the centerline of an ideal ridge (cf. Figure~\ref{fig:idealridge2d}), the image is locally even-symmetric with respect to a symmetry axis defined by the tangent of the centerline. A two-dimensional ridge measure can in principle be obtained by simply interchanging the roles of the odd- and even-symmetric molecules $\mo$ and $\me$ in the definition of the edge measure $\EM(f,y)$. However, to fully retain contrast invariance, it is necessary to also take into account the width of the ridge, that is, the distance between the two jump singularities that define a ridge in an idealized setting. Contrary to the case of edge detection, even-symmetric coefficients of $L^1$-normalized molecules that are centered around the location of a ridge are not invariant to scaling but depend on the scaling parameter as well as the width of the respective ridge. This behavior is illustrated with respect to ridges of three different widths in Figure~\ref{fig:ridgeproblem}.

For an even-symmetric generator  $\gene\in\Psiee$, let us denote the coefficient at the location of an ideal ridge whose tangent direction in $y$ agrees with the orientation of $\gene$ and that has width $2r$ and height $1$ by
\begin{equation}
K_{\gene}(r) = K_{\psie}(r) =  \int_{-r}^r\psie(x)\mathrm{d}x,
\end{equation}
where $\psie$ is the one-dimensional even-symmetric wavelet associated with the two-dimensional generator $\gene$ (cf. \eqref{eq:gene}). We furthermore define the radius of $\gene$ as half of the width of the ideal ridge which yields the most significant coefficient, that is,
\begin{equation}
r_{\gene} = \argmax_{ r \in \bR_{\geq 0}}\abs{K_{\gene}(r)},
\label{eq:genradius}
\end{equation}
and extend this notion to arbitrarily shifted and rotated molecules by setting $r_{\me_{0,\theta,y}} = r_{\me_{0,0,0}} = r_{\gene}$. The radius of a dilated molecule can directly be computed as a function of the scale parameter and the underlying generator, that is, $r_{\me_{j,\theta,y}} = a^{-j}r_{\me_{0,0,0}} =  a^{-j}r_{\gene}$.
Note that in most practical cases, the radius $r_{\gene}$ is equivalent to the distance between the origin and the first zero crossing of the associated one-dimensional wavelet $\psie$ (cf. Figure~\ref{fig:ridgeproblem1}).

Let $\gene\in\Psiee$, $\geno\in\Psioo$ be a pair of odd-symmetric and even-symmetric generators, $\alpha\in [0,1]$, \rev{$a>1$}, $J = \{j_n\}_{n=1}^{N_J} \subset\bZ$ be an increasing sequence of $N_J \in \bN$ scaling parameters, and $\Theta \subset \bT$ be a set of orientation parameters. Furthermore let $f\in L^2(\bR^2)$ be a two-dimensional image. Analogous to the edge detection case, we denote the scaling and rotation parameters associated with the most significant even-symmetric molecule at a point $y\in\bR^2$ in the image plane by
\begin{equation}
\label{eq:mostsignificantridge}
(j^*(y),\theta^*(y)) = \argmax\limits_{(j,\theta)\in J\times \Theta}\abs{\ip{f}{\me_{j,\theta,y}}}.
\end{equation}
For each point $y$, we first estimate the width of an assumed ideal ridge whose centerline passes through $y$ and whose tangent direction at $y$ agrees with the most significant orientation parameter $\theta^*(y)$. We use $j^-(y)$ and $j^+(y)$ to denote the scaling parameters in the strictly increasing sequence $J$ that precede, respectively succeed, the most significant scale $j^*(y)$. The width measure is obtained by computing a \rev{refinement of the most significant scaling parameter as} the maximum point of a parabola fit through the points 
\begin{equation}
\left(\rev{j},\ip{f}{\me_{\rev{j},\theta^*(y),y}}\right), \quad j \in \{ j^-\rev{(y)}, j^\ast\rev{(y)}, j^+\rev{(y)}\}.
\end{equation}
%$\left(a^{-j^-(y)}r_{\gene},\ip{f}{\me_{j^-(y),\theta^*(y),y}}\right)$, $\left(a^{-j^*(y)}r_{\gene},\ip{f}{\me_{j^*(y),\theta^*(y),y}}\right)$, and $\left(a^{-j^+(y)}r_{\gene},\ip{f}{\me_{j^+(y),\theta^*(y),y}}\right)$. 
Following \eqref{eq:maximumestfinal}, we define the local width measure as
\begin{equation}
\WMR(f,y) = 2\left(a^{\frac{c_1}{2c_2}-j^*(y)}r_{\gene}\right),
\label{eq:widthmeasureridge}
\end{equation}
where the values $c_1$ and $c_2$ are chosen according to~\eqref{eq:maximumestparabola}. \rev{If $j^*(y)$ is the first or last index in $J$, we set $\WMR(f,y) = 2\left(a^{-j^*(y)}r_{\gene}\right)$.}
%\eqref{eq:parabolaa} and \eqref{eq:parabolab}. 
Under the assumption that an ideal ridge with width $\WMR(f,y)$, height $1$, and a local tangent direction that agrees with $\theta^*(y)$ passes through $y$, we can directly compute the associated even-symmetric coefficients of differently dilated molecules with rotation parameter $\theta^*(y)$ and denote
\begin{equation}
K_{\gene}(f,j,y) = K_{\gene}\left(a^{j}\WMR(f,y)/2\right) = \int_{-a^{j}\WMR(f,y)/2}^{a^{j}\WMR(f,y)/2}\psie(x)\mathrm{d}x.
\end{equation}
Based on the estimated width $\WMR(f,y)$, we can further define a measure that computes the height of an assumed ideal ridge whose centerline passes through $y$ as the ratio between the most significant coefficient and the value $K_{\gene}(f,j^*(y),y)$, that is,
\begin{equation}
\HMR(f,y) = \frac{\ip{f}{\me_{j^*(y),\theta^*(y),y}}}{K_{\gene}(f,j^*(y),y)}.
\label{eq:heightmeausreridge}
\end{equation}

A two-dimensional ridge measure can now be defined by carefully modifying the edge measure $\EM(f,y)$. To detect locally even-symmetric features instead of odd-symmetric ones, the roles of the even-symmetric molecules $\me$ and odd-symmetric molecules $\mo$ need to be interchanged. While the odd-symmetric coefficients at the location of an ideal edge are invariant to scaling, this is not the case for the even-symmetric coefficients  at the location of an ideal ridge. This can be taken into account by adjusting the normalizing factor for every scaling parameter $j \in J$ using the values $\HMR(f,y) K_{\gene}(f,j,y)$, which correspond to the even-symmetric coefficients at the location of an ideal ridge with width $\WMR(f,y)$ and height $\HMR(f,y)$. Finally, when using strongly oscillating generators, even the sign of the even-symmetric coefficients at the location of an ideal ridge is not invariant to scaling. This can be addressed by considering the expected sign $\sgn(\HMR(f,y) K_{\gene}(f,j,y))$ for each scale parameter when taking the sum of the even-symmetric coefficients. To simplify our notation, we write $\hkcoeff(f,j,y) = \HMR(f,y) K_{\gene}(f,j,y)$ and a two-dimensional ridge measure is now given by
\begin{equation}
\RMT(f,y) = \frac{\sum\limits_{j\in J}\sgn\left(\hkcoeff(f,j,y)\right)\ip{f}{\me_{j,\theta^*(y),y}} - \sum\limits_{j \in J}\abs{\ip{f}{\mo_{j-\jo,\theta^*(y),y}}} - \beta N_J\abs{K_{\gene}(f,j^*(y),y)}}{\sum\limits_{j\in J}\max\left\{\abs{\ip{f}{\me_{j,\theta^*(y),y}}},\abs{\hkcoeff(f,j,y)}\right\} + \epsilon},
\label{eq:ridgemeasure2d1}
\end{equation}
where $\epsilon > 0$ prevents division by zero, $\beta > 0$ is a soft-thresholding parameter, and $\jo\in \bR_{>0}$ denotes the scaling offset between even- and odd-symmetric molecules. As in the case of edge detection, we finally set
\begin{equation}
\RM(f,y)  =\max \{0,\RMT(f,y)\}.
\label{eq:ridgemeasure2d2}
\end{equation}

\subsection{Blob Detection}
\label{sec:blobmeasure}
The measures $\EM(f,y)$ and $\RM(f,y)$ detect features that exhibit symmetry properties that are locally defined by a singular symmetry axis, namely the normal of the tangent of an edge contour or the centerline of a ridge. A similar approach can be used to detect features in two-dimensional images that have less anisotropic symmetry properties in the sense that locally, the center of such features is a point of symmetry for more than one or even all possible directions. Let us consider an idealized blob as a filled square, which has four axes of even symmetry, or a filled circle, for which each line that passes through its center is an axis of even symmetry (cf. Figure~\ref{fig:idealblob2d}).

To detect such blobs, we will use a generator $\genet \in \Psieet$ that is defined by the tensor product of an even-symmetric wavelet with itself (cf. \eqref{eq:genetilde}). Like a square, such a generator has by construction four axes of symmetry and is invariant to rotations \rev{by an angle of} $\frac{\pi}{2}$. Let us denote the coefficient at the location of an ideal square-shaped blob with side length (width) $2r$ and height $1$ by
\begin{equation}
\widetilde{K}_{\genet}(r) =  \int_{-r}^r\int_{-r}^r\psie(x_1)\psie(x_2)\mathrm{d}x_2\mathrm{d}x_1 = K_{\psie}(r)^2,
\label{eq:kpsiet}
\end{equation}
and retain the definition of the radius of $\genet$ as half of the side length of the square-shaped blob which maximizes its response, that is,
\begin{equation}
r_{\genet} = \argmax_{ r \in \rev{\bR_{\geq 0}}} K_{\psie}(r)^2 = \argmax_{ r \in \rev{\bR_{\geq 0}}}\abs{K_{\psie}(r)}.
\label{eq:genradiusblob}
\end{equation}

Let $\geno\in\Psioo$ be an odd-symmetric generator, \rev{$a>1$}, $J = \{j_n\}_{n=1}^{N_J} \subset\bZ$ be an increasing sequence of $N_J \in \bN$ scaling parameters, and $\Theta \subset \bT$ be a set of orientation parameters. Due to the isotropic nature of the features we aim to detect, and contrary to the cases of edge and ridge detection, we set $\alpha = 1$ and denote elements from the molecule system $\SM(\genet,1,a,J,\Theta)$ with $\met_{j,\theta,y}$ and elements from the system $\SM(\geno,1,a,J,\Theta)$ with $\mo_{j,\theta,y}$. For a two-dimensional image $f\in L^2(\bR^2)$, we again denote the most significant pair of scaling and rotation parameters with respect to the even-symmetric generator $\genet$ at a point $y\in\bR^2$ with
\begin{equation}
(j^*,\theta^*) = \argmax\limits_{(j,\theta)\in J\times \Theta}\abs{\ip{f}{\met_{j,\theta,y}}}.
\end{equation}
As in the case of ridge detection, the even-symmetric coefficients at the center of a blob are not invariant to scaling but depend on both the scaling parameter $j$ and the width of the respective blob. For each point $y$, we again estimate the width of an assumed ideal blob by fitting a parabola through the points
\begin{equation}
\rev{\left(j,\ip{f}{\met_{\rev{j},\theta^*(y),y}}\right), \quad j \in \{ j^-(y), j^\ast(y), j^+(y)\},}
\end{equation}
where $j^-(y)$ denotes the predecessor and $j^+(y)$ the successor of $j^*(y)$ in the sequence $J$ and compute
\begin{equation}
\WMB(f,y) = 2\left(a^{\frac{c_1}{2c_2}-j^*(y)}r_{\genet}\right),
\label{eq:widthmeasureblob}
\end{equation}
where the values $c_1$ and $c_2$ are chosen according to~\eqref{eq:maximumestparabola}. 
%\eqref{eq:parabolaa} and \eqref{eq:parabolab}. 
Under the assumption that an ideal square-shaped blob with width $\WMB(f,y)$ and height $1$ is centered on $y$, we can also directly compute the associated even-symmetric coefficients of differently dilated molecules and denote
\begin{equation}
\widetilde{K}_{\genet}(f,j,y) = \widetilde{K}_{\genet}\left(a^{j}\WMB(f,y)/2\right) = K_{\psie}\left(a^{j}\WMB(f,y)/2\right)^2.
\end{equation}
The height of the blob can then be computed by
\begin{equation}
\HMB(f,y) = \frac{\ip{f}{\met_{j^*(y),\theta^*(y),y}}}{\widetilde{K}_{\genet}(f,j^*(y),y)}.
\label{eq:heightmeausreblob}
\end{equation}
We are now ready to define a measure for blob detection that can be understood as a generalization of the ridge measure $\RM(f,y)$ in the sense that we aim to detect features that are associated with more than one axis of even symmetry. In particular, we will take into account that axes of even symmetry induce rotation invariance. A perfect square, for instance, has four symmetry axes and is invariant to rotations by $\frac{\pi}{2}$, while a pentagon has five axes of even symmetry and is invariant to rotations by $\frac{2\pi}{5}$. Let for each point $\rev{y \in \bR^2}$ in the image plane $\Thetat(y) \subset \Theta$ denote the subset of rotation parameters that correspond to the expected rotation invariance properties of the blob centered on $y$. The subset $\Thetat(y)$ thus depends on the most significant rotation parameter $\theta^*(y)$ and on the symmetry and rotation invariance properties of the specific shape we aim to detect. Note that in the case of circles, $\Thetat(y) = \Theta$ independently of  the most significant scale paramter $j^*(y)$ for all points $y$. With a soft-thresholding parameter $\beta > 0$ and a \rev{possibly non-integer} scaling offset $\jo\in \bR_{>0}$, the blob measure is given by
\begin{equation}	
\resizebox{.9\hsize}{!}{$\BMT(f,y) = \frac{\sum\limits_{j\in J}\min\limits_{\theta\in\Thetat(y)}\left\{\sgn\left(\HMB(f,y)\right)\ip{f}{\met_{j,\theta,y}}\right\} - \sum\limits_{j \in J}\max\limits_{\theta\in\Thetat(y)}\left\{\abs{\ip{f}{\mo_{j-\jo,\theta,y}}}\right\} - \beta N_J \widetilde{K}_{\genet}(f,j^*(y),y)}{\sum\limits_{j\in J}\max\left\{\abs{\ip{f}{\met_{j,\theta^*(y),y}}},\abs{\HMB(f,y)} \widetilde{K}_{\genet}(f,j,y)\right\}},$}
\label{eq:blobmeasure2d1}
\end{equation}
where $\epsilon > 0$ prevents division by zero. We define
\begin{equation}	
\BM(f,y)  = \max\{0,\BMT(f,y)\}.
\label{eq:blobmeasure2d2}
\end{equation}
\subsection{Estimation of Local Tangent Directions for Edges and Ridges}
\label{sec:orientationmeasure}

The width estimates $\WMR(f, y)$ and $\WMB(f, y)$ exploit the fact that even-symmetric generators from the sets $\Psiee$ and $\Psieet$ are themselves associated with a width \rev{which, in combination with the scale parameter $j$, also determines the width of the elements in the associated system of symmetric molecules}. By considering the orientation parameter $\theta$, a similar approach can be used to estimate local tangent directions of anisotropic features such as edges or ridges. Let us first consider the case of edges and a system $\SM(\geno, \alpha, a, J, \Theta)$ of odd-symmetric molecules with $\geno \in \Psioo$, $\alpha\in [0,1]$, \rev{$a>1$}, an increasing sequence $J = \{j_n\}_{n=1}^{N_J} \subset\bZ$ of $N_J \in \bN$ scaling parameters, and a clockwise (or counter-clockwise) ordered sequence $\Theta = \{\theta_n\}_{n=1}^{N_{\Theta}}  \subset \bT$ of $N_\Theta\in \bN$ orientation parameters on the torus. Each odd-symmetric molecule $\mo_{j,\theta,y} \in \SM(\geno, \alpha, a, J, \Theta)$ (cf.\ \eqref{eq:symmetricmoleculesystem}) has a preferred orientation defined by the rotation parameter $\theta$ and the preferred orientation of the generator $\geno$. At the location of an ideal edge, the odd-symmetric coefficient is maximized with respect to rotation if the direction of the one-dimensional wavelet $\psio$ used in the construction of the generator $\geno$ is orthogonal to the local tangent of the edge. For an image $f\in L^2(\bR^2)$ and a point $y\in\bR^2$, let us again denote the most significant scaling and orientation parameters \rev{by}
\begin{equation}
\label{eq:mostsignificantorientationmeasure}
(j^*(y),\theta^*(y)) = \argmax\limits_{(j,\theta)\in J\times \Theta}\abs{\ip{f}{\mo_{j,\theta,y}}}.
\end{equation}
Analogous to the width measures $\WMR(f, y)$ and $\WMB(f, y)$, we can use $\theta^*(y)$ as a first approximation of the tangent direction of a potential edge going through the point $y$. Again, this approximation can be refined by computing the maximum point of a parabola fit through the points $\left(\theta^{-}(y),\ip{f}{\me_{j^*(y),\theta^{-}(y),y}}\right)$, $\left(\theta^{*}(y),\ip{f}{\me_{j^*(y),\theta^{*}(y),y}}\right)$, and $\left(\theta^{+}(y),\ip{f}{\me_{j^{*}(y),\theta^{+}(y),y}}\right)$, where $\theta^{-}(y)$ denotes the preceding and $\theta^+(y)$ the succeeding orientation parameter of $\theta^*(y)$ in the sequence $\Theta$. The local tangent orientation can then be estimated by applying \eqref{eq:maximumestfinal}, that is,
\begin{equation}
\OME(f, y) = \theta^{*}(y) - \frac{c_1}{2c_2},
\label{eq:orientationmeasureedge}
\end{equation}
where the values $c_1$ and $c_2$ are chosen as in~\eqref{eq:maximumestparabola}. 
%\eqref{eq:parabolaa} and \eqref{eq:parabolab}. 
To obtain a function $\OMR(f, y)$ that estimates local tangent orientation of ridges, it suffices to replace the odd-symmetric generator $\geno$ with an even-symmetric generator $\gene\in\Psiee$.
\section{Performance Evaluation on Synthetic Images}
\label{sec:syntheval}
To evaluate the performance of the proposed measures, we first consider sets of synthetic digital grayscale images for which the locations and properties of features are unambiguously defined. For each type of feature, we created two test images of varying complexity that contain exemplary instances of the respective feature. To test the stability of the proposed measures in adverse conditions, we furthermore applied a combination of Gaussian and Poisson (shot) noise to each test image with three different degrees of severity (no noise, medium noise, severe noise). The Supplemental Materials contain visualizations of all six synthetic images, their distorted versions as well as the corresponding ground truths in Section~\ref{sec:evalsyn_supp} and an explanation of the \matlab{} variables used in Section~\ref{sec:digitalimplementation}.

In all cases, both the synthetic image and the ground truth are based on a geometry that was defined semi-randomly on the unit square and then discretized on a grid of $768\times 768$ pixels. Hence, the ground truth is not directly tied to the discrete digital image and therefore less likely to be biased towards any of the different algorithms we consider in our numerical experiments. For all three features, the feature location ground truth is stored as a binary image where white pixels denote the contours of edges, centerlines of ridges, and centers of blobs, respectively. In the case of edges and ridges, the ground truth also contains the orientation of the tangent for each pixel that lies on the contour of an edge or the centerline of a ridge. The same holds for local feature widths in the case of ridges and blobs. \rev{The height measures $\HMR(f,y)$, and $\HMB(f,y)$ are essentially defined as functions of the width measures $\WMR(f,y)$, and $\WMB(f,y)$ (cf. \eqref{eq:heightmeausreridge}, respectively \eqref{eq:heightmeausreblob}). The validity of a height measure therefore almost exclusively depends on the precision of the associated width measure. For brevity, the measures $\HMR(f,y)$, and $\HMB(f,y)$ were thus omitted from the numerical evaluation.}
%%
%\begin{figure}[t!]
%	\centering
%	\subfloat[Input (\synthone, medium noise)]{\label{fig:synthimages_edges_SFD_input}\spyon{3.2}{figs/01_edgeEvalSimple_gauss_30_poiss_5_input.pdf}{4cm}{(0.67,0.55)}{(-0.3,0.3)}{2.75cm}}
%	\subfloat[SymFD edge measure $\EM(f,y)$]{\label{fig:synthimages_edges_SFD_fm}\spyon{3.2}{figs/04_edgeEvalSimple_gauss_30_poiss_5_SFD_featureMap.pdf}{4cm}{(0.69,0.55)}{(-0.3,0.3)}{2.75cm}}\\
%	\subfloat[Tangent directions $\OME(f,y)$ after post-processing]{\label{fig:synthimages_edges_SFD_thinned}\spyon{3.2}{figs/10_edgeEvalSimple_gauss_30_poiss_5_SFD_orientationMap_thinned.pdf}{4cm}{(0.69,0.55)}{(-0.3,0.3)}{2.75cm}}
%	\subfloat[Error map for $\OME(f,y)$]{\label{fig:synthimages_edges_SFD_orierror}\spyon{3.2}{figs/11_edgeEvalSimple_gauss_30_poiss_5_SFD_errorMap_orientation.pdf}{4cm}{(0.68,0.55)}{(-0.3,0.3)}{2.75cm}}
%	\label{fig:synthimages_edges_SFD}
%	\caption{Detection of edges and local tangent orientations in a synthetic image yielded by SymFD with parameters $\psio = \frac{\HT G_2}{\normL{\HT G_2}}$, $\psie = \frac{G_2}{\normL{G_2}}$, $\SFDparammath{maxFeatureWidth} = 16$, $\SFDparammath{maxFeatureLength} = 16$, $\alpha = \frac{1}{2}$, $\SFDparammath{minFeatureWidth} = 4$, $\SFDparammath{scalesPerOctave} = 2$, $\SFDparammath{nOrientations} = 16$, $\je = 1$, and $\beta = 15$. The input image is of size $768\times 768$ and was distorted by a combination of Gaussian and Poisson noise.}
%\end{figure}
		
We consider different metrics to evaluate the results obtained by a feature detection algorithm with respect to the corresponding ground truth. For a given image, let $\Ngt$ denote the number of pixels that are marked as feature points in the ground truth and $\Ndet$ the number of points that were detected as feature points by the algorithm under evaluation. We denote the set of feature points in the ground truth with $\Pgt = \{p_n\}_{n = 1}^{\Ngt}\subset \bN^2$ and the set of detected feature points with $\Pdet = \{\rev{q}_n\}_{n = 1}^{\Ndet} \subset \bN^2$. Furthermore, let $\Ddet = \{ \min_{\rev{p} \in \Pgt}\|\rev{q - p}\| \colon \rev{q} \in \Pdet \}$ denote the set of all Euclidean distances from pixel coordinates in $\Pdet$ to the set $\Pgt$, and $\Dgt = \{ \min_{\rev{q} \in \Pdet}\|\rev{p - q}\| \colon \rev{p} \in \Pgt \}$ denote the set of all distances from points in $\Pgt$ to the set $\Pdet$. A figure of merit that evaluates how successful an algorithm was at correctly detecting the location of features can then be defined by
\begin{equation}
\FOM = \frac{1}{\max\{\Ngt,\Ndet\}}\sum_{n = 1}^{\Ngt}\frac{1}{1 + \gamma {d_n}^2},
\label{eq:FOM}
\end{equation}
where $d_n \in D = \Dgt \cup \Ddet$ denotes the $n$-th largest element in the union of both sets of distances, and $\gamma > 0$. All $\FOM$-values reported in this section were computed with $\gamma = \frac{1}{9}$. Note that for $D = \Ddet$ and accordingly adjusted summation bounds (i.e., replacing $\Ngt$ with $\Ndet$), \eqref{eq:FOM} yields exactly the definition of Pratt's figure of merit ($\PFOM$) \cite{AbPr79}. The main reason for using a slightly adjusted version here is that the original definition of $\PFOM$ tends to be too forgiving of false positives in the case that $\Ndet < \Ngt$.

For a fixed radius $r > 0$, we define the set of successfully detected points in the ground truth (true positives) as $\TPset = \{ \rev{p} \in \Pgt \colon \min_{\rev{q} \in \Pdet}\|\rev{p - q}\| \leq r \}$ and let $\Ntp \in \bN$ denote the number of points in $\TPset$. To evaluate width and tangent orientation measures, we compute the mean of the set of absolute errors obtained by comparing the corresponding ground truth values at points in $\TPset$ with the values yielded by an algorithm at the nearest neighbors of the ground truth points in the set $\Pdet$. Formally, we define the mean absolute error ($\MAE$) of an arbitrary measure $\Meas(f, {y})$ computed on a digital image $f$ as
\begin{equation}
\MAEall = \frac{1}{\Ntp} \sum\limits_{n = 1}^{\Ntp}\abs{\MeasGt(f,p_n) - \Meas\left(f,\argmin\limits_{\rev{q}\in\Pdet}\|p_n  - \rev{q}\|\right)},
\label{eq:MAE}
\end{equation}
where $p_n \in \TPset$ and $\MeasGt(f,p_n)$ denotes the ground truth of the estimated property at the pixel $p_n$. Note that in the case of tangent orientation measures, the error is computed on the torus. Finally, we define the number of true positives ($\TP$) and the ratio of successfully detected feature points ($\TPR$) as
\begin{equation}
\TP = \Ntp,\text{ and }\; \TPR = \frac{\Ntp}{\Ngt},
\label{eq:tpfptpr}
\end{equation}
respectively.

Throughout this section, the results of SymFD \rev{(the proposed algorithm)} are not only compared to the ground truth but also to the results of other well-known feature detection algorithms. An overview of all considered algorithms is given in Table~\ref{tab:synthalgorithms} in the Supplementary Materials, which lists for each method the types of features it can detect, the number of parameters of the algorithm and the average time it took to process one synthetic digital image. The average processing time was measured on an Intel Core i7-4790 CPU clocked at 3.60 GHz.

The parameters of each method were separately optimized for each test image and each noise configuration. In the cases of edge and ridge detection, the optimization was carried out in terms of the $\FOM$-value. For blob detection, the goal was to maximize the number of correctly detected blobs while keeping the number of false positives close to zero.

\rev{Note that SymFD also implements two post-processing steps, namely hard thresholding and morphological thinning to obtain thinned binary feature maps. These steps, along with the overall implementation details of SymFD, are explained in detail in Section~\ref{sec:digitalimplementation} in the Supplementary Materials.}

All \matlab{} scripts and configurations that were used to obtain the numerical results presented in this section can be downloaded from \url{http://www.math.uni-bremen.de/cda/software.html}.
\subsection{Edge Detection}
The synthetic images for the evaluation the edge measure $\EM(f,y)$ and the tangent orientation measure $\OME(f,y)$ are based on shapes that are defined by closed smooth spline curves and polygons. The main advantage of considering splines and polygons is that in both cases, with the exception of corner points in a polygon, the tangent can easily be computed for every point on the respective curve. The binary images that represent the feature location ground truth were obtained by marking each pixel which intersects with any of the defined curves. Each polygon and each spline curve is constructed from a finite sequence of points. These point sequences were predefined in the case of \synthone{} and semi-randomly selected in \synthtwo{}. The shapes in both images were filled using a range of constant values and different types of gradients to ensure a certain amount of contrast variation. Furthermore, both images were overlaid with a smooth surface to test the capability of an algorithm to properly distinguish between sharp and smooth transitions. The synthetic images, their distorted versions, and the corresponding ground truths are compiled in Figure~\ref{fig:synthimages_edges} in the Supplementary Materials.
\begin{table}[t!]
\centering
\caption{Edge Detection Performance on Synthetic Images}
\label{tab:synthimages_edges_comparison}
\begin{scriptsize}
	\begin{threeparttable}
		{\setlength{\tabcolsep}{4pt}\renewcommand{\arraystretch}{1.3}\begin{tabular}{lrrrcrrrcrrr}
				\toprule[0.5mm]
				&\multicolumn{3}{c}{No Noise} & \phantom{abc}&\multicolumn{3}{c}{Medium Noise} & \phantom{abc}&\multicolumn{3}{c}{Severe Noise}\\
				&$\FOM$ & $\MAEOME$ & $\TPR$ & &$\FOM$ & $\MAEOME$ & $\TPR$ & &$\FOM$ & $\MAEOME$ & $\TPR$\\
				\midrule
				\synthone{}\\
				SymFD & 0.91 & 1.79\,$^\circ$ & 100\,\%& & 0.89 & 3.19\,$^\circ$ & 100\,\%& & 0.72 & 4.46\,$^\circ$ & 83\,\%\\
				Shearlet \cite{YLEK2009} & 0.85 & 3.78\,$^\circ$ & 100\,\%& & 0.82 & 5.78\,$^\circ$ & 98\,\%& & 0.64 & 8.71\,$^\circ$ & 84\,\%\\
				PhaseCong \cite{Kov1999} & 0.91 & 6.13\,$^\circ$ & 100\,\%& & 0.85 & 8.48\,$^\circ$ & 95\,\%& & 0.58 & 15.59\,$^\circ$ & 72\,\%\\
				Canny \cite{Can1986} & 0.92 & \multicolumn{1}{c}{n/a} & 100\,\%& & 0.90 & \multicolumn{1}{c}{n/a} & 100\,\%& & 0.72 & \multicolumn{1}{c}{n/a} & 86\,\%\\
				Sobel \cite{SF68} & 0.93 & \multicolumn{1}{c}{n/a} & 100\,\%& & 0.73 & \multicolumn{1}{c}{n/a} & 83\,\%& & 0.11 & \multicolumn{1}{c}{n/a} & 17\,\%\\
				\synthtwo{}\\
				SymFD & 0.90 & 4.55\,$^\circ$ & 100\,\%& & 0.85 & 7.43\,$^\circ$ & 96\,\%& & 0.64 & 13.28\,$^\circ$ & 76\,\%\\
				Shearlet \cite{YLEK2009} & 0.85 & 6.58\,$^\circ$ & 100\,\%& & 0.79 & 9.67\,$^\circ$ & 96\,\%& & 0.55 & 13.32\,$^\circ$ & 73\,\%\\
				PhaseCong \cite{Kov1999} & 0.87 & 9.46\,$^\circ$ & 97\,\%& & 0.72 & 12.11\,$^\circ$ & 82\,\%& & 0.42 & 17.01\,$^\circ$ & 52\,\%\\
				Canny \cite{Can1986} & 0.91 & \multicolumn{1}{c}{n/a} & 100\,\%& & 0.87 & \multicolumn{1}{c}{n/a} & 99\,\%& & 0.65 & \multicolumn{1}{c}{n/a} & 79\,\%\\
				Sobel \cite{SF68} & 0.93 & \multicolumn{1}{c}{n/a} & 100\,\%& & 0.65 & \multicolumn{1}{c}{n/a} & 76\,\%& & 0.16 & \multicolumn{1}{c}{n/a} & 22\,\%\\\midrule[0.5mm]
			\end{tabular}}
			\begin{tablenotes}\footnotesize
				\item $\FOM$: figure of merit, $\MAEOME$: mean \rev{absolute} error of the respective tangent direction measure, $\TPR$: true positive rate. Note that $\MAEOME$ is computed only with respect to points in the ground truth that are sufficiently close to points detected by an algorithm. The ratio of considered points is indicated by $\TPR$.
			\end{tablenotes}
		\end{threeparttable}
	\end{scriptsize}
\end{table}

\begin{figure}[t!]
	\centering
	\subfloat[Input (\synthtwo{}, severe noise)]{\label{fig:synthimages_edges_comparison_input}\spyon{3.2}{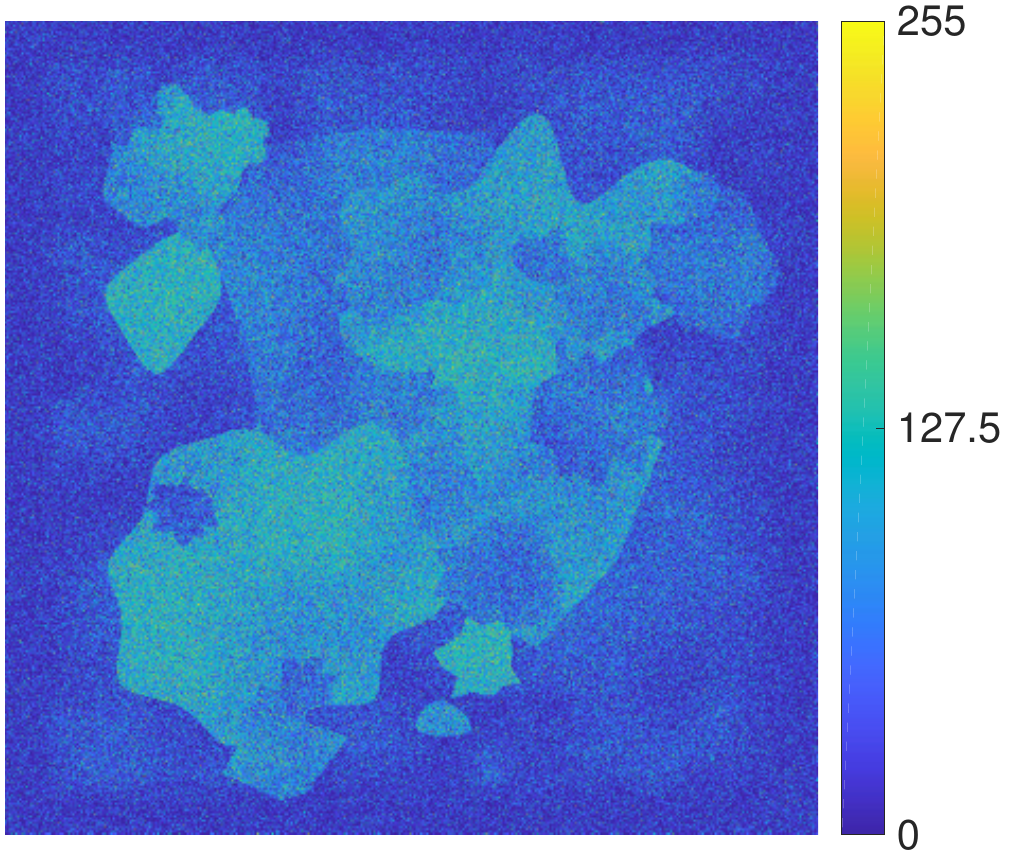}{4cm}{(0.2,0.43)}{(-0.3,0.3)}{2.75cm}}
	\subfloat[SymFD]{\label{fig:synthimages_edges_comparison_SFD}\spyon{3.2}{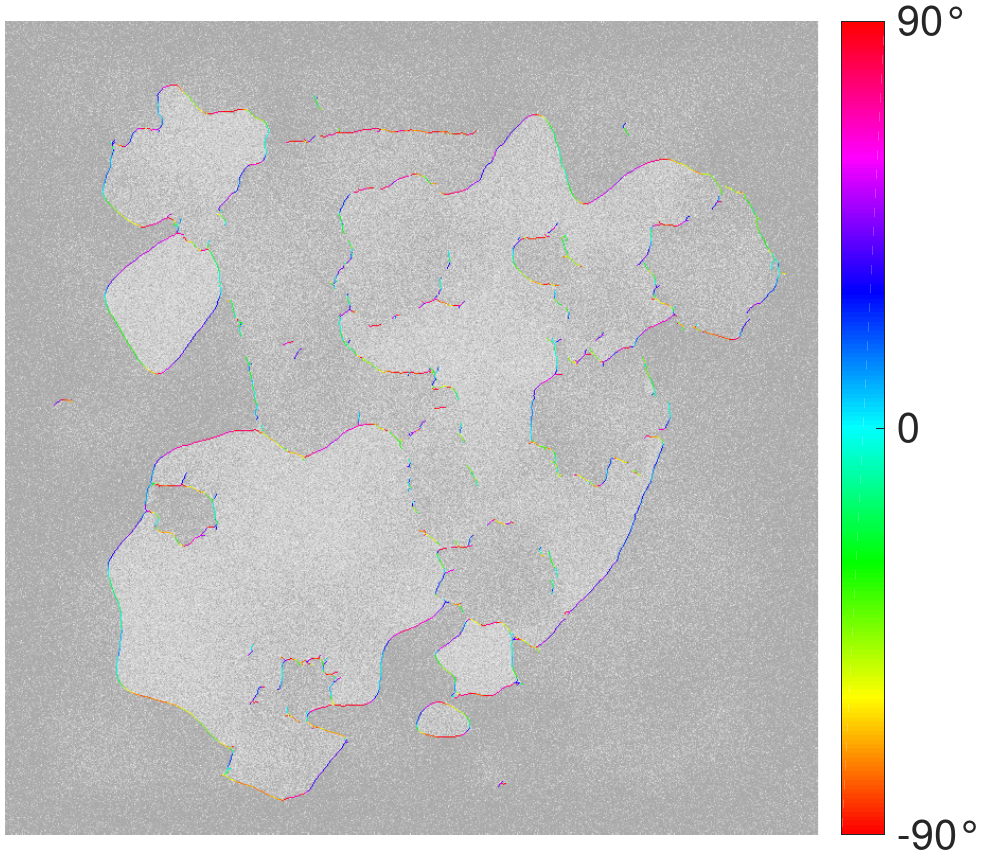}{4cm}{(0.2,0.43)}{(-0.3,0.3)}{2.75cm}}\\
	\subfloat[Shearlet-based edge detector \cite{YLEK2009}]{\label{fig:synthimages_edges_comparison_shearlet}\spyon{3.2}{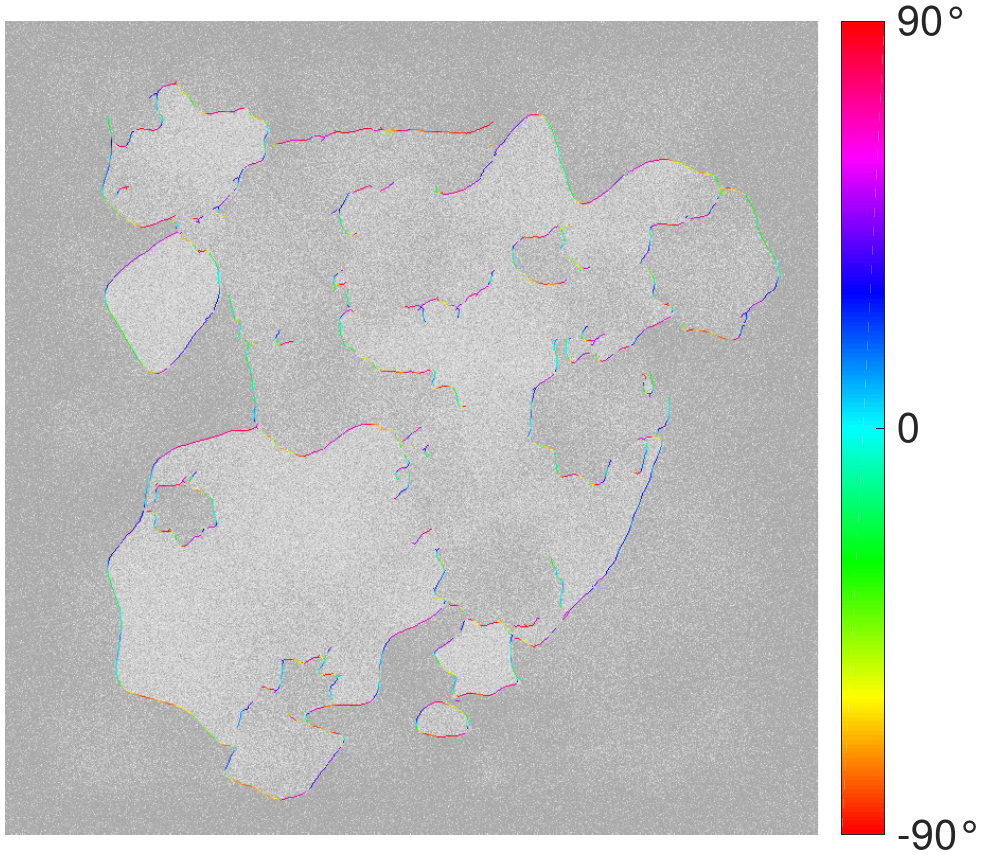}{4cm}{(0.2,0.43)}{(-0.3,0.3)}{2.75cm}}
	\subfloat[Phase congruency measure \cite{Kov1999}]{\label{fig:synthimages_edges_comparison_phasecong}\spyon{3.2}{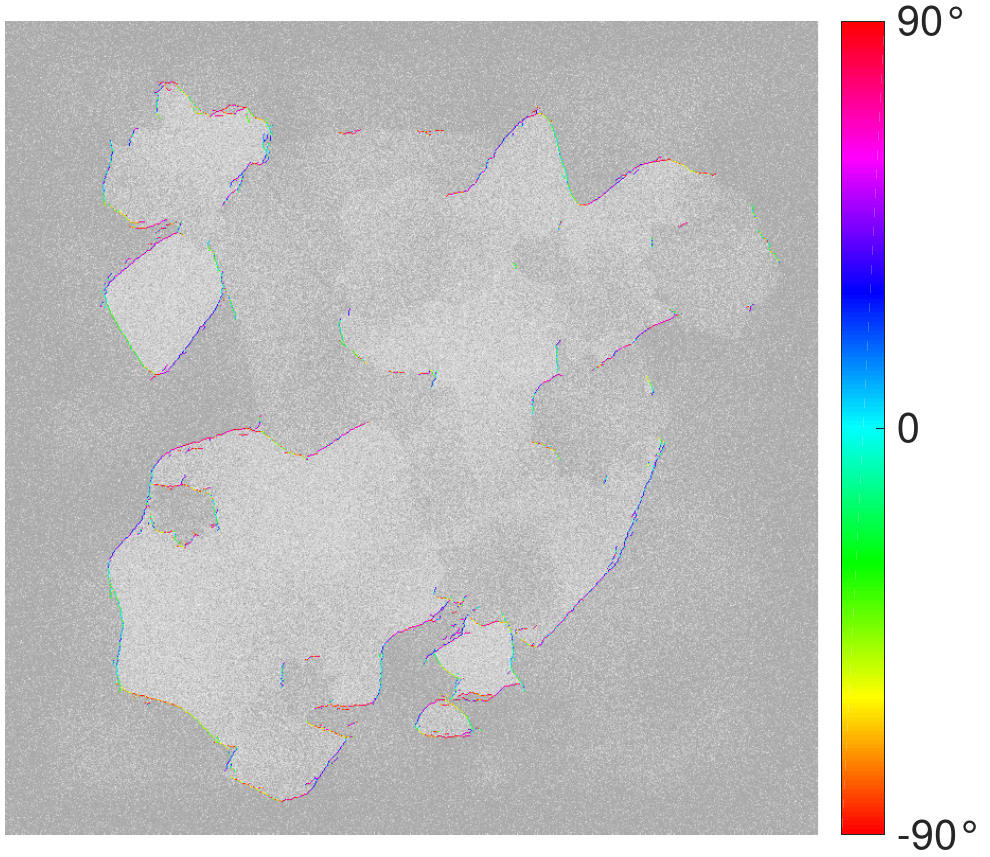}{4cm}{(0.2,0.43)}{(-0.3,0.3)}{2.75cm}}\\
	\subfloat[Canny \cite{Can1986}]{\label{fig:synthimages_edges_comparison_canny}\spyon{3.2}{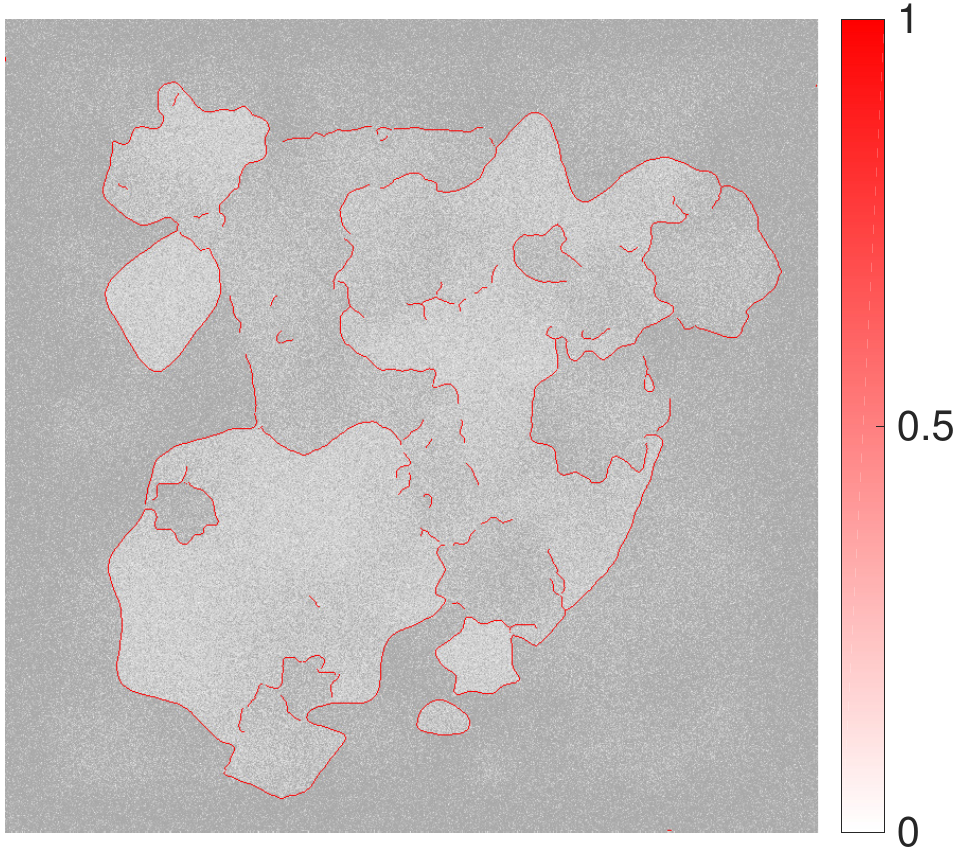}{4cm}{(0.2,0.43)}{(-0.3,0.3)}{2.75cm}}
	\subfloat[Sobel \cite{SF68}]{\label{fig:synthimages_edges_comparison_sobel}\spyon{3.2}{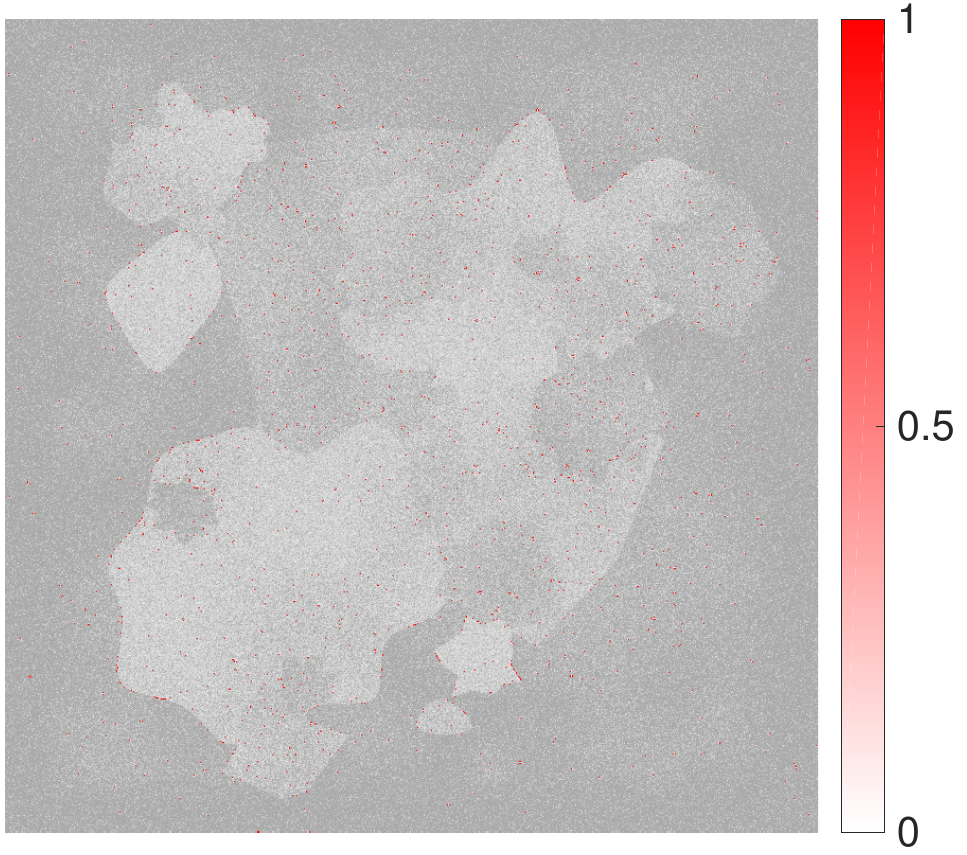}{4cm}{(0.2,0.43)}{(-0.3,0.3)}{2.75cm}}
	\caption{Visual comparison of the detection results yielded by different edge detection methods. Where applicable, the detected edge contour is shown in combination with estimates of the local tangent directions.}
	\label{fig:synthimages_edges_comparison}
\end{figure}

We compare the performance of SymFD with results obtained from the Canny edge detector \cite{Can1986}, the Sobel edge filter \cite{SF68}, the phase congruency measure \cite{Kov1999}, and a shearlet-based edge detector \cite{YLEK2009}, which can be seen as a shearlet-based generalization of Canny. In the case of the Canny and Sobel edge detectors, we applied the implementation provided by \matlab{}'s \SFDparam{edge} function. The phase congruency measure was computed with the \SFDparam{phasecong3} function \cite{KovONLINE} and an implementation of the shearlet-based edge detector was kindly provided by the authors of \cite{YLEK2009}. A visual comparison of the results yielded by the different methods when processing \synthtwo{} with severe distortions may be found in Figure~\ref{fig:synthimages_edges_comparison}. The SymFD edge measure $\EM(f,y)$ and the tangent orientation measure $\OME(f,y)$ after post-pro\-cess\-ing are plotted in Figure~\ref{fig:synthimages_edges_SFD} in the Supplementary Materials in the case of \synthone{} distorted with medium noise. 

The detection performance of the considered algorithms for both synthetic images and three different levels of noise are summarized in Table~\ref{tab:synthimages_edges_comparison}. The edge detection performance was measured in terms of the $\FOM$. Estimates of the local tangent directions are compared to the ground truth via the $\MAEOME$. Note that the $\TPR$ is equal to the percentage of ground truth feature points that were used in the computation of the $\MAEOME$ (cf. \eqref{eq:MAE} and \eqref{eq:tpfptpr}). The radius defining the set of true positives $\TPset$ was set to three pixels. Note that out of the algorithms included in the evaluation, only SymFD, the phase congruency measure, and the shearlet-based edge detector provide estimates of tangent orientations.

\subsection{Ridge Detection}
\begin{figure}[t!]
	\centering
	\subfloat[Input (\synththree, medium noise)]{\label{fig:synthimages_ridges_SFD_input}\spyon{3.2}{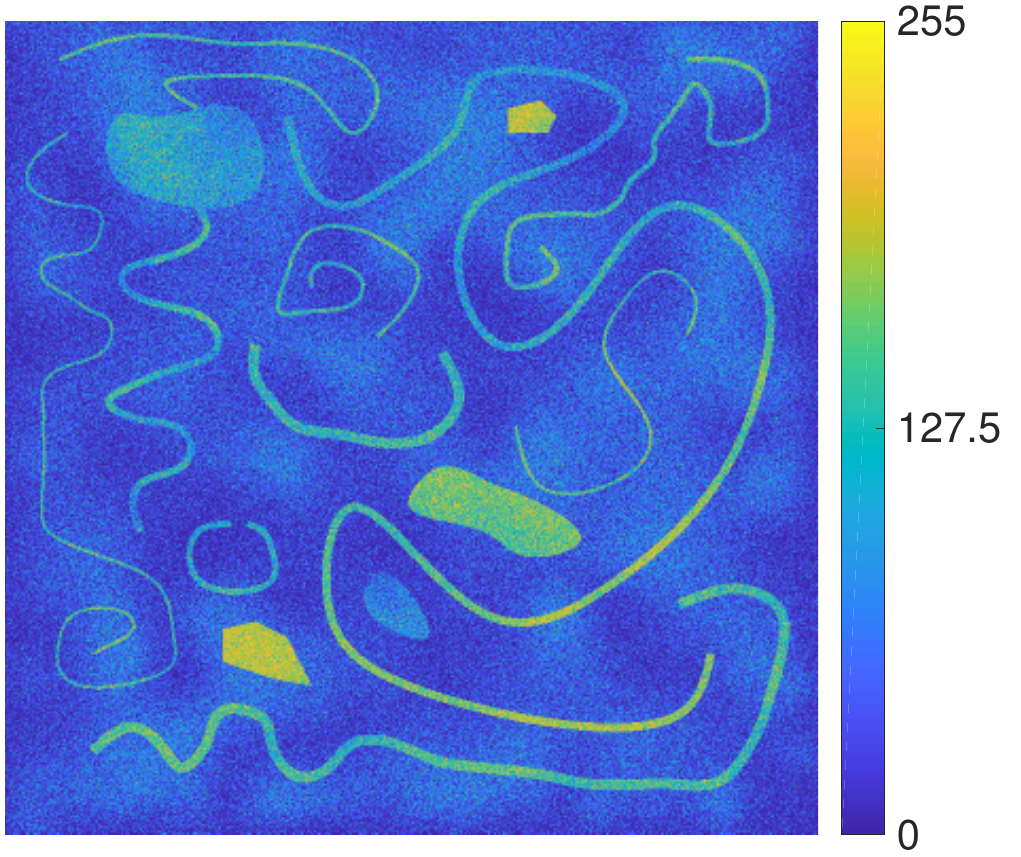}{3.6cm}{(0.525,0.655)}{(-0.3,0.3)}{2.4cm}}
	\subfloat[SymFD ridge measure $\RM(f,y)$]{\label{fig:synthimages_ridges_SFD_fm}\spyon{3.2}{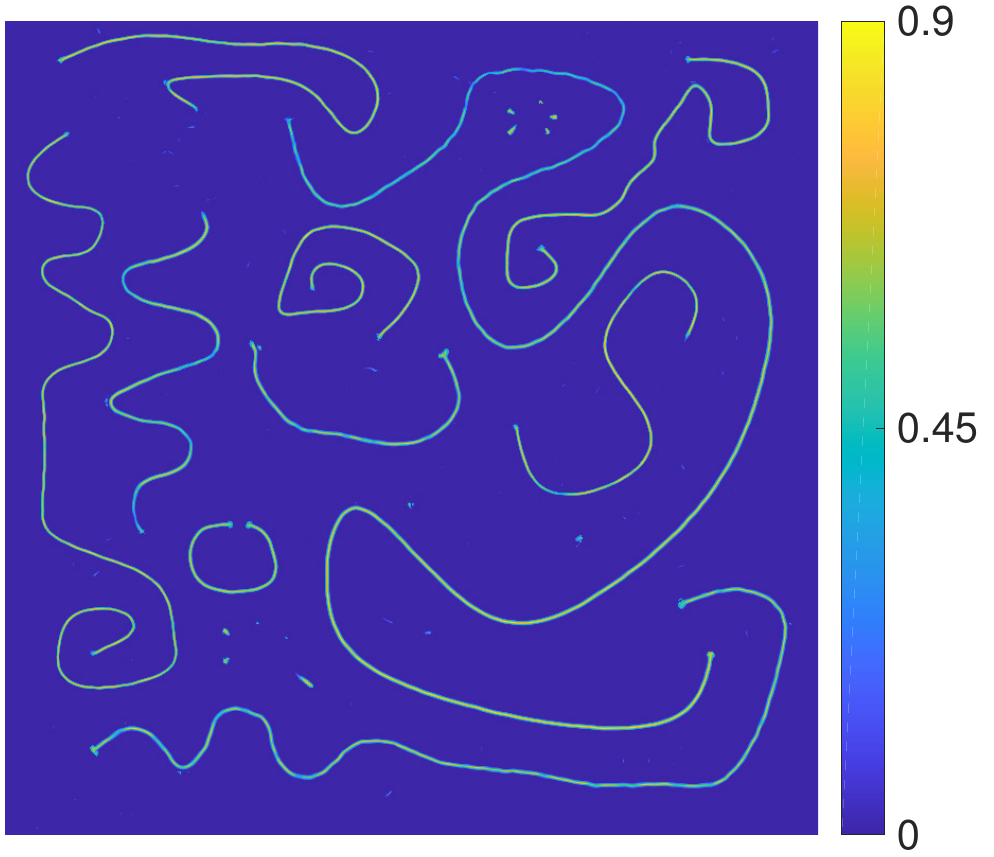}{3.6cm}{(0.54,0.655)}{(-0.3,0.3)}{2.4cm}}\\
	\subfloat[Tangent directions $\OMR(f,y)$ after post-processing]{\label{fig:synthimages_ridges_SFD_thinned_width}\spyon{3.2}{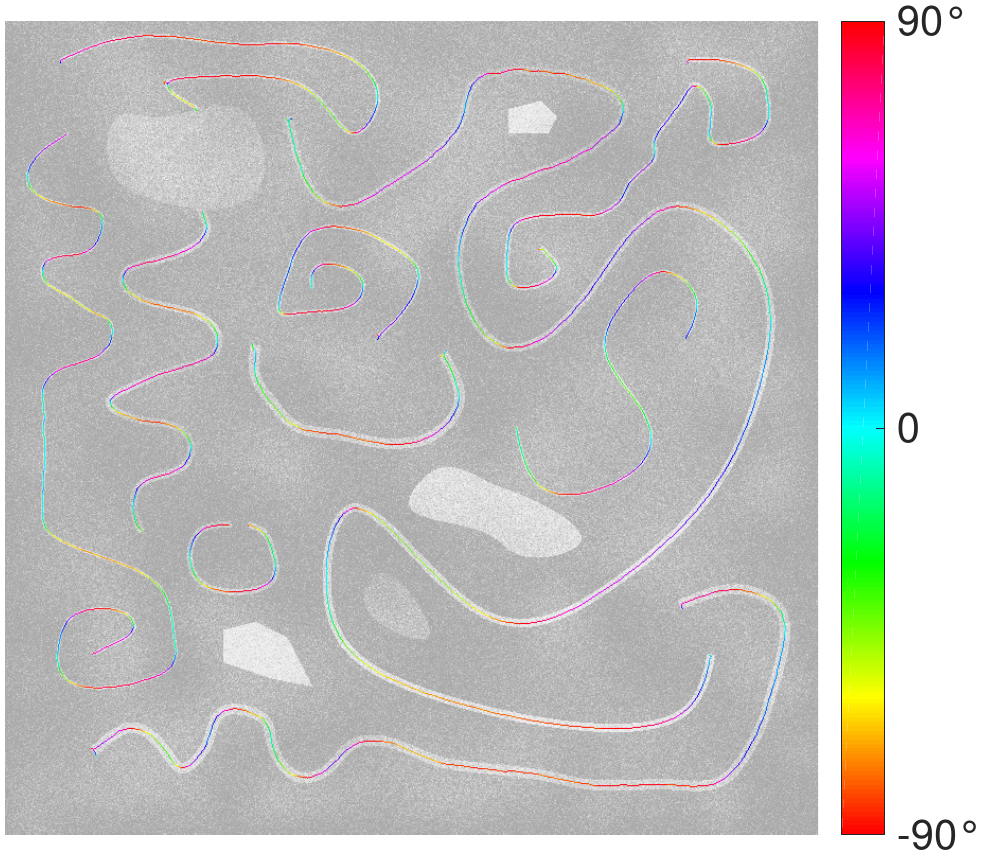}{3.6cm}{(0.545,0.655)}{(-0.3,0.3)}{2.4cm}}
	\subfloat[Widths $\WMR(f,y)$ after post-processing]{\label{fig:synthimages_ridges_SFD_thinned_orientation}\spyon{3.2}{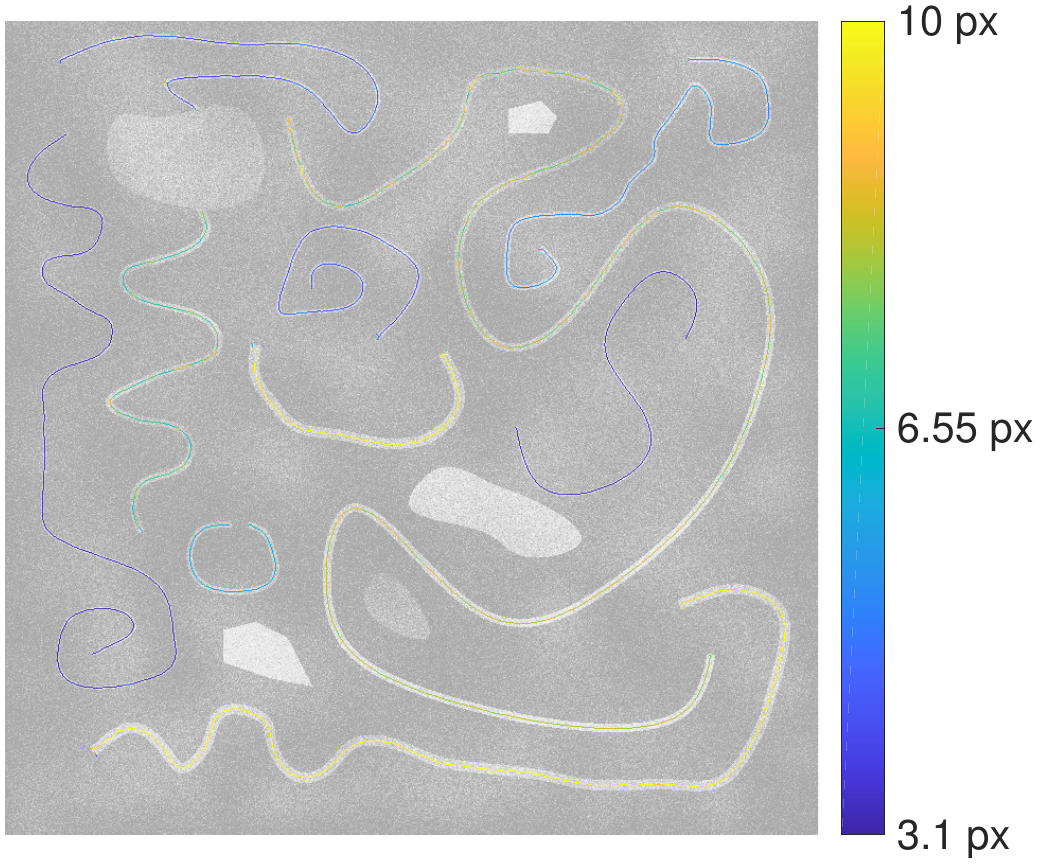}{3.6cm}{(0.51,0.655)}{(-0.3,0.3)}{2.4cm}}
	\caption{Detection of ridges, local tangent orientations, and ridge widths in a synthetic image. The results were yielded by SymFD with parameters $\psie = \frac{G_2}{\normL{G_2}}$, $\psio = \frac{\HT G_2}{\normL{\HT G_2}}$, $\SFDparammath{maxFeatureWidth} = 10$, $\SFDparammath{maxFeatureLength} = 15$, $\alpha = 0.2$, $\SFDparammath{minFeatureWidth} = 3$, $\SFDparammath{scalesPerOctave} = 6$, $\SFDparammath{nOrientations} = 16$, $\jo = 1$, and $\beta = 20$. All ridges in the input have positive contrast. The analysis is therefore restricted to locations $y$ where the height measure is positive, i.e., $\HMR(f,y) \geq 0$. The input image is of size $768\times 768$ and was distorted by a combination of Gaussian and Poisson noise.}
	\label{fig:synthimages_ridges_SFD}
\end{figure}
Similar to the case of edge detection, we generated synthetic digital images that contain exemplary ridges whose centerlines are based on smooth spline curves. The sequences of points defining the spline curves were again predefined in \synththree{} and chosen in a semi-random fashion in \synthfour{}. To obtain ridges, the spline curves were thickened by a value ranging from three to ten pixels. In the case of \synththree{}, the ridge-shapes were filled using different constant values and different types of gradients, while in \synthfour{}, all ridge-shapes are filled with the same value. To test the capability of an algorithm to correctly differentiate between edges and ridges, both synthetic images also contain filled shapes that are based on closed curves. Furthermore, both images were overlaid with a smooth surface. Visualizations of the noise-free and distorted versions of \synththree{} and \synthfour{} as well as the corresponding ground truths may be found in Figure~\ref{fig:synthimages_ridges} in the Supplementary Materials.%Appendix~\ref{sec:appendixsynth}.
\begin{table}[t]
\caption[Evaluation of four ridge detection methods on synthetic images]{Evaluation of four ridge detection methods on synthetic images.}
\centering
\begin{threeparttable}
	{\setlength{\tabcolsep}{.9pt}\renewcommand{\arraystretch}{1.3}\scriptsize
		\begin{tabular}{lrrrrcrrrrcrrrr}
			\toprule
			&\multicolumn{4}{c}{No noise} & \phantom{ab}&\multicolumn{4}{c}{Medium noise} & \phantom{ab}&\multicolumn{4}{c}{Severe noise}\\
			&$\FOM$ & $\MAEWMR$ & $\MAEOMR$ & $\TPR$ & &$\FOM$ & $\MAEWMR$ & $\MAEOMR$ & $\TPR$ & &$\FOM$ & $\MAEWMR$ & $\MAEOMR$ & $\TPR$\\
			\midrule
			\Synththree{}\\
			SymFD & 0.90 & 0.12\px & 1.20\,$^\circ$ & 100\perc& & 0.89 & 0.26\px & 2.15\,$^\circ$ & 100\perc& & 0.81 & 0.84\px & 4.80\,$^\circ$ & 94\perc\\
			Steger~\cite{steger1998unbiased} & 0.90 & 0.71\px & \multicolumn{1}{c}{n/a} & 100\perc& & 0.85 & 1.31\px & \multicolumn{1}{c}{n/a} & 100\perc& & 0.81 & 2.56\px & \multicolumn{1}{c}{n/a} & 98\perc\\
			Frangi~\cite{frangi1998multiscale} & 0.80 & \multicolumn{1}{c}{n/a} & 1.24\,$^\circ$ & 99\perc& & 0.80 & \multicolumn{1}{c}{n/a} & 2.18\,$^\circ$ & 99\perc& & 0.20 & \multicolumn{1}{c}{n/a} & 3.26\,$^\circ$ & 23\perc\\
			PhaseCong~\cite{Kov1999} & 0.89 & \multicolumn{1}{c}{n/a} & 15.67\,$^\circ$ & 99\perc& & 0.81 & \multicolumn{1}{c}{n/a} & 31.39\,$^\circ$ & 96\perc& & 0.48 & \multicolumn{1}{c}{n/a} & 36.07\,$^\circ$ & 52\perc\\
			\Synthfour{}\\
			SymFD & 0.88 & 0.55\px & 4.18\,$^\circ$ & 99\perc& & 0.87 & 0.59\px & 4.28\,$^\circ$ & 98\perc& & 0.84 & 0.77\px & 5.12\,$^\circ$ & 97\perc\\
			Steger~\cite{steger1998unbiased} & 0.88 & 1.12\px & \multicolumn{1}{c}{n/a} & 98\perc& & 0.87 & 1.80\px & \multicolumn{1}{c}{n/a} & 99\perc& & 0.83 & 1.85\px & \multicolumn{1}{c}{n/a} & 95\perc\\
			Frangi~\cite{frangi1998multiscale} & 0.85 & \multicolumn{1}{c}{n/a} & 4.30\,$^\circ$ & 98\perc& & 0.85 & \multicolumn{1}{c}{n/a} & 4.55\,$^\circ$ & 97\perc& & 0.82 & \multicolumn{1}{c}{n/a} & 6.94\,$^\circ$ & 97\perc\\
			PhaseCong~\cite{Kov1999} & 0.81 & \multicolumn{1}{c}{n/a} & 25.71\,$^\circ$ & 97\perc& & 0.82 & \multicolumn{1}{c}{n/a} & 29.18\,$^\circ$ & 96\perc& & 0.64 & \multicolumn{1}{c}{n/a} & 33.55\,$^\circ$ & 74\perc\\\midrule
		\end{tabular}}
		\begin{tablenotes}\footnotesize
			\item $\FOM$: figure of merit, $\MAEWMR$, $\MAEOMR$: mean \rev{absolute} error of the respective width and tangent direction measures, $\TPR$: true positive rate. Note that $\MAEWMR$ and $\MAEOMR$ are computed only with respect to points in the ground truth that are sufficiently close to points detected by an algorithm. The ratio of considered points is indicated by $\TPR$.
		\end{tablenotes}
	\end{threeparttable}
	\label{tab:synthimages_ridges_comparison}
\end{table}
				
Detection results for a noisy version of \synththree{} yielded by the ridge measure $\RM(f,y)$, the tangent orientation measure $\OMR(f,y)$, and the width measure $\WMR(f,y)$ are depicted in Figure~\ref{fig:synthimages_ridges_SFD}. The ridge detection performance of SymFD is furthermore compared with results obtained from the phase congruency measure \cite{Kov1999}, the vessel enhancement filter proposed by Frangi et al.\ \cite{frangi1998multiscale}, and a widely used scale space-based ridge detection method proposed by Steger \cite{steger1998unbiased}. For the phase congruency measure, we applied the same implementation already used during the evaluation of different edge detection methods. An implementation of the Frangi vesselness measure may be found on the \matlab{} Central File Exchange \cite{frangiOnline}. To obtain binary detection results, the feature maps computed by the Frangi vesselness measure were further processed with the same thresholding and thinning methods that are implemented in SymFD. For the Steger ridge detection method, we used an implementation provided by a freely available ImageJ plugin \cite{stegerOnline}.

The ridge detection performances of SymFD, the phase congruency measure, the Frangi vessel enhancement filter, and the method proposed by Steger are summarized in Table~\ref{tab:synthimages_ridges_comparison}. The similarity between the ridge location ground truth and the map of feature points detected by a single algorithm is again measured in terms of the $\FOM$. Estimates of the local tangent directions and ridge widths are compared to the ground truth via the $\MAEOMR$ and the $\MAEWMR$, respectively. The radius defining the set of true positives $\TPset$ was set to three pixels. Similar to the evaluation of edge detection methods, errors in width and tangent direction measurements are only computed for feature points in the set $\TPset$. Out of the algorithms included in the evaluation, SymFD, the phase congruency measure and the implementation applied to obtain the Frangi vesselness measure also provide estimates of local tangent directions. Furthermore, only SymFD and the Steger ridge detector yield estimates of local ridge widths. A visual comparison of the results yielded by the different methods when processing \synthfour{} with severe distortions may be found in Figure~\ref{fig:synthimages_ridges_comparison} in the Supplementary Materials.
\subsection{Blobs}
%%
%\begin{figure}[t!]
%	\centering
%	\subfloat[Input (\synthsix{}, medium noise)]{\label{fig:synthimages_blobs_SFD_input}\spyon{3.2}{figs/01_blobEvalSmall_gauss_50_poiss_1_input.pdf}{3.6cm}{(0.22,0.825)}{(-0.3,0.3)}{2.4cm}}
%	\subfloat[SymFD blob measure $\BM(f,y)$]{\label{fig:synthimages_blobs_SFD_fm}\spyon{3.2}{figs/03_blobEvalSmall_gauss_50_poiss_1_SFD_featureMap.pdf}{3.6cm}{(0.235,0.825)}{(-0.3,0.3)}{2.4cm}}\\
%	\subfloat[Widths $\WMB(f,y)$ after post-processing]{\label{fig:synthimages_blobs_SFD_thinned}\spyon{3.2}{figs/05_blobEvalSmall_gauss_50_poiss_1_SFD_featureMap_thinned.jpg}{3.6cm}{(0.215,0.825)}{(-0.3,0.3)}{2.4cm}}
%	\subfloat[Error map for $\WMB(f,y)$]{\label{fig:synthimages_blobs_SFD_error_width}\spyon{3.2}{figs/10_blobEvalSmall_gauss_50_poiss_1_SFD_errorMap_width.jpg}{3.6cm}{(0.222,0.825)}{(-0.3,0.3)}{2.4cm}}
%	\label{fig:synthimages_blobs_SFD}
%	\caption{Detection of blobs and blob widths (diameters) in a synthetic image. The results were yielded by SymFD with parameters $\psie = \frac{\HT G_1}{\normL{\HT G_1}}$, $\SFDparammath{maxFeatureWidth} = 16$, $\SFDparammath{maxFeatureLength} = 16$, $\alpha = 1$, $\SFDparammath{minFeatureWidth} = 6$, $\SFDparammath{scalesPerOctave} = 8$, $\SFDparammath{nOrientations} = 16$, $\jo = 0$, and $\beta = 5$. All blobs in the input have positive contrast. The analysis is therefore restricted to locations $y$ where the height measure is positive, i.e., $\HMB(f,y) \geq 0$. The input image is of size $768\times 768$ and was distorted by a combination of Gaussian and Poisson noise.}
%\end{figure}
				
To evaluate the proposed blob measure $\BM(f,y)$ and width estimates $\WMB(f,y)$, we consider two synthetic images that show aggregations of filled circles with varying diameters and contrasts. \Synthfive{} consists of 31 large circles with diameters ranging from 30 to 50 pixels while \synthsix{} contains a total of 200 small blobs whose diameters are between 7 and 13 pixels. In both cases, the center points of the circles were semi-randomly selected in the sense that two center points have at least a distance of 20 pixels in \synthsix{} and 100 pixels in \synthfive{}. Noisy versions of \synthfive{} and \synthsix{} and the corresponding ground truths are visualized in Figure~\ref{fig:synthimages_edges} in the Supplementary Materials.%Appendix~\ref{sec:appendixsynth}.

The circle detection performance of SymFD is compared to the circular Hough transform \cite{atherton1999size,davies2005machinevision}, which can be computed in \matlab{} via the function \SFDparam{imfindcircles}. To compare detection results of an algorithm with the ground truth, we consider the number $\TP$ of true positives with respect to a radius of six pixels, as defined in \eqref{eq:tpfptpr}, and the number $\FP$ of false positives. Here, $\FP$ is equal to the number of detected blob center points which are not in the vicinity of an actual center of a blob. That is, $\FP$ is equal to the number of elements in the set $\FPset = \{ p_1 \in \Pdet \colon \min_{p_2 \in \Pgt}\|p_1 - p_2\| > 6 \}$. We further consider the mean absolute errors in terms of center point location and blob width denoted by $\MAECTR$ and $\MAEWMB$, respectively. The results of the numerical evaluation of the blob detection performance of SymFD and the circular Hough transform are summarized in Table~\ref{tab:synthimages_blobs_comparison}.
\begin{table}[t]
\centering
\caption[Evaluation of two blob detection methods on synthetic images]{Evaluation of two blob detection methods on synthetic images.}
\begin{threeparttable}
	{\setlength{\tabcolsep}{.9pt}\renewcommand{\arraystretch}{1.3}\scriptsize\begin{tabular}{lrrrrcrrrrcrrrr}
			\toprule
			&\multicolumn{4}{c}{No noise} & \phantom{abc}&\multicolumn{4}{c}{Medium noise} & \phantom{abc}&\multicolumn{4}{c}{Severe noise}\\
			&$\TP$ & $\FP$ & $\MAECTR$ & $\MAEWMB$ & &$\TP$ & $\FP$ & $\MAECTR$ & $\MAEWMB$ & &$\TP$ & $\FP$ & $\MAECTR$ & $\MAEWMB$\\
			\midrule
			\Synthfive{} (31 blobs)\\
			SymFD & 31 & 0 & 2.16\px & 2.62\px& & 31 & 0 & 2.16\px & 2.95\px& & 31 & 0 & 2.09\px & 3.13\px\\
			Circular Hough~\cite{atherton1999size,davies2005machinevision} & 31 & 0 & 0.99\px & 0.61\px& & 30 & 0 & 1.82\px & 2.33\px& & 20 & 0 & 2.03\px & 8.15\px\\
			\Synthsix{} (200 blobs)\\
			SymFD & 200 & 0 & 1.10\px & 0.87\px& & 198 & 0 & 1.04\px & 0.97\px& & 155 & 1 & 1.30\px & 1.19\px\\
			Circular Hough~\cite{atherton1999size,davies2005machinevision} & 200 & 0 & 0.88\px & 0.56\px& & 194 & 3 & 0.99\px & 1.04\px& & 47 & 1 & 1.10\px & 1.66\px\\\midrule
		\end{tabular}}
				\begin{tablenotes}\footnotesize
					\item \rev{$\TP$: number of true positives, $\FP$: number of false positives}, $\MAECTR$: mean \rev{absolute} error with respect to the location of center points, $\MAEWMB$: mean \rev{absolute} error of the respective width measure. Note that $\MAECTR$ and $\MAEWMB$ are computed only with respect to points in the ground truth that are sufficiently close to points detected by an algorithm.
				\end{tablenotes}
	\end{threeparttable}
	\label{tab:synthimages_blobs_comparison}
\end{table}

\begin{figure}[t!]
	\centering
	\subfloat[Input 1 (\synthfive{}, severe noise)]{\label{fig:synthimages_blobs_comparison_input1}\spyon{3.2}{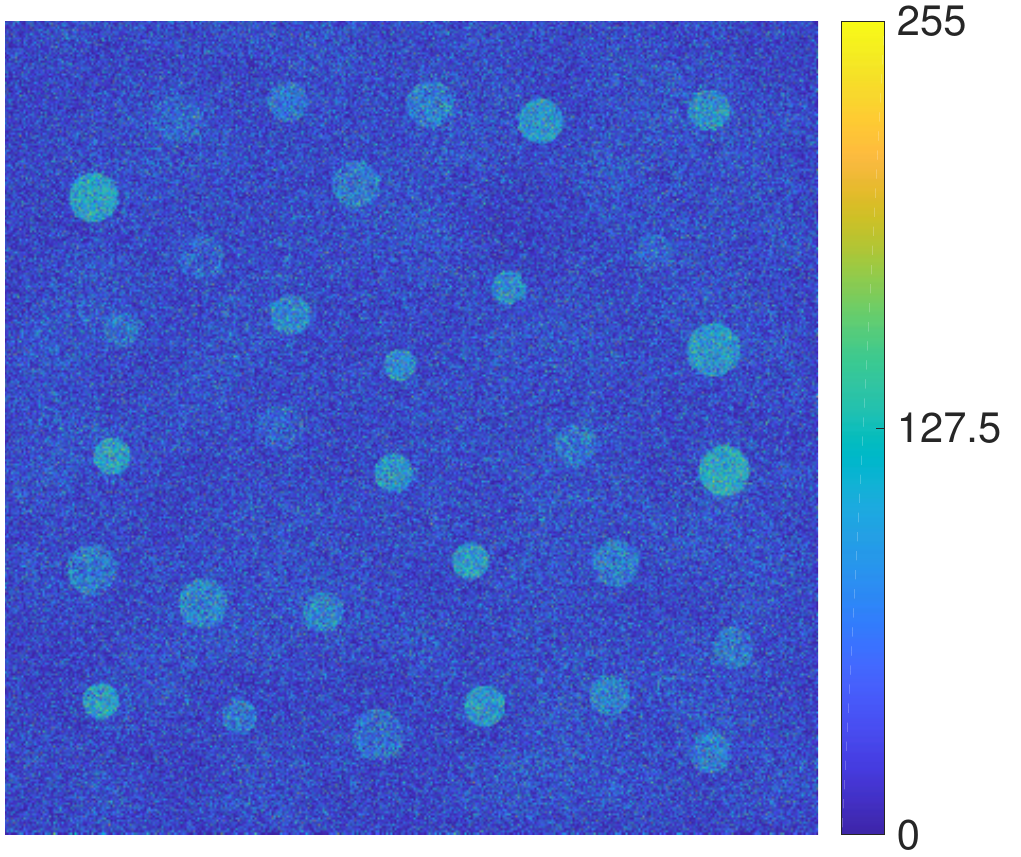}{3.6cm}{(0.24,0.67)}{(-0.3,0.3)}{2.4cm}}
	\subfloat[Input 2 (\synthsix{}, severe noise)]{\label{fig:synthimages_blobs_comparison_input2}\spyon{3.2}{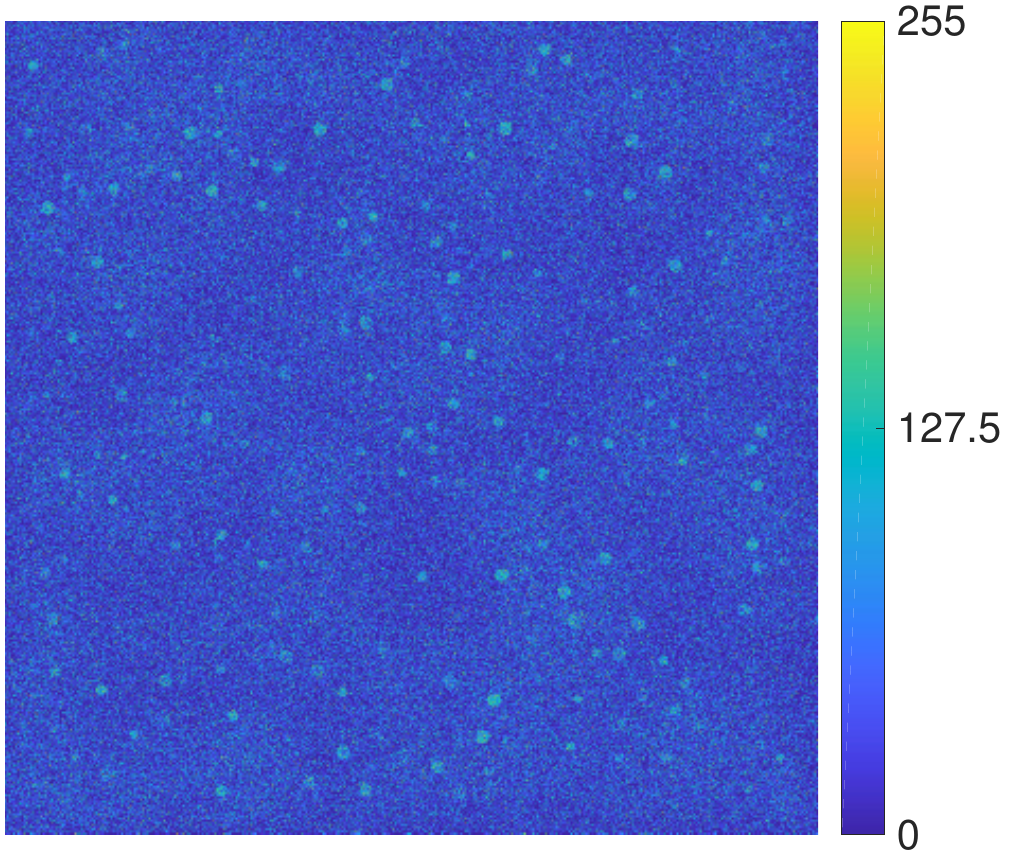}{3.6cm}{(0.22,0.825)}{(-0.3,0.3)}{2.4cm}}\\
	\subfloat[SymFD (input 1)]{\label{fig:synthimages_blobs_comparison_SFD1}\spyon{3.2}{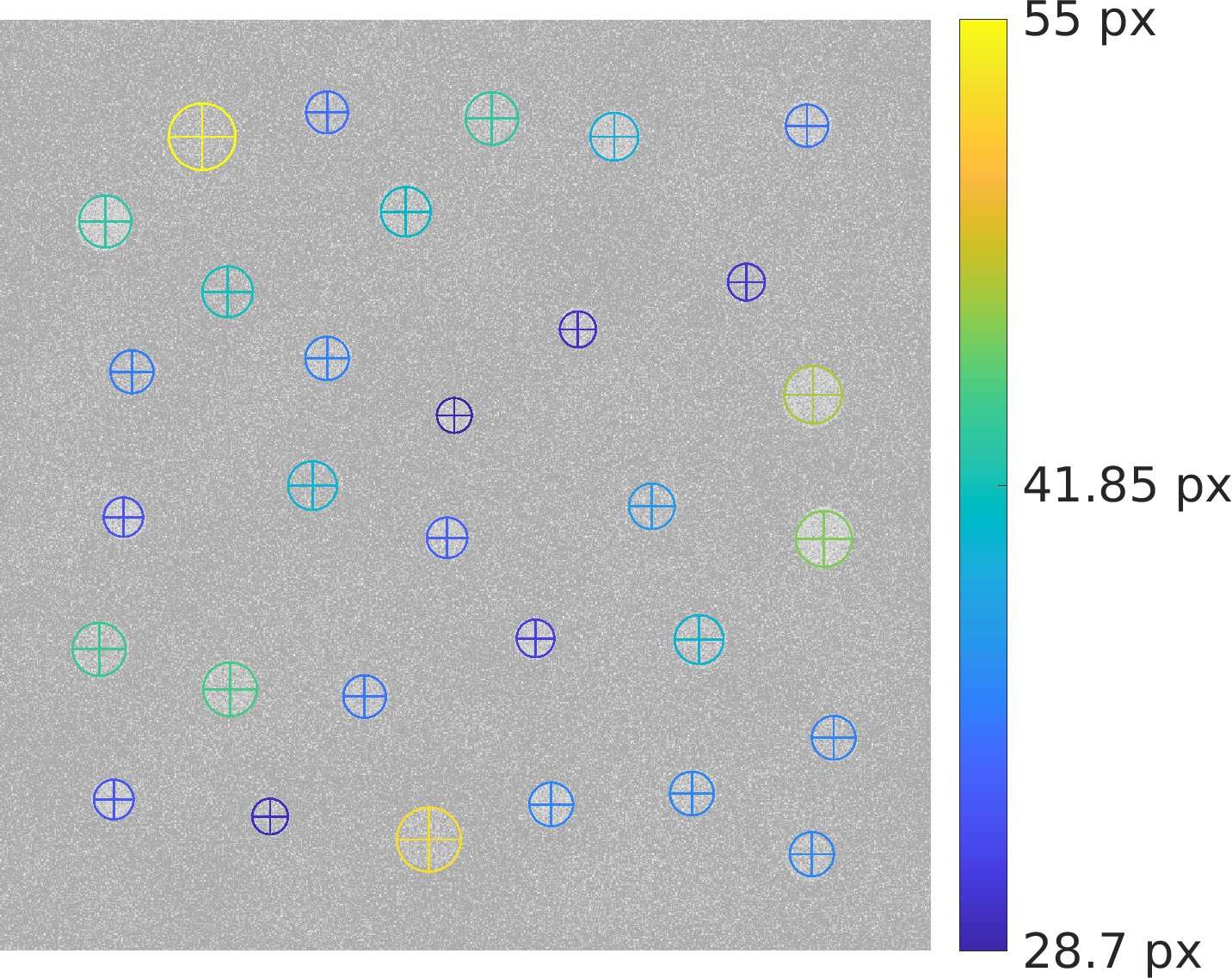}{3.6cm}{(0.24,0.67)}{(-0.3,0.3)}{2.4cm}}
	\subfloat[SymFD (input 2)]{\label{fig:synthimages_blobs_comparison_SFD2}\spyon{3.2}{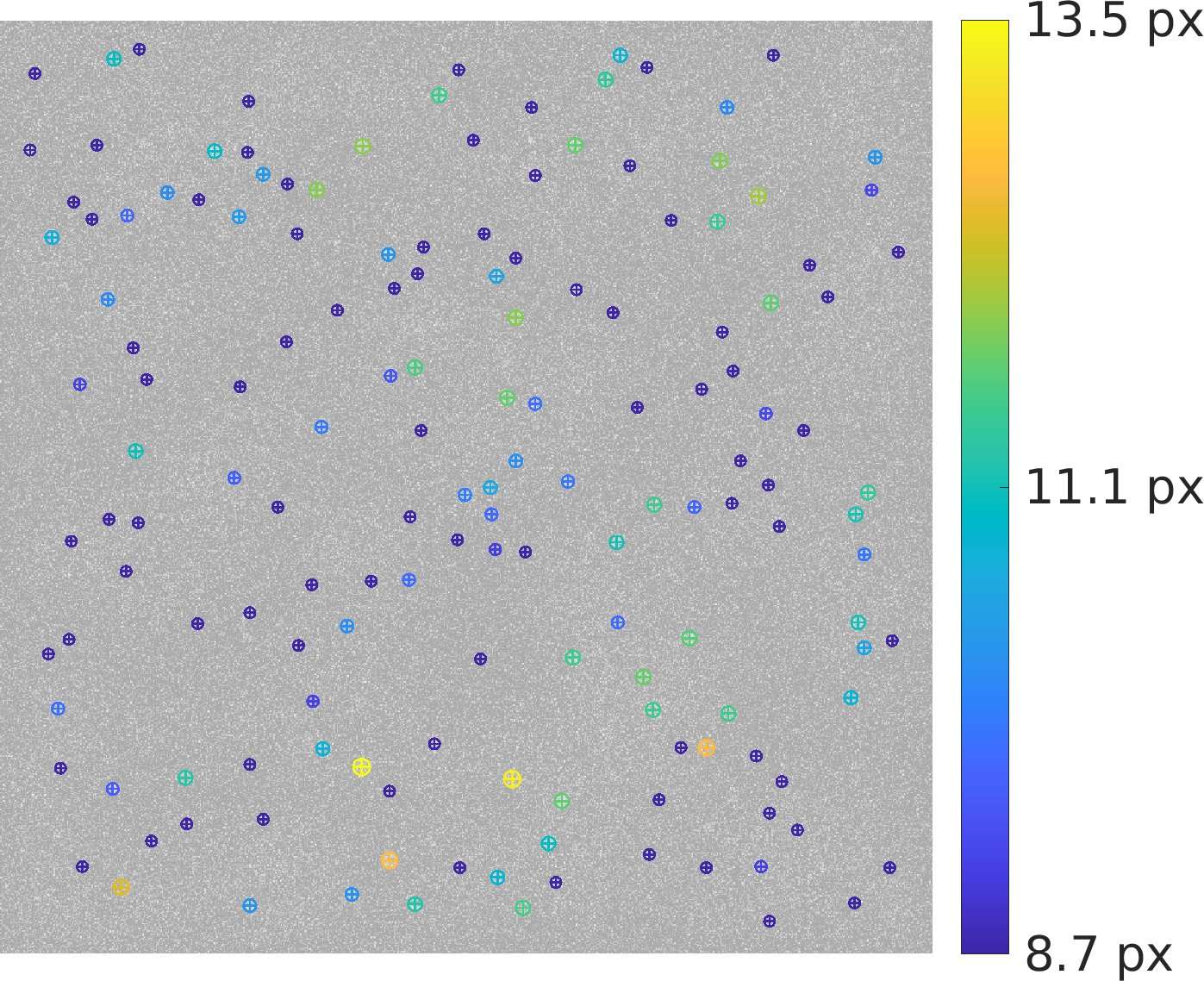}{3.6cm}{(0.215,0.825)}{(-0.3,0.3)}{2.4cm}}\\
	\subfloat[Circular Hough \cite{atherton1999size,davies2005machinevision} (input 1)]{\label{fig:synthimages_blobs_comparison_hough1}\spyon{3.2}{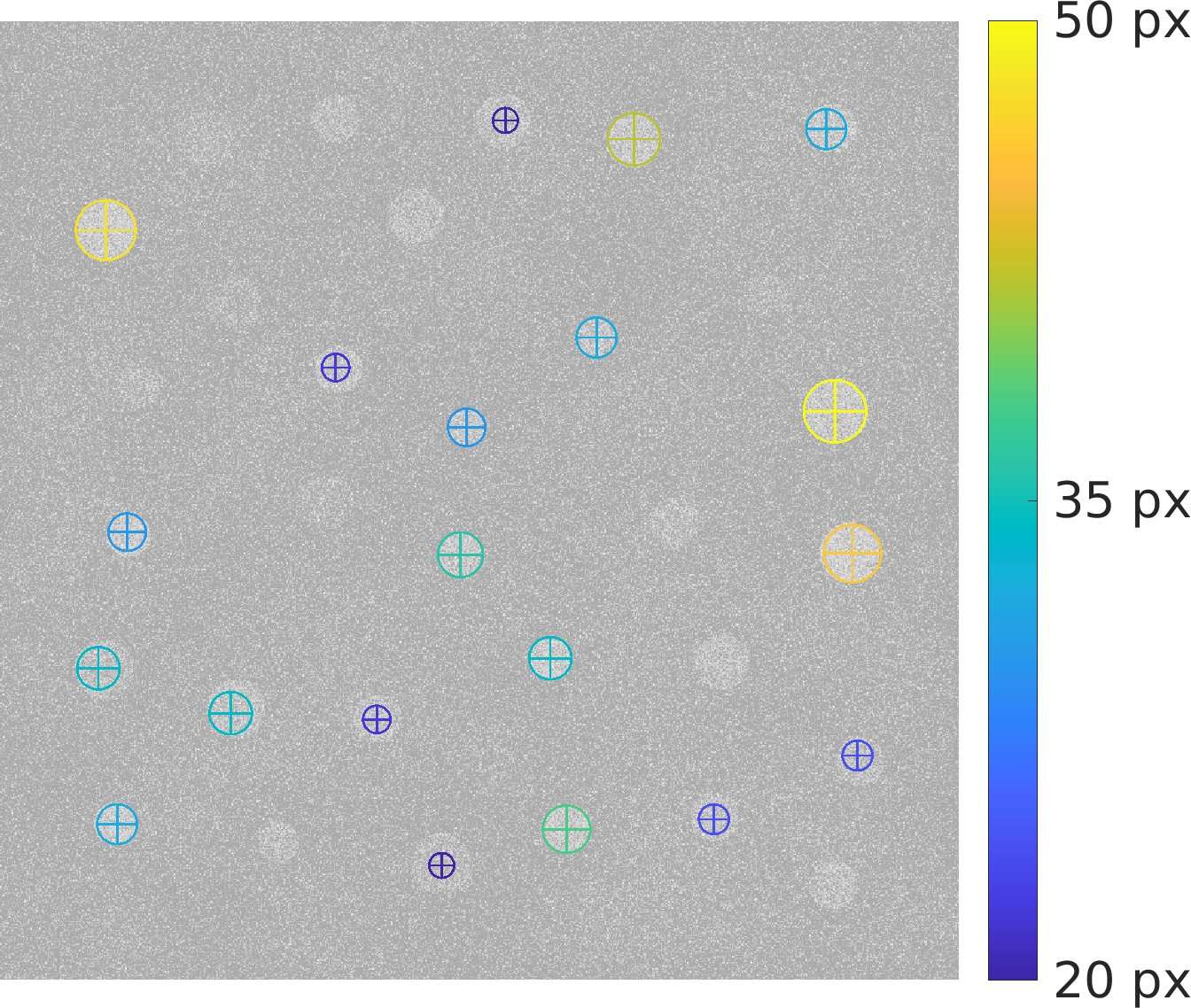}{3.6cm}{(0.24,0.67)}{(-0.3,0.3)}{2.4cm}}
	\subfloat[Circular Hough \cite{atherton1999size,davies2005machinevision} (input 2)]{\label{fig:synthimages_blobs_comparison_hough2}\spyon{3.2}{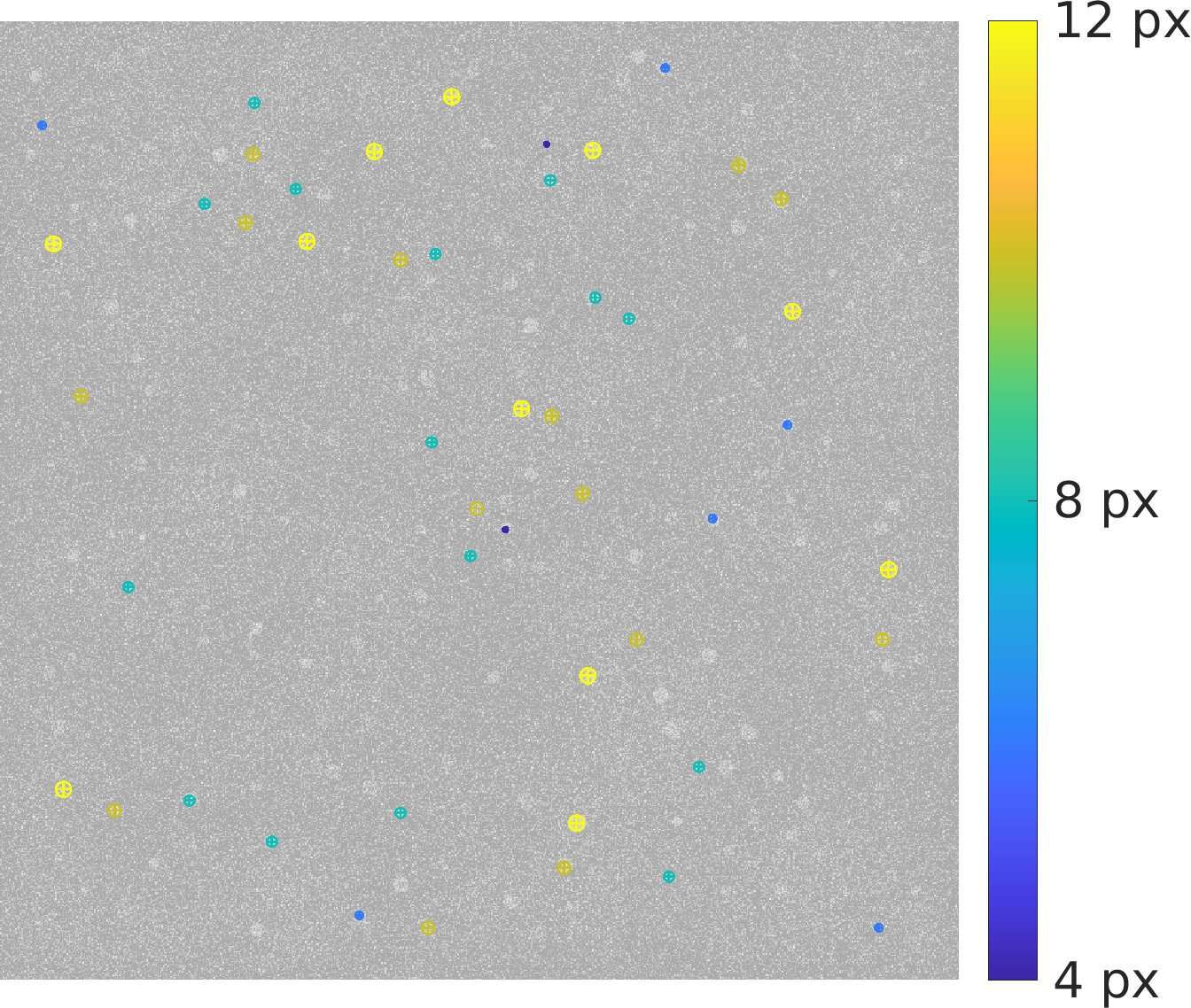}{3.6cm}{(0.222,0.825)}{(-0.3,0.3)}{2.4cm}}
	\caption{Visual comparison of the detection results yielded by the measures $\BM(f,y)$ and $\WMB(f,y)$ implemented in SymFD and the circular Hough transform.}
	\label{fig:synthimages_blobs_comparison}
\end{figure}

A visual comparison of the results yielded by SymFD and the circular Hough transform in the cases of \synthfive{} and \synthsix{} distorted by severe Gaussian and Poisson noise is presented in Figure~\ref{fig:synthimages_blobs_comparison}, and SymFD blob detection results with width error map for \synthsix{} distorted by medium noise are shown in Figure~\ref{fig:synthimages_blobs_SFD} in the Supplementary Materials.

\section{Applications}
\label{sec:applications}
Previous versions of $\EM(f,y)$ and $\RM(f,y)$ that are based on complex-valued shearlets \cite{Rei14,KRKLLH,reisenhofer2016shearlet} were already successfully applied in different feature extraction tasks such as the detection and characterization of flame fronts \cite{reisenhofer2016shearlet,KRKLLH}, the detection of borders of tidal flats in the Wadden Sea \cite{KRKLLH}, the extraction of fracture-traces in rock masses \cite{Bolkas2017}, the detection of channel boundaries in seismic data \cite{karbalaali2017channel,karbalaali2018channel}, and the automated detection of the boundaries of touching cells in scanning electron (SEM) images \cite{visapp18}.
%
%\begin{figure}[t!]
%	\centering
%	\subfloat[Image 7 of the REVIEW VDIS dataset with manually annotated vessel profiles.]{\label{fig:review_gt1}\spyon{6}{figs/05_VDIS007_gt_widths.jpg}{2.8cm}{(0.3,0.295)}{(-0.32,0.3)}{2.3cm}}\hspace{.2cm}
%	\subfloat[Image 2 of the REVIEW CLRIS dataset with manually annotated vessel profiles.]{\label{fig:review_gt2}\spyon{5}{figs/06_CLRIS002_gt_oris.jpg}{2.8cm}{(0.442,0.87)}{(-0.32,0.3)}{2.3cm}}
%	\caption{Examples of digital images in the REVIEW retinal vessel reference dataset \cite{aldiri2008review} with manually annotated ground truth vessel profiles.}
%	\label{fig:review_gt}
%\end{figure}

To further evaluate the applicability of the proposed measures, we consider two tasks which require the correct localization and characterization of certain morphological components in biomedical images. In Section~\ref{sec:retinal}, we demonstrate how the width measure $\WMR(f,y)$ and estimates of local tangent directions yielded by $\OMR(f,y)$ can be used in combination with the ridge measure $\RM(f,y)$ to obtain characterizations of the geometry of retinal blood vessels. In particular, the blood vessel profiles yielded by SymFD are validated and compared to other state-of-the-art algorithms on the publicly available REVIEW database \cite{aldiri2008review}. In Section~\ref{sec:cellcol}, we investigate the applicability of the blob measure $\BM(f,y)$ for automatically counting the number of grown cell colonies in a Petri dish.
\subsection{Determining Widths and Orientations of Retinal Blood Vessels}
\label{sec:retinal}
Retinal imaging is one of only a few existing techniques which provides non-invasive observations of the human circulatory system. Variations in the geometry of the retinal vasculature can furthermore be linked to a wide range of ocular diseases and systemic diseases with a vascular component such as diabetes \cite{cheung2012retinalvascular,bekkers2015curvature}, Alzheimer's disease \cite{williams2015retinalmicro}, or high blood pressure \cite{ikram2006hypertension}. In recent years, many methods have been proposed for automatically extracting and analyzing the geometry of blood vessels in digital retinal images \cite{abramoff2010retinal,patton2006progressinretinal,DEKB11}.

To give an example of how SymFD can be applied in the context of retinal image analysis, we consider the REVIEW retinal vessel reference dataset \cite{aldiri2008review}. For each image in the REVIEW database, three independent experts were asked to manually define vessel edge profiles on a number of preselected segments. Each profile consists of a pair of opposite edge points that indicate the boundaries of a blood vessel. A single profile thus not only contains information about the location of a retinal blood vessel but provides a complete characterization in terms of the local width and the local normal direction of the centerline. Two examples of images in the REVIEW database are shown along with manually marked vessel profiles in Figure~\ref{fig:review_gt} in the Supplemental Materials.

The REVIEW database consists of four subsets of digital images that aim to provide meaningful benchmarks for different types of challenges in retinal image processing. The \emph{high resolution image set} (HRIS) contains four images that show severe cases of diabetic retinopathy. Each image in the HRIS data set has a resolution of $896\times609\times3$ pixels but has previously been sub-sampled by a factor of four such that, disregarding human error, the vessel widths are known to a precision of a quarter of a pixel. The \emph{vascular disease image set} (VDIS) contains eight retinal images of size $1360\times 1024\times 3$ that were randomly selected from a database of images of patients attending a retinopathy clinic. The \emph{central light reflex image set} (CLRIS) consists of only two images of a resolution of $2160\times 1440 \times 3$ pixels that show early atherosclerotic changes in retinal vessels. Finally, the \emph{kick point image set} (KPIS) contains two sub-sampled images where the normal direction of the centerline of a given location was computationally determined from the neighboring pixels on the centerline and the human observers where only asked to identify the kick-points determining the width. In total, the REVIEW database contains 5066 manually marked vessel profiles. 

\begin{figure}[t!]
	\centering
	\subfloat[Input (image 7 of VDIS dataset)]{\label{fig:review_SFD_input}\spyon{3.2}{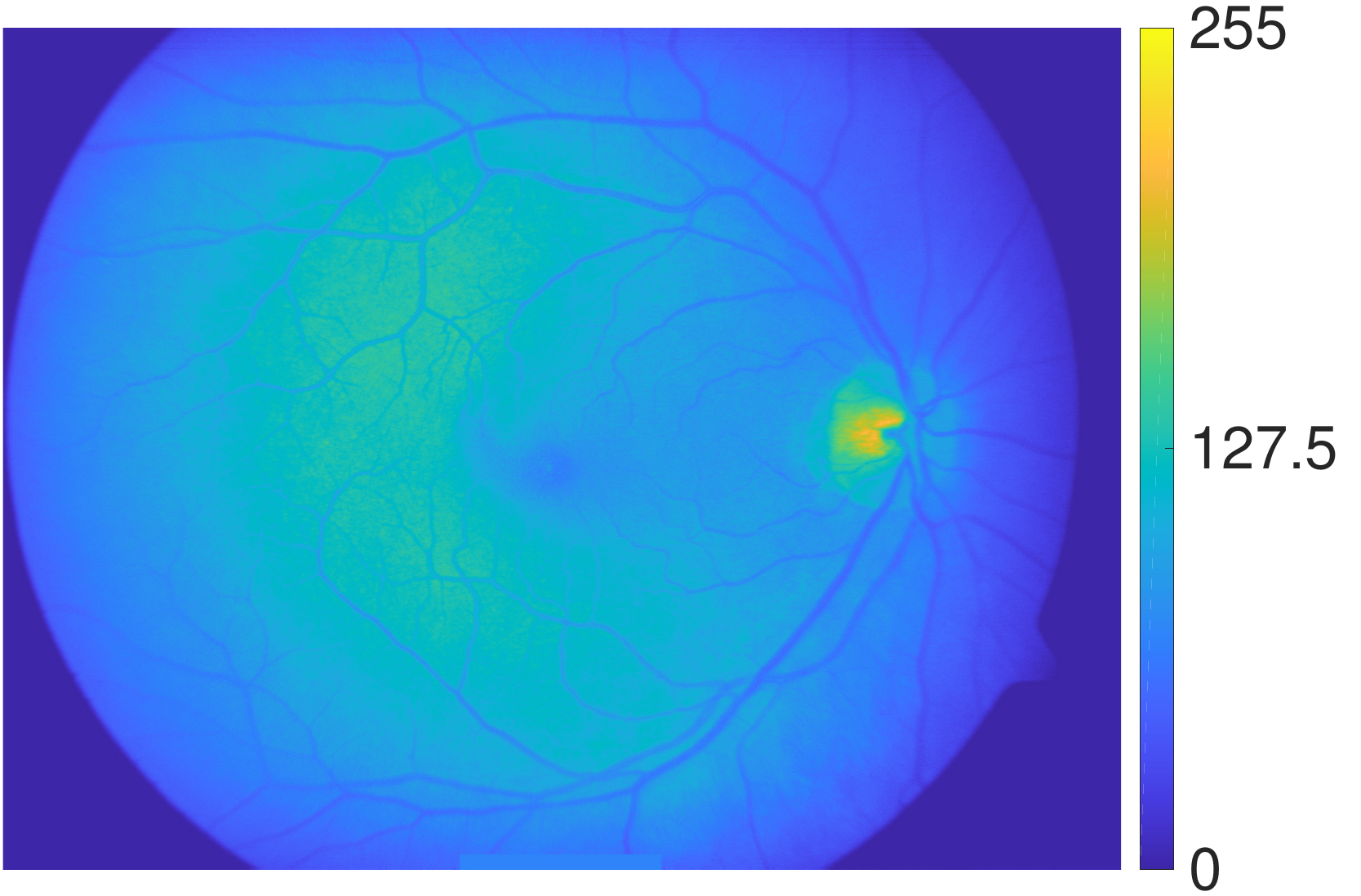}{2.8cm}{(0.3,0.33)}{(-0.31,0.35)}{2.5cm}}\hspace{.2cm}
	\subfloat[SymFD ridge measure $\RM(f,y)$]{\label{fig:review_SFD_fm}\spyon{3.2}{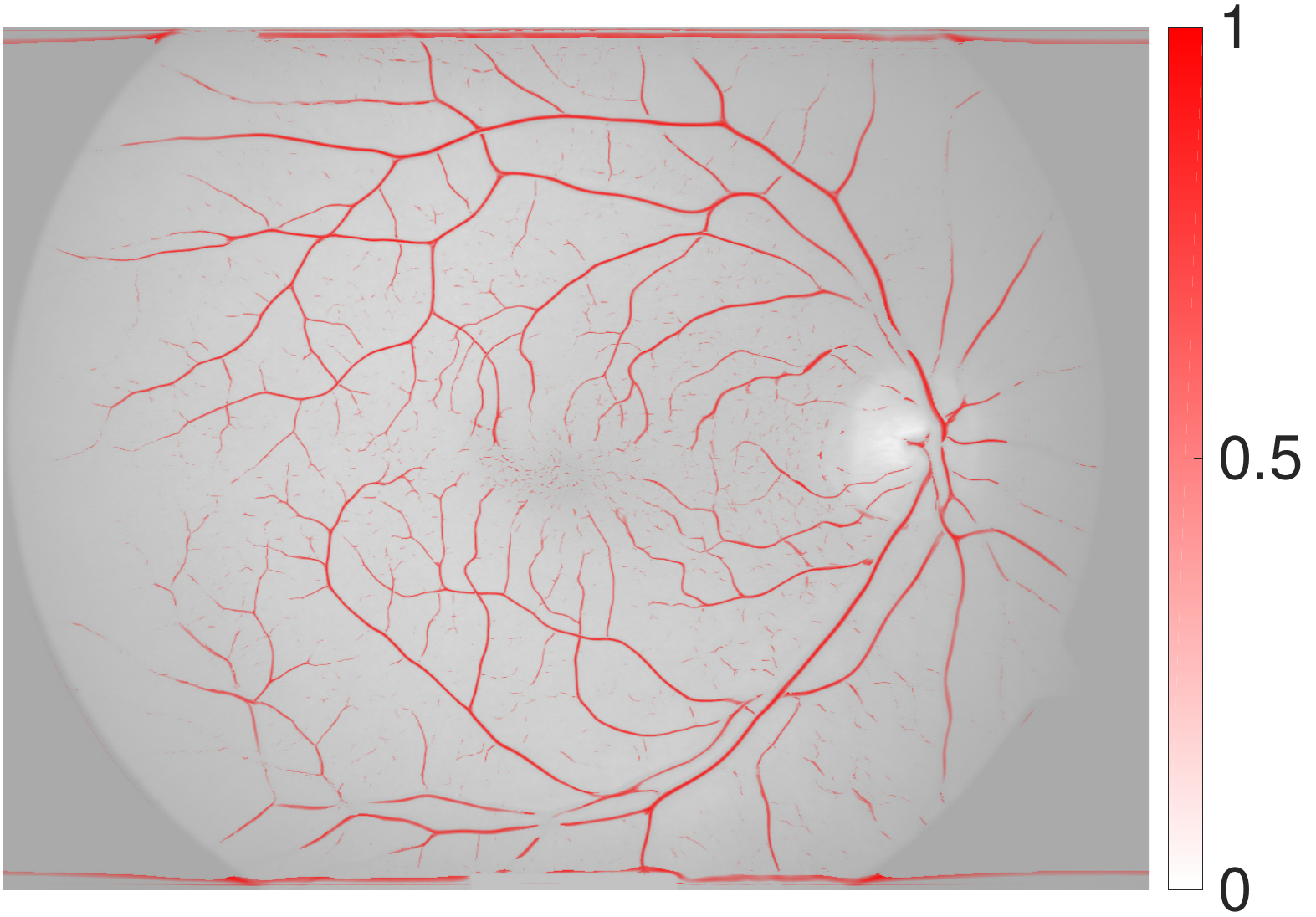}{2.8cm}{(0.3,0.33)}{(-0.32,0.35)}{2.5cm}}\\
	\subfloat[SymFD width measure $\WMR(f,y)$]{\label{fig:review_SFD_width}\spyon{3.2}{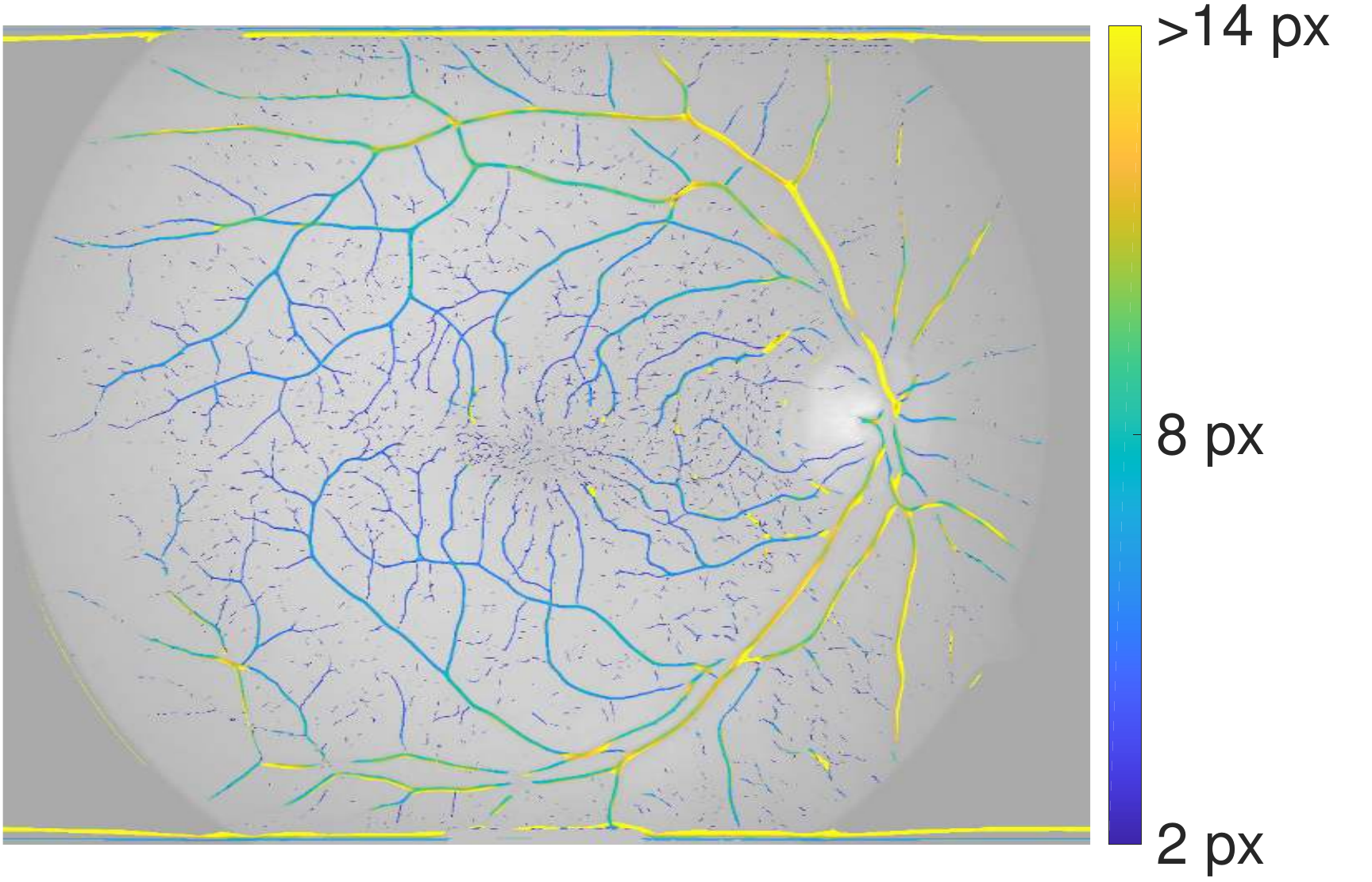}{2.8cm}{(0.28,0.345)}{(-0.3,0.35)}{2.5cm}}\hspace{.2cm}
	\subfloat[SymFD tangent direction estimates $\OMR(f,y)$]{\label{fig:review_SFD_ori}\spyon{3.2}{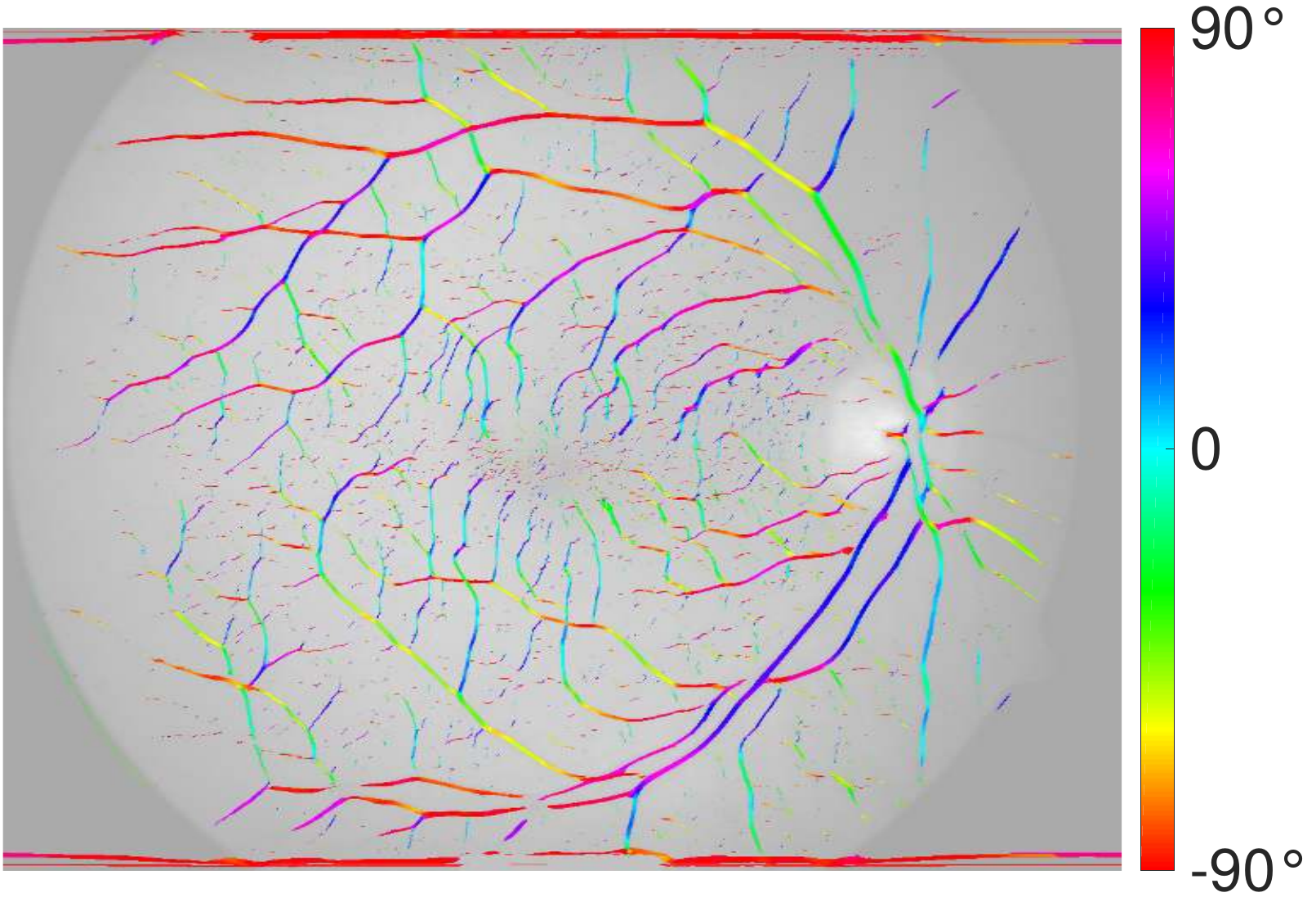}{2.8cm}{(0.3,0.33)}{(-0.32,0.35)}{2.5cm}}\\
	\subfloat[Errors for $\WMR(f,y)$. Widths and orientations of the plotted vessel edge profiles are based on the SymFD-measures $\WMR(f,y)$ and $\OMR(f,y)$.]{\label{fig:review_SFD_error_width}\spyon{6}{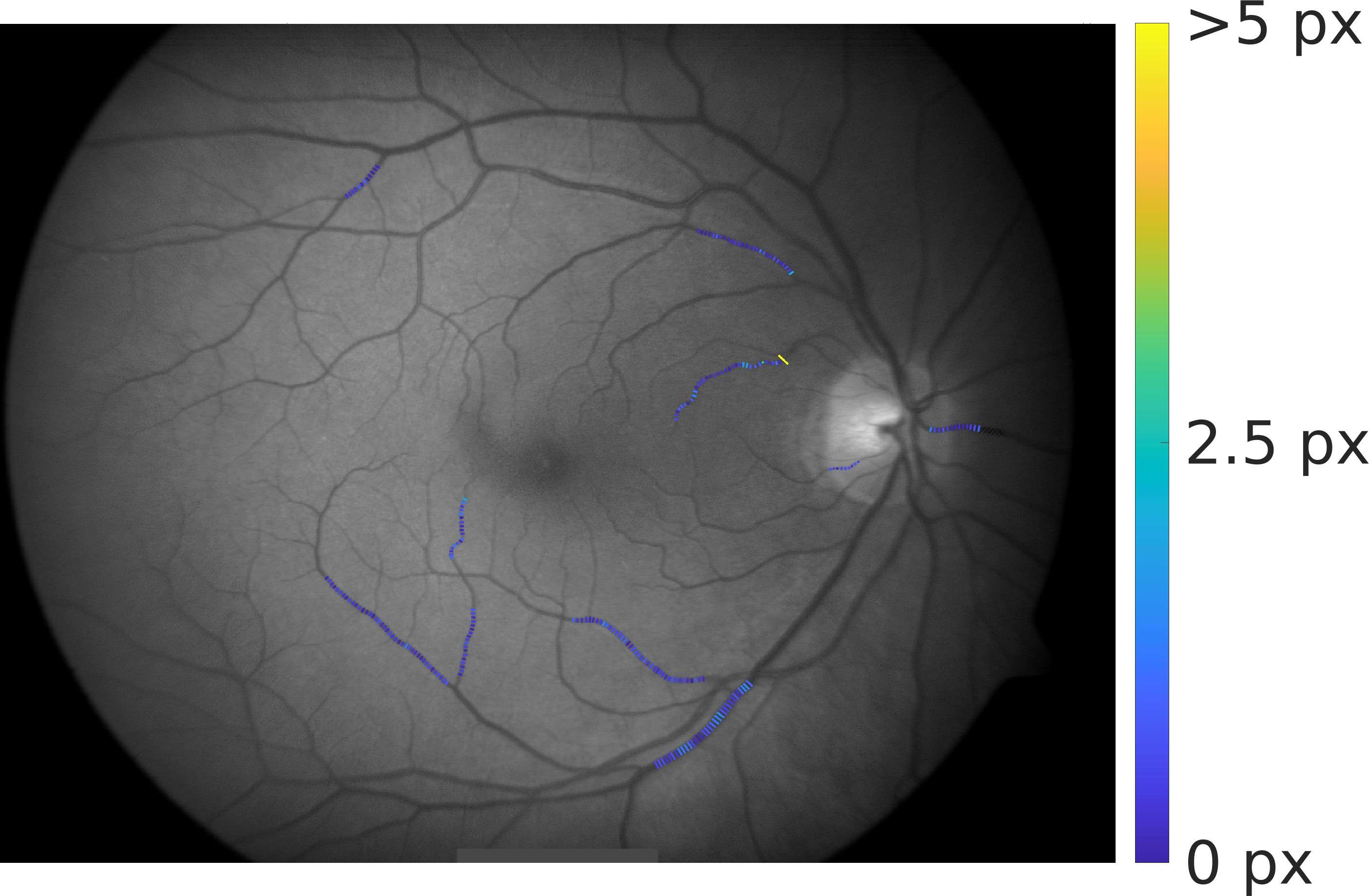}{2.8cm}{(0.305,0.295)}{(-0.32,0.35)}{2.5cm}}\hspace{.2cm}
	\subfloat[Errors for $\OMR(f,y)$. Widths and orientations of the plotted vessel edge profiles are based on the SymFD-measures $\WMR(f,y)$ and $\OMR(f,y)$.]{\label{fig:review_SFD_error_ori}\spyon{6}{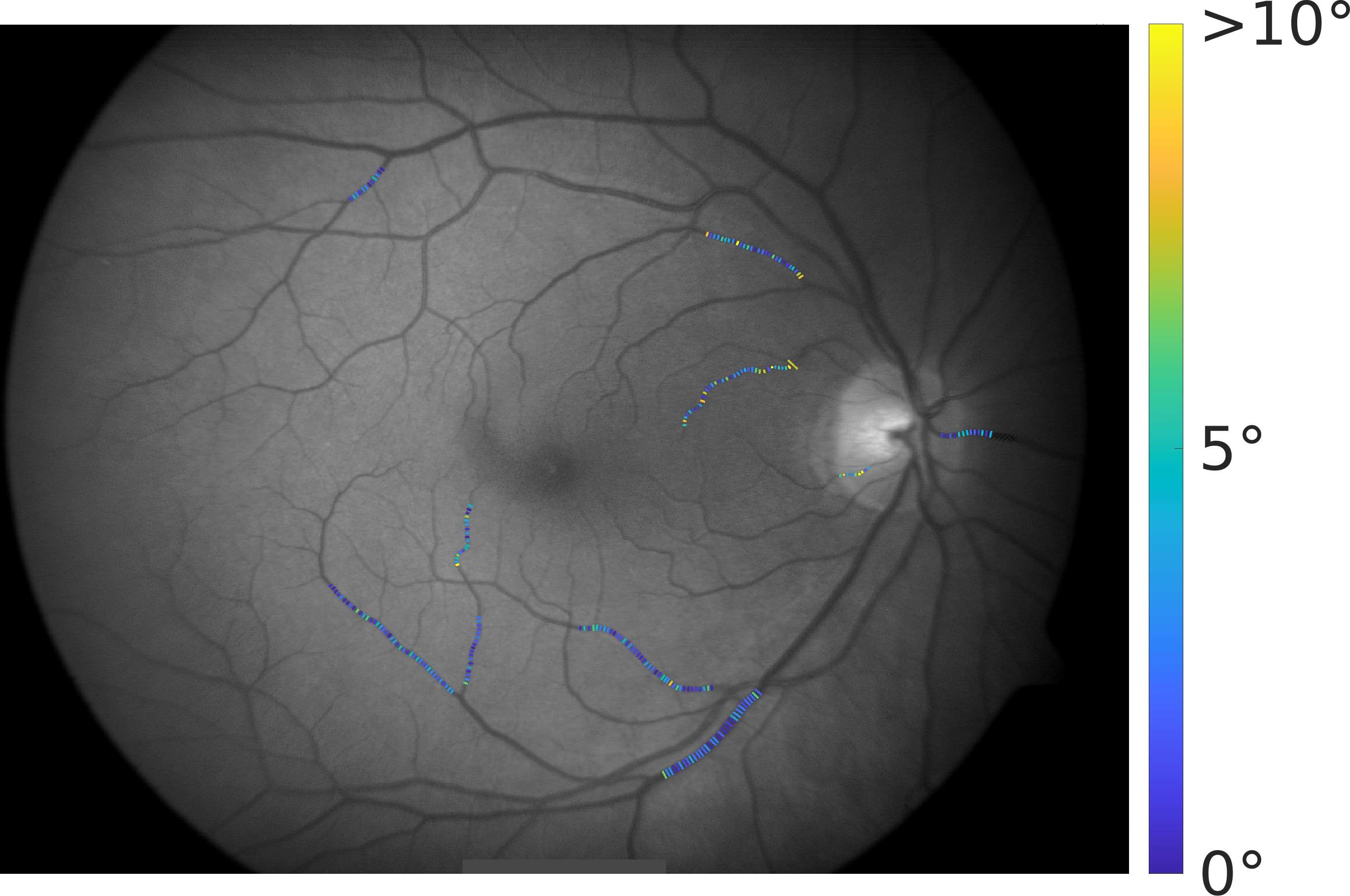}{2.8cm}{(0.315,0.29)}{(-0.32,0.35)}{2.5cm}}
	\caption{Width and orientation measurements of retinal blood vessels yielded by SymFD. The input image is part of the VDIS dataset from the REVIEW retinal vessel reference database.}	
	\label{fig:review_SFD}
\end{figure}

The diameters of blood vessels that are visible in the images of the database roughly range from $2$ to $20$ pixels. To obtain a precise detection and characterization of both thin and wide blood vessels, we use not a single pair but two pairs of systems of even- and odd-symmetric molecules that cover different but overlapping regions in the frequency domain. Both pairs of systems are based on the even-symmetric one-dimensional wavelet $\psie = \frac{\HT G_1}{\normL{\HT G_1}}$ and the odd-symmetric wavelet $\psio = \frac{G_1}{\normL{G_1}}$ and cover two octaves with $\SFDparammath{scalesPerOctave} = 4$,  $\alpha = 1$, and $\SFDparammath{nOrientations} = 16$. The first pair of systems aims \rev{to detect} of thin vessel and is defined by the parameters $\SFDparammath{maxFeatureWidth}_1 = 8$, $\SFDparammath{maxFeatureLength}_1 = 24$, and $\SFDparammath{minFeatureWidth}_1 = 2$. The second pair is capable of detecting vessels with a diameter of up to 24 pixels and is defined by the parameters $\SFDparammath{maxFeatureWidth}_2 = 24$, $\SFDparammath{maxFeatureLength}_2 = 72$, and $\SFDparammath{minFeatureWidth}_2 = 6$. Each image in the database is processed by computing the ridge measure $\RM(f,y)$, the width measure $\WMR(f,y)$ and the tangent direction measure $\OMR(f,y)$ for both pairs of systems of even- and odd-symmetric molecules with parameters $\jo = 0$, and $\beta = 2$. In each case, the computations are carried out solely on the green channel of the processed RGB image, in which the blood vessels typically have the most significant contrast. For every pixel location $y$, we keep the values yielded by the pair of systems associated with the larger ridge measure $\RM(f,y)$. We further restrict the analysis to ridges with negative contrast, that is, to locations where $\HMR(f,y) \leq 0$. As the width measure $\WMR(f,y)$ is not only based on the even-symmetric coefficient on the most significant scale but also on the coefficients with respect to the preceding and succeeding scaling parameters, measurements where the first or last scales are most significant are not considered. The obtained feature maps in the case of image 7 from the VDIS dataset are plotted in \rev{Figures~\ref{fig:review_SFD_fm}~to~\ref{fig:review_SFD_ori}}.

To validate the obtained measurements, the width and tangent direction maps yielded by SymFD are compared to the widths and orientations of the manually marked profiles. For each vessel edge profile in the ground truth, we consider the difference with respect to the width measure $\WMR(f,y)$ and the orientation measure $\OMR(f,y)$ at the nearest location $y$ in a two-pixel radius for which the ridge measure $\RM(f,y)$ is greater than zero. If no such pixel exists, the corresponding vessel edge profile counts as undetected. Visualizations of the tangent direction error and the width error for measurements obtained from SymFD are visualized in the case of image 7 from the VDIS dataset in Figures~\ref{fig:review_SFD_error_width}~\rev{and}~\ref{fig:review_SFD_error_ori}. For the tangent orientation estimates $\OMR(f,y)$, we compute the mean absolute error $\MAEOMR$ on the torus with respect to all manually marked profiles for each of the four subsets of images. The respective values along with the \emph{success rate} $\SR$, which denotes the percentage of successfully matched vessel profiles, are reported in Table~\ref{tab:review_results_orientation}.
\begin{table}[t!]
\centering
\caption{Evaluation of Orientation Measurements on the REVIEW Database}
\label{tab:review_results_orientation}
\begin{scriptsize}
\begin{threeparttable}
	{\setlength{\tabcolsep}{4pt}	\renewcommand{\arraystretch}{1.3}\begin{tabular}{lrrcrrcrrcrr}
			\toprule[0.5mm]
			&\multicolumn{2}{c}{VDIS} & \phantom{abc}&\multicolumn{2}{c}{KPIS} & \phantom{abc}&\multicolumn{2}{c}{HRIS} & \phantom{abc}&\multicolumn{2}{c}{CLRIS}\\
			&$\MAEOMR$ & $\SR$ & &$\MAEOMR$ & $\SR$ & &$\MAEOMR$ & $\SR$ & &$\MAEOMR$ & $\SR$\\
			\midrule
			\textit{Human Observers}\\
			O1 & 2.69\,$^\circ$ & 100\perc& & 2.40\,$^\circ$ & 100\perc& & 2.82\,$^\circ$ & 100\perc& & 1.62\,$^\circ$ & 100\perc\\
			O2 & 2.97\,$^\circ$ & 100\perc& & 2.52\,$^\circ$ & 100\perc& & 2.41\,$^\circ$ & 100\perc& & 1.61\,$^\circ$ & 100\perc\\
			O3 & 3.47\,$^\circ$ & 100\perc& & 3.18\,$^\circ$ & 100\perc& & 2.70\,$^\circ$ & 100\perc& & 1.51\,$^\circ$ & 100\perc\\
			\textit{Algorithm}\\
			SymFD & 3.30\,$^\circ$ & 99\perc& & 2.74\,$^\circ$ & 100\perc& & 2.46\,$^\circ$ & 100\perc& & 2.93\,$^\circ$ & 95\perc\\\midrule[0.5mm]
		\end{tabular}}
		\begin{tablenotes}\footnotesize
			\item $\MAEOMR$: mean \rev{absolute} error of the tangent direction measure, $\SR$: success rate. Note that $\MAEOMR$ is computed only with respect to points in the ground truth that are sufficiently close to points detected by an algorithm. The ratio of considered points is indicated by $\SR$.
		\end{tablenotes}
	\end{threeparttable}
\end{scriptsize}
\end{table}

For the validation of \rev{the} width measurements $\WMR(f,y)$, we follow the procedure proposed in \cite{aldiri2009active,bankhead2012fastretinal,xu2011vesselboundary} and consider the standard deviation $\SD_{\WMR}$ of differences between the measure $\WMR(f,y)$ and the ground truth vessel diameters instead of the mean absolute error. This way, the comparison is independent of possibly different implicit definitions of vessel widths. We further compare SymFD to a number of classical and state-of-the-art methods for the extraction of vessel diameters from digital retinal images. In \cite{zhou1994detection}, vessel widths are obtained by fitting a \emph{one-dimensional Gaussian} (1DG) model to the vessel intensity profiles. A similar approach using a \emph{two-dimensional Gaussian} (2DG) model  was proposed in \cite{lowell2004measurement}. The \emph{half height full width} (HHFW) model defines the width of a vessel as the distance between the points at which the vessel intensity profile reaches half of the maximum intensity left and right of an assumed center point \cite{brinchmann1986theoretical}. Gregson et al.\ proposed to obtain vessel widths by fitting a rectangle to the intensity profile such that the area under the rectangle is equal to the area under the intensity profile \cite{gregson1995automated}. A graph-based edge segmentation technique for measuring vessel widths was proposed in \cite{xu2011vesselboundary}, while the \emph{extraction of segment profiles} (ESP) algorithm proposed in \cite{aldiri2009active} is based on an active contour model. The \emph{automated retinal image analyzer} (ARIA) employs a wavelet-based approach for vessel segmentation which is followed by a refinement of edge locations \cite{bankhead2012fastretinal}. Finally, \emph{edge tracking based on orientation scores} (ETOS) is an algorithm that yields state-of-the-art vessel width measurements by simultaneously tracking both edges of a vessel in the coupled space of positions and orientations \cite{bekkers2014multiorientation}. The accuracy of vessel width measurements on images from the REVIEW database obtained by SymFD and all aforementioned methods in terms of the standard deviation $\SD_{\WMR}$ and success rate $\SR$ are compiled in Table~\ref{tab:review_results_width}.  
\begin{table}[t!]
\centering
\caption{Evaluation of Width Measurements on the REVIEW Database}
\label{tab:review_results_width}
\begin{scriptsize}
	\begin{threeparttable}
		{\setlength{\tabcolsep}{4pt}\renewcommand{\arraystretch}{1.3}\begin{tabular}{lrrcrrcrrcrr}
				\toprule[0.5mm]
				&\multicolumn{2}{c}{VDIS} & \phantom{abc}&\multicolumn{2}{c}{KPIS} & \phantom{abc}&\multicolumn{2}{c}{HRIS} & \phantom{abc}&\multicolumn{2}{c}{CLRIS}\\
				&$\SD_{\WMR}$ & $\SR$ & &$\SD_{\WMR}$ & $\SR$ & &$\SD_{\WMR}$ & $\SR$ & &$\SD_{\WMR}$ & $\SR$\\
				\midrule
				\textit{Human Observers}\\
				O1 & 0.54\px & 100\perc& & 0.23\px & 100\perc& & 0.29\px & 100\perc& & 0.57\px & 100\perc\\
				O2 & 0.62\px & 100\perc& & 0.21\px & 100\perc& & 0.26\px & 100\perc& & 0.70\px & 100\perc\\
				O3 & 0.67\px & 100\perc& & 0.23\px & 100\perc& & 0.28\px & 100\perc& & 0.57\px & 100\perc\\
				\textit{Algorithms}\\
				SymFD & 1.20\px & 99\perc& & 0.41\px & 100\perc& & 0.60\px & 100\perc& & 1.44\px & 95\perc\\
				ARIA \cite{bankhead2012fastretinal} & 0.95\px & 99\perc& & 0.29\px & 100\perc& & 0.32\px & 100\perc& & 0.95\px & 100\perc\\
				ETOS \cite{bekkers2014multiorientation} & 0.80\px & 100\perc& & 0.36\px & 100\perc& & 0.45\px & 100\perc& & 0.53\px & 100\perc\\
				ESP \cite{aldiri2009active} & 0.77\px & 100\perc& & 0.33\px & 100\perc& & 0.42\px & 100\perc& & 1.47\px & 93\perc\\
				Graph \cite{xu2011vesselboundary} & 1.43\px & 96\perc& & 0.67\px & 99\perc& & 0.57\px & 100\perc& & 1.78\px & 94\perc\\
				Gregson \cite{gregson1995automated} & 1.49\px & 100\perc& & 0.60\px & 100\perc& & 2.84\px & 100\perc& & 2.84\px & 100\perc\\
				HHFW \cite{brinchmann1986theoretical} & 0.88\px & 78\perc& & 0.39\px & 96\perc& & 0.93\px & 88\perc& & \multicolumn{1}{c}{n/a} & 0\perc\\
				1DG \cite{zhou1994detection} & 2.11\px & 100\perc& & 0.40\px & 100\perc& & 4.14\px & 100\perc& & 4.14\px & 99\perc\\
				2DG \cite{lowell2004measurement} & 1.33\px & 77\perc& & 0.34\px & 100\perc& & 6.02\px & 99\perc& & 6.02\px & 27\perc\\\midrule[0.5mm]
			\end{tabular}}
				\begin{tablenotes}\footnotesize
					\item $\SD_{\WMR}$: standard deviation from the ground truth of the width measure, $\SR$: success rate. Note that $\SD_{\WMR}$ is computed only with respect to points in the ground truth that are sufficiently close to points detected by an algorithm. The ratio of considered points is indicated by $\SR$.
				\end{tablenotes}
		\end{threeparttable}
	\end{scriptsize}
\end{table}
\subsection{Automated Cell Colony Counting}
\label{sec:cellcol}
Counting the number of cell colonies growing on an agar disk is a common but time-consuming task in many areas of biomedical research. To evaluate the applicability of the blob measure $\BM(f,y)$ for automatically obtaining the number of colonies from digital images, we consider a set of pictures which was originally developed for the evaluation of the OpenCFU software package \cite{geissmann2013opencfu}. The set consists of digital images of 19 different plates containing between 10 and 1000 \emph{Staphylococcus aureus} colonies and can be downloaded from \url{http://opencfu.sourceforge.net/}. Each plate was photographed once with a high-definition camera and once with a low-cost webcam. To provide a reference for the evaluation of automated counting methods, the number of colonies on each plate was independently counted by seven trained humans. For a single plate, the ground truth was then defined as the median of the numbers reported by the human experts. Examples of the considered images are displayed in Figure~\ref{fig:cellcol_input} in the Supplemental Materials.

\begin{figure}[t!]
	\centering
	\subfloat[Input 1 (HD image with a resolution of $1538\times 1536$ pixels. Median human colony count: 359.)]{\label{fig:cellcol_SFD_input1}\spyon{3.2}{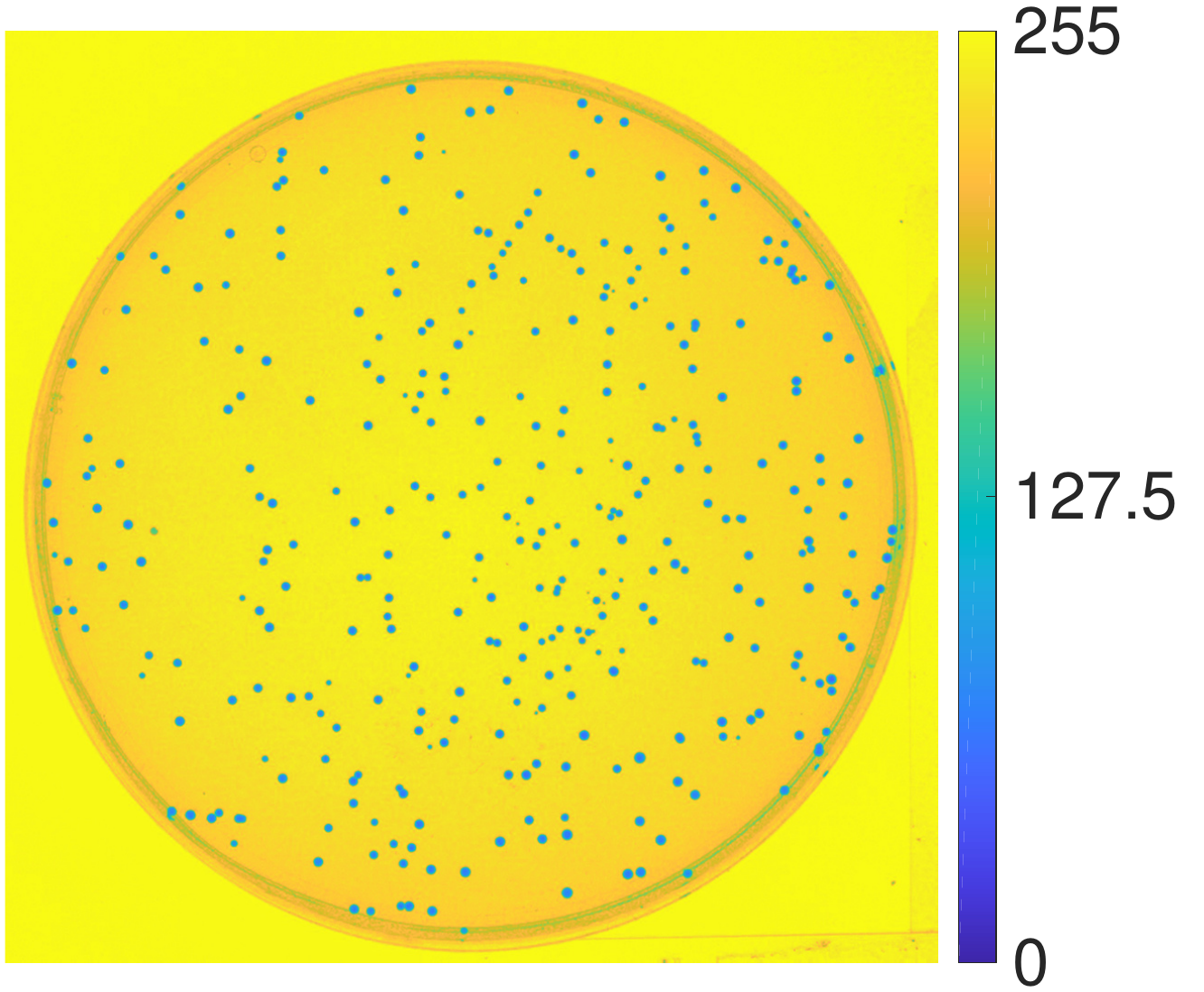}{3.5cm}{(0.24,0.67)}{(-0.33,0.3)}{2.4cm}}\hspace{.4cm}
	\subfloat[Input 2 (Webcam image with a resolution of $1000\times 1000$ pixels. Median human colony count: 525.)]{\label{fig:cellcol_SFD_input2}\spyon{3.2}{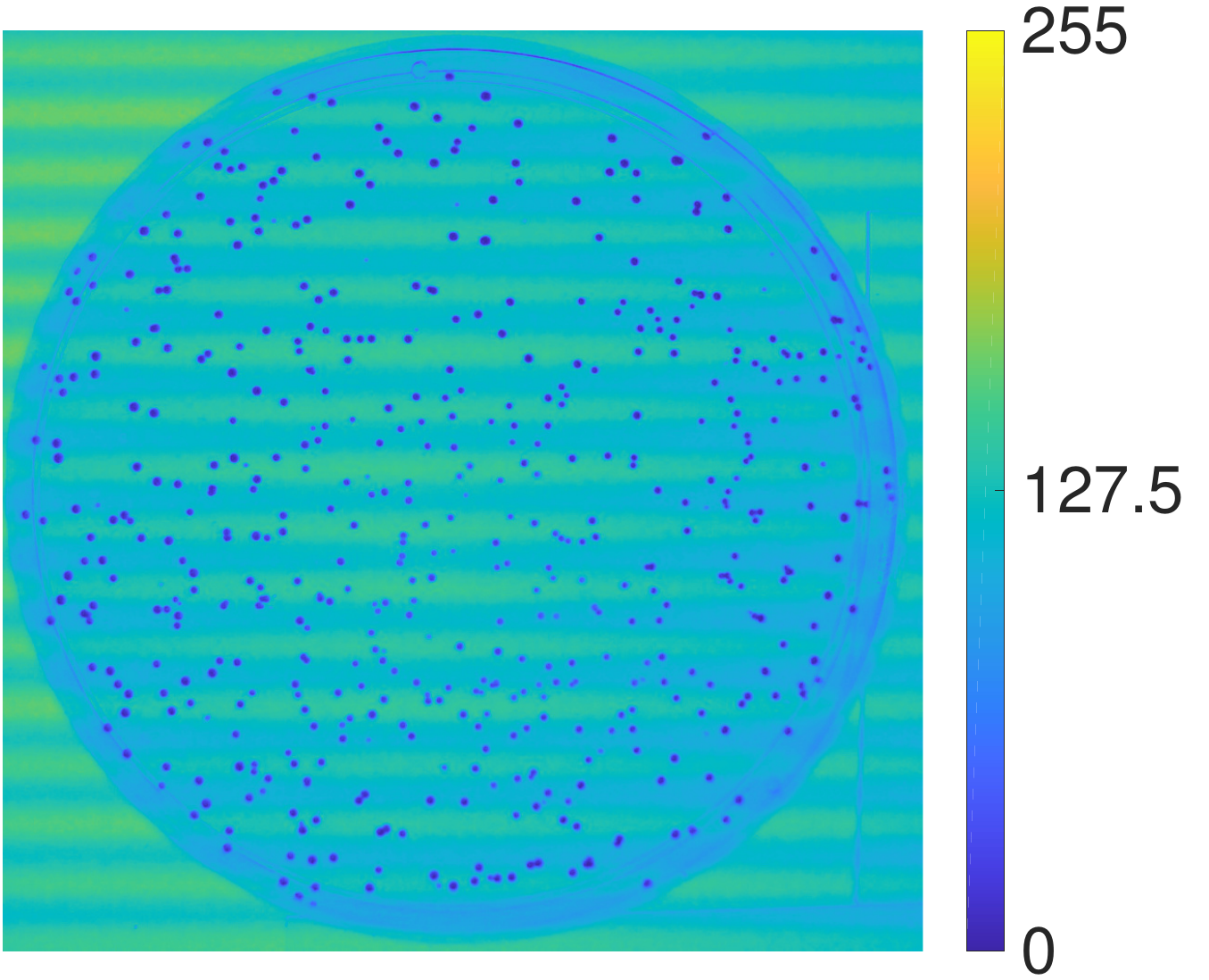}{3.5cm}{(0.22,0.825)}{(-0.33,0.3)}{2.4cm}}\\
	\subfloat[Cell colonies detected in input 1 by applying the blob measure $\BM(f,y)$ implemented in SymFD. Number of detected colonies: 357.]{\label{fig:cellcol_SFD_fm_thinned1}\spyon{3.2}{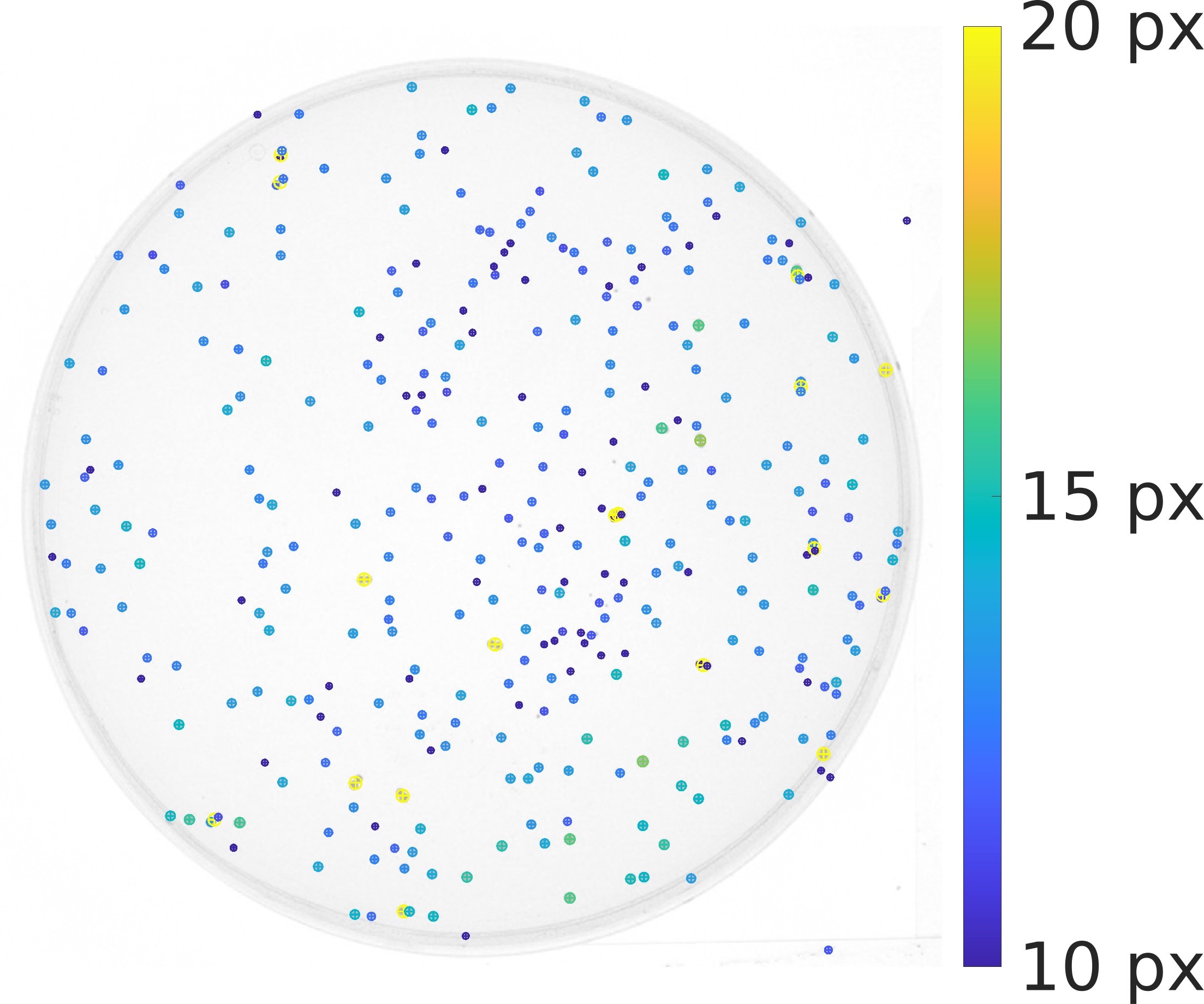}{3.5cm}{(0.24,0.67)}{(-0.33,0.3)}{2.4cm}}\hspace{.4cm}
	\subfloat[Cell colonies detected in input 2 by applying the blob measure $\BM(f,y)$ implemented in SymFD. Number of detected colonies: 512.]{\label{fig:cellcol_SFD_fm_thinned2}\spyon{3.2}{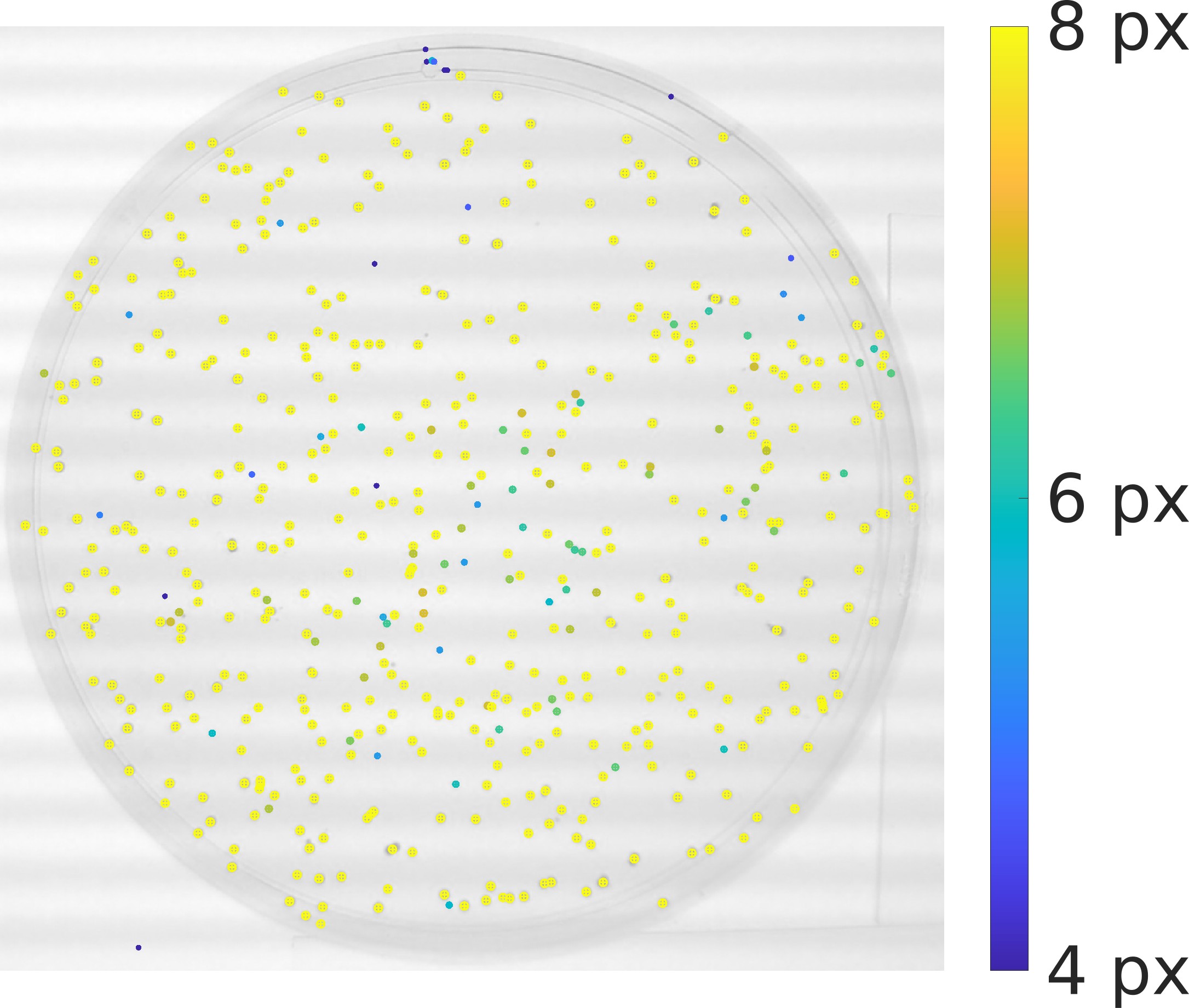}{3.5cm}{(0.222,0.825)}{(-0.33,0.3)}{2.4cm}}
	\caption{SymFD-based detection of cell colonies in digital photographs of Petri dishes.}
	\label{fig:cellcol_SFD}
\end{figure}

To automatically count the number of cell colonies in the high-definition images, all 19 images were processed with SymFD by computing the blob measure $\BM(f,y)$ on the blue color channels using the parameters $\psie = \frac{\HT G_1}{\normL{\HT G_1}}$, $\jo = -1$, $\SFDparammath{maxFeatureWidth}_{\text{hd}} = 20$, $\SFDparammath{maxFeatureLength}_{\text{hd}} = 20$, $\SFDparammath{minFeatureWidth}_{\text{hd}} = 10$, $\alpha = 1$, $\SFDparammath{scalesPerOctave} = 3$, $\SFDparammath{nOrientations} = 16$, and $\beta_{\text{hd}} = 15$. The obtained feature maps were then transformed into binary images by applying the threshold $0.03$. Centers of cell colonies were eventually obtained by computing the centroids of all connected components in the binary image via \matlab{}'s \SFDparam{regionprops} function. For processing the webcam images, the parameters specifying the effective support of the applied symmetric molecules had to be adapted to accommodate for the lower resolution and were set to $\SFDparammath{maxFeatureWidth}_{\text{wc}} = 8$, $\SFDparammath{maxFeatureLength}_{\text{wc}} = 8$, and $\SFDparammath{minFeatureWidth}_{\text{wc}} = 4$. Furthermore, mostly due to horizontal stripe artifacts %(cf. Figure~\ref{fig:cellcol_input_webcam}), 
the contrasts of the cell colonies are slightly lower in the images captured by a webcam than in their high-definition counterparts. Thus, for processing the webcam images, the parameter specifying the minimal contrast was chosen as $\beta_{\text{wc}} = 6$. Examples of cell colony detection results obtained by SymFD for one high-definition image and one webcam image are visualized in Figure~\ref{fig:cellcol_SFD}. 
%
%\begin{figure}[t!]
%	\centering
%	\subfloat[The shown image was acquired with a high-definition camera with a resolution of $1538\times 1536$ pixels. Median human count of cell colonies: 1090.]{\label{fig:cellcol_input_HD}\spyon{3.2}{figs/F.jpg}{3.8cm}{(0.4,0.6)}{(-0.4,0.3)}{2.75cm}}\hspace{.4cm}
%	\subfloat[The shown image was obtained from a low-cost webcam with horizontal stripe artifacts and acquired at a resolution of $1000\times 1000$ pixels. Median human count of cell colonies: 69.]{\label{fig:cellcol_input_webcam}\spyon{3.2}{figs/wc_A.png}{3.8cm}{(0.48,0.41)}{(-0.4,0.3)}{2.75cm}}
%	\caption{Two pictures showing grown cell colonies in a Petri dish. The dataset containing the displayed images was originally developed for the evaluation of the OpenCFU software package \cite{geissmann2013opencfu}.}
%	\label{fig:cellcol_input}
%\end{figure}

To validate the accuracy of the cell colony counts obtained by SymFD, we follow the procedure proposed in \cite{geissmann2013opencfu} and consider the median relative deviation from the counts provided by the human experts across all 19 images. We further compare the accuracy of SymFD with three other methods for the automated detection and counting of cell colonies. The \emph{NIST's Integrated Colony Enumerator} (NICE) \cite{clarke2010lowcost} is a freely available software package based on a combination of extended-minima transforms and thresholding operations. We also consider an ImageJ plugin developed by Cai et al.\ \cite{cai2011optimized} that is based on adaptive thresholding, the application of a watershed algorithm, and particle filtering. The results obtained by SymFD are compared to OpenCFU which is a highly versatile and efficiently implemented open source library for the detection of cell colonies \cite{geissmann2013opencfu}. The median relative deviations from human counts for SymFD and the aforementioned methods with respect to high-definition and webcam images of all 19 plates are compiled in Table~\ref{tab:cellcol_results}. The cell colony counts for OpenCFU, NICE, and the ImageJ plugin proposed by Cai et al.\ were kindly provided by the author of \cite{geissmann2013opencfu}. In particular NICE and the ImageJ macro are sensitive to artifacts in the background and at the border of the Petri dish. To facilitate a fair comparison, Table~\ref{tab:cellcol_results} also reports the median relative errors when using an additional foreground mask. To illustrate the robustness of the considered methods with respect to images of agar plates with both small and large numbers of cell colonies, semilog graphs plotting for each algorithm the relative deviation for a single plate against the number of cell colonies in the corresponding plate are shown Figure~\ref{fig:cellcol_graphs}.

\section{Discussion of Numerical Results}
\label{sec:discussion}
The results of the numerical evaluation on synthetic images summarized in Tables~\ref{tab:synthimages_edges_comparison},~\ref{tab:synthimages_ridges_comparison},~and~\ref{tab:synthimages_blobs_comparison} indicate that the detection accuracy of the proposed measures $\EM(f,y)$, $\RM(f,y)$, and $\BM(f,y)$ at least matches the performance of some of the most popular state-of-the-art algorithms for edge, ridge, and blob detection. In particular, all three measures show a high robustness in the presence of noise. Maybe the most significant instance of this may be found in the case of blob detection, where the circular Hough transform clearly fails to reliably detect filled circles in severely distorted images, while SymFD still yields a perfect detection result in the case of \synthfive{} and correctly identifies three quarters of the blobs in \synthsix{} (see Table~\ref{tab:synthimages_blobs_comparison} and Figure~\ref{fig:synthimages_blobs_comparison}). However, the experiments also show that regarding the mere detection of features, SymFD does not provide a significant advantage over some of the already established methods for the considered test inputs. In the case of edge detection, the well-known Canny edge detector \cite{Can1986} matches or even slightly outperforms the detection rate of SymFD for all of the six test images (see Table~\ref{tab:synthimages_edges_comparison}) while the same is true in the case of ridge detection for the Steger algorithm \cite{steger1998unbiased} (see Table~\ref{tab:synthimages_ridges_comparison}).

One of the most significant benefits of the proposed framework is that it not only allows for the detection of different types of features but also yields a comprehensive characterization of their geometry in terms of local tangent directions and diameters. The respective mean absolute errors reported in Tables~\ref{tab:synthimages_edges_comparison}~and~\ref{tab:synthimages_ridges_comparison} indicate a comparatively high accuracy of the tangent orientation measures $\OME(f,y)$ and $\OMR(f,y)$, and the width measure $\WMR(f,y)$. However, the experiments also show that for all considered test images, the width and orientation measures are sensitive to noise, where the most significant decrease in accuracy may be found in the case of $\OME(f,y)$ and \synthtwo{} (see Table~\ref{tab:synthimages_edges_comparison}). The visualization of the local error of the tangent direction estimates shown in Figure~\ref{fig:synthimages_edges_SFD_orierror} in the Supplemental Materials furthermore reveals that the accuracy of the measure $\OME(f,y)$ deteriorates in the proximity of points on which the edge curve is not smooth (e.g. corner points). In the case of blob detection, the errors of the width measure $\WMB(f,y)$ summarized in Table~\ref{tab:synthimages_blobs_comparison} and visualized in Figure~\ref{fig:synthimages_blobs_SFD_error_width} in the Supplemental Materials are relatively large, even in comparison with the results yielded by the circular Hough transform. This can in part be explained by the fact that the two-dimensional generators from the set $\Psieet$ are based on tensor products of one-dimensional even-symmetric wavelets and thus designed to detect square-shaped rather than circular blobs, which is also reflected in the definition of the function $\widetilde{K}_{\genet}(r)$ in Section~\ref{sec:blobmeasure} (see \eqref{eq:kpsiet}). We suspect that the accuracy of the measure $\WMB(f,y)$ with respect to filled circles would significantly be improved by using circular two-dimensional generators and adjusting the definition of $\widetilde{K}_{\genet}(r)$ accordingly. A first choice for such a generator could be the two-dimensional Mexican hat wavelet. We further suspect that using a generator that actually matches the symmetry properties and shape of the objects that are to be detected would also have a positive effect on the overall accuracy of the measure $\BM(f,y)$.
\begin{table}[t!]
\centering
\caption{Accuracy of Cell Colony Counts}
\label{tab:cellcol_results}
	\begin{small}
		\begin{threeparttable}
			{\setlength{\tabcolsep}{4pt}\renewcommand{\arraystretch}{1.3}\begin{tabular}{lrrcrr}
				\toprule[0.5mm]
				&\multicolumn{2}{c}{HD} & \phantom{abc}&\multicolumn{2}{c}{Webcam}\\
				&With Mask & Without Mask & &With Mask & Without Mask\\
				\midrule
				SymFD & 1.95\,\% & 3.57\,\%& & 2.23\,\% & 5.95\,\%\\
				OpenCFU \cite{geissmann2013opencfu} & 1.93\,\% & 3.85\,\%& & 2.79\,\% & \multicolumn{1}{c}{n/a}\\
				ImageJ \cite{cai2011optimized} & 6.64\,\% & 80.93\,\%& & 11.11\,\% & \multicolumn{1}{c}{n/a}\\
				NICE \cite{clarke2010lowcost} & 9.91\,\% & 20.56\,\%& & 13.04\,\% & \multicolumn{1}{c}{n/a}\\\midrule[0.5mm]
			\end{tabular}}
			\begin{tablenotes}
				\item The accuracy is reported in terms of the median relative deviation from counts provided by human experts.
			\end{tablenotes}
		\end{threeparttable}
	\end{small}
\end{table}
								
The numerical experiments conducted in Section~\ref{sec:retinal} show that the measures $\RM(f,y)$, $\OMR(f,y)$, and $\WMR(f,y)$ can be used to reliably describe the geometry of blood vessels in different types of retinal images. In particular, the tangent direction estimates yielded by $\OMR(f,y)$ are highly accurate in the sense that their mean deviation from the ground truth is only slightly above the variation within the group of human experts (see Table~\ref{tab:review_results_orientation}). While the width measure $\WMR(f,y)$ yields more accurate estimates of local vessel diameters than many of the classical approaches it is also clearly outperformed by some of the more recent algorithms that were specifically developed for processing retinal images, such as ARIA \cite{bankhead2012fastretinal} and ETOS \cite{bekkers2014multiorientation} (see Table~\ref{tab:review_results_width}).
\begin{figure}[t!]
	\centering
	\subfloat[High-definition images with additional foreground mask.]{\label{fig:cellcol_graphs_hd_mask}\includegraphics[width=0.45\textwidth]{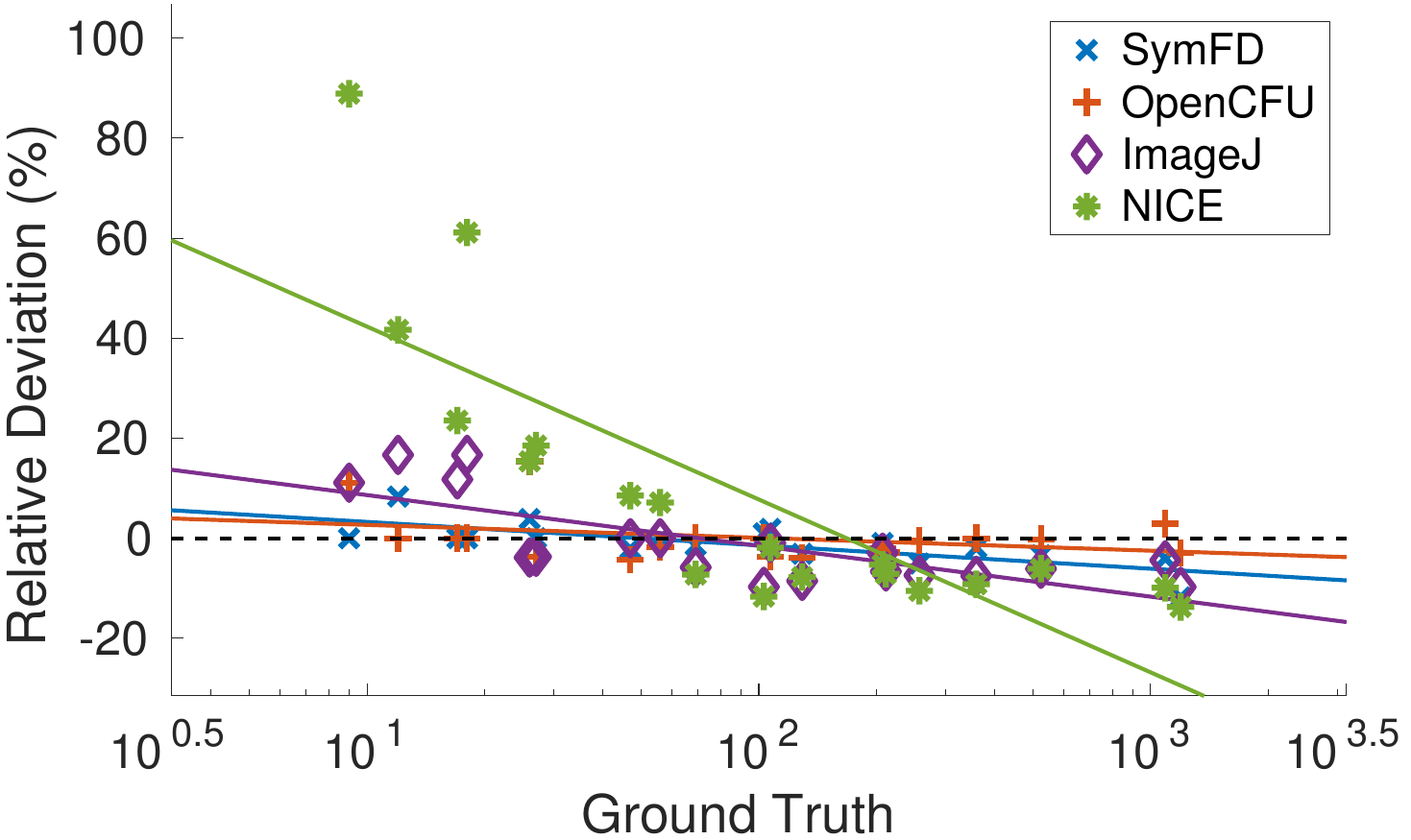}}\myhspace\subfloat[High-definition images without mask.]{\label{fig:cellcol_graphs_hd}\includegraphics[width=0.45\textwidth]{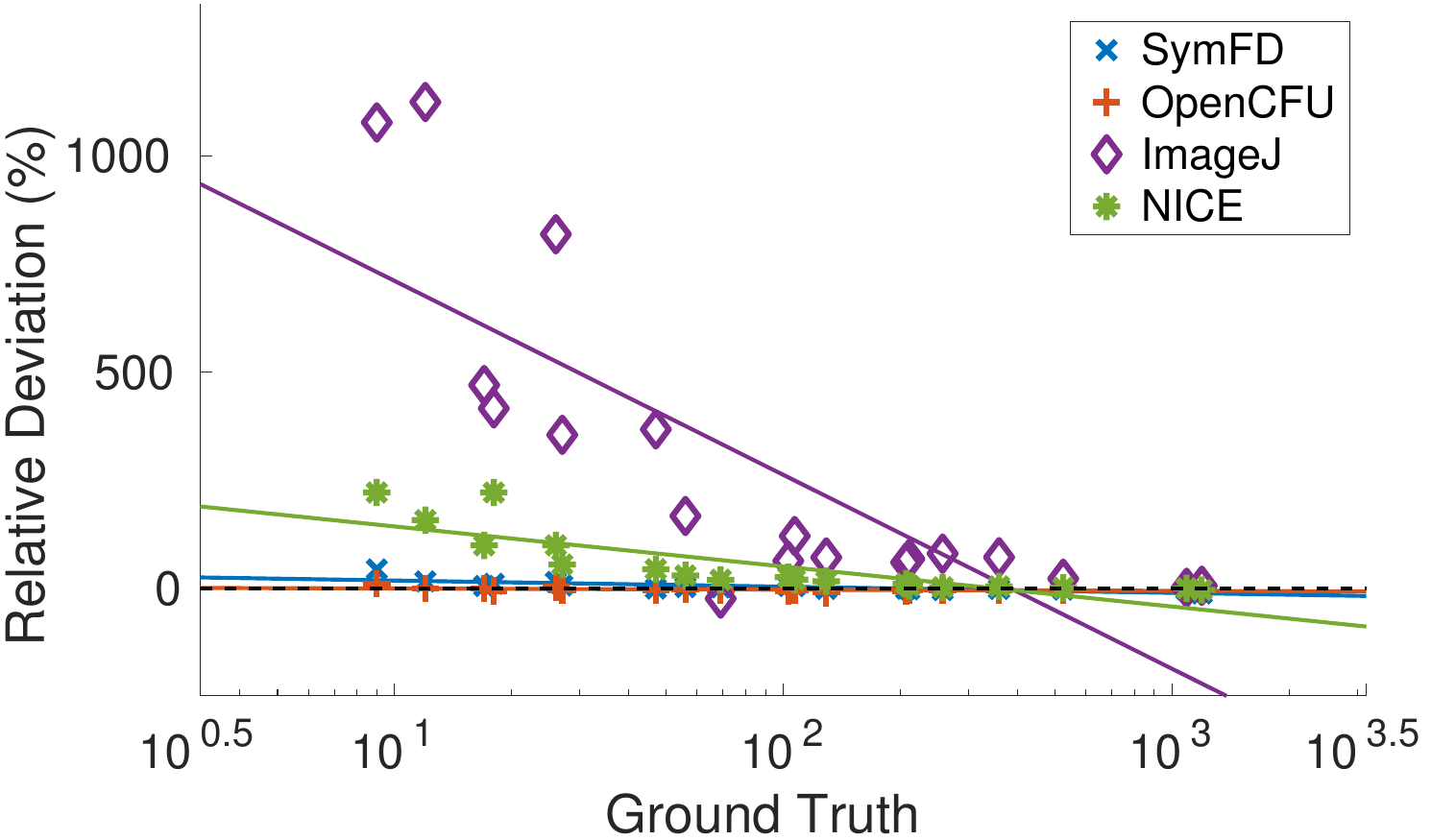}}\\
	\subfloat[Webcam images with additional foreground mask.]{\label{fig:cellcol_graphs_webcam_mask}\includegraphics[width=0.45\textwidth]{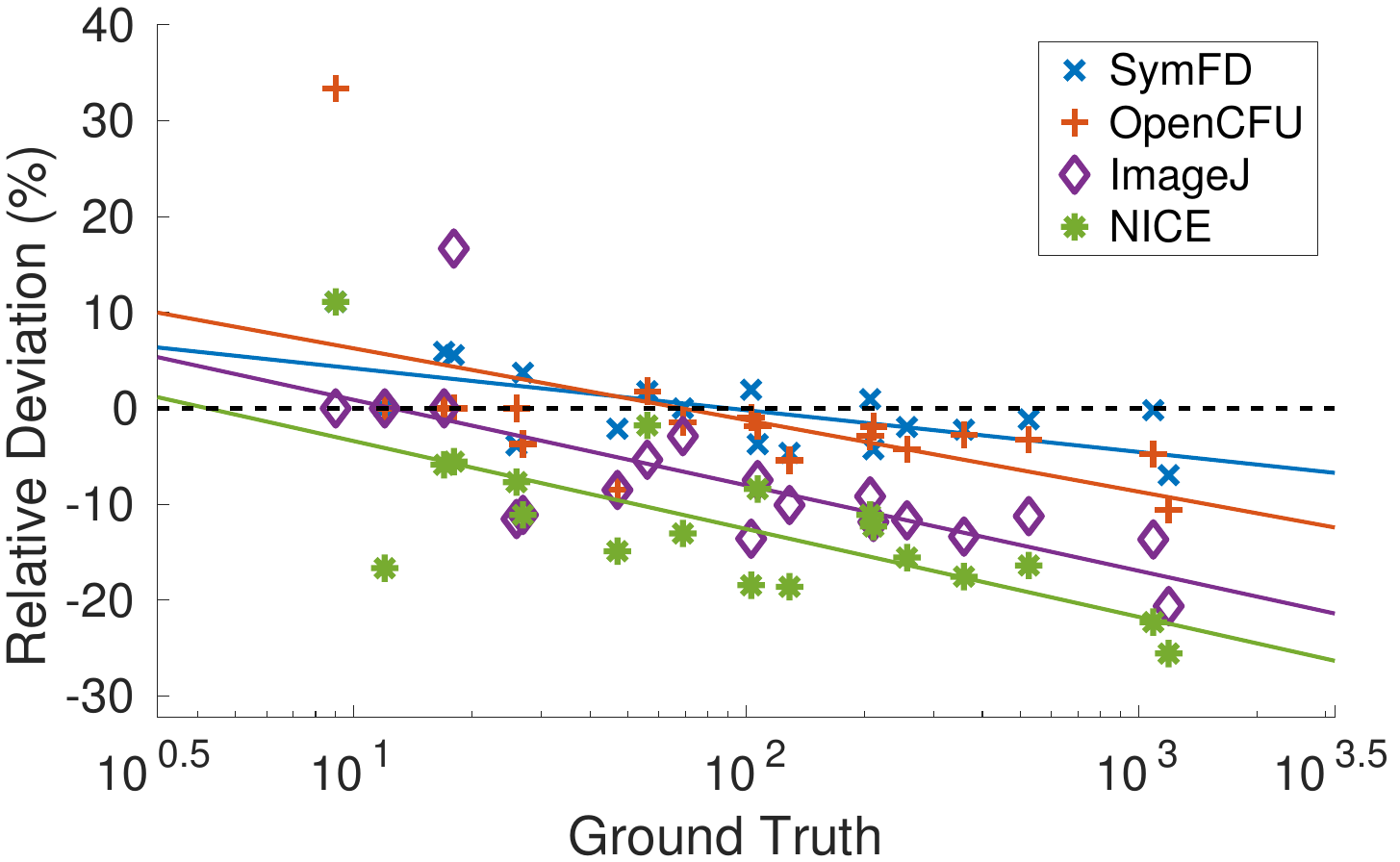}}\myhspace\subfloat[Webcam images without mask.]{\label{fig:cellcol_webcam}\includegraphics[width=0.45\textwidth]{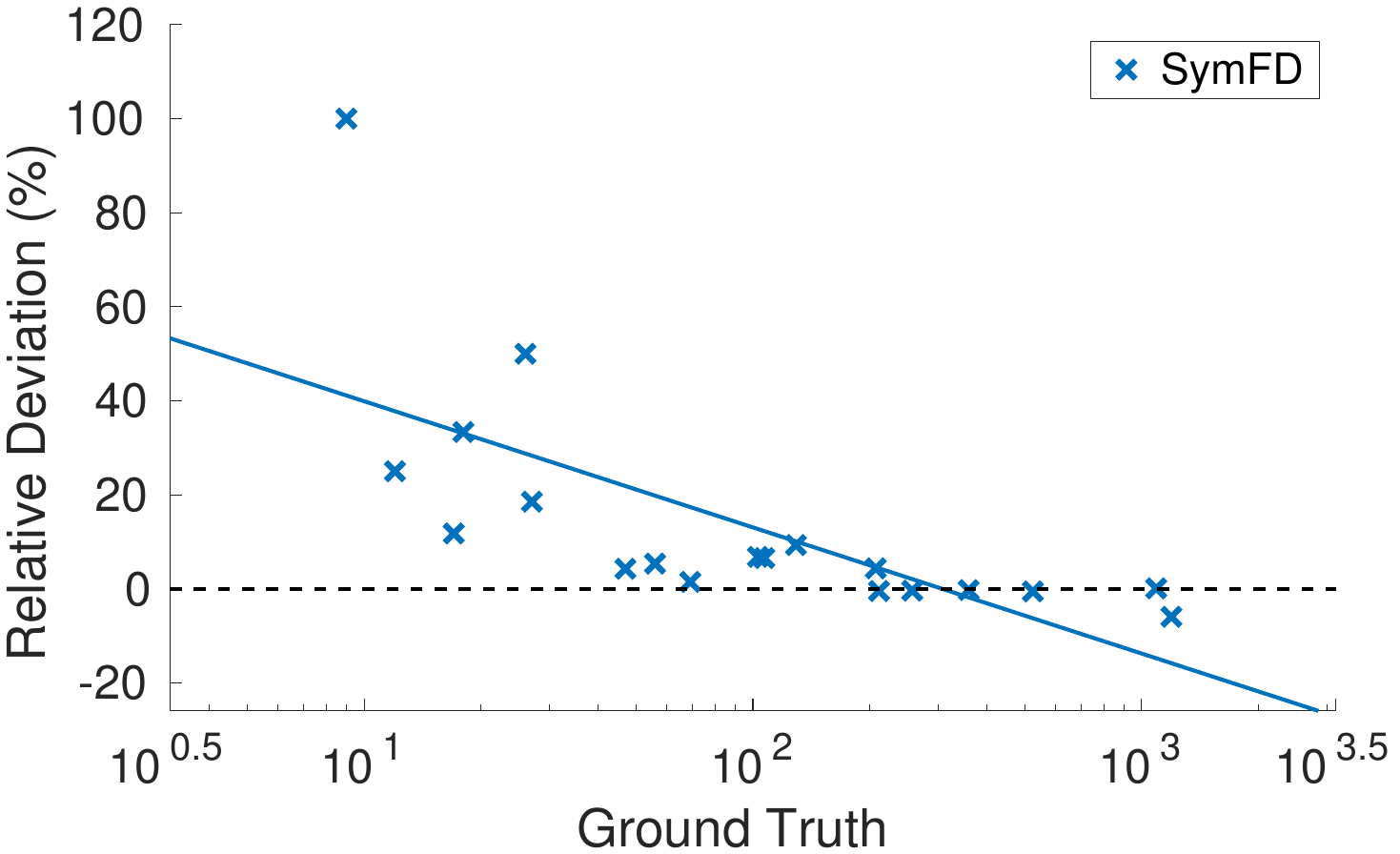}}
	\caption{Semilog graphs and trendlines for four automated cell counting methods plotting the relative deviation for a single plate against the number of cell colonies in the corresponding plate. }
	\label{fig:cellcol_graphs}
\end{figure}

The applicability of the blob measure $\BM(f,y)$ for automatically counting the number of cell colonies on an agar plate was investigated in Section~\ref{sec:cellcol}. It has been suggested that in practice, an average relative deviation of less than $20\perc$ will often be negligible in comparison to noise introduced by other experimental factors \cite{geissmann2013opencfu}. When using an additional foreground mask, SymFD achieves a median relative deviation from the ground truth of about $2\perc$ on a set of 19 high-definition images but also on a set of 19 low-quality webcam images (cf. Table~\ref{tab:cellcol_results}). This accuracy is comparable to the accuracy of the widely used OpenCFU software package \cite{geissmann2013opencfu} and suggests a high reliability of the results obtained by SymFD when applied for cell colony detection. The values reported in Table~\ref{tab:cellcol_results} also show only a slight decrease in accuracy for SymFD when omitting the foreground mask, which indicates that the measure $\BM(f,y)$ is not sensitive to structures in the image background or at the border of a Petri dish. Similar to other methods, SymFD tends to slightly overestimate the number of cell colonies on sparse plates while underestimating the cell colony count for high-density plates (cf. Figure~\ref{fig:cellcol_graphs}). This is mostly due to the high significance of false positives when only a few colonies are present and the difficulty of correctly identifying single colonies in dense clusters which often occur on high-density plates.

The numerical experiments described in Sections~\ref{sec:syntheval} and \ref{sec:applications} suggest that the proposed measures indeed define a highly versatile and powerful framework that can be reliably applied in a wide range of different feature detection and characterization tasks. There is, however, a trade-off for this high degree of flexibility, namely the relatively large number of parameters that are required to be selected when configuring SymFD for a specific task. %(cf. Table~\ref{tab:synthalgorithms}). 
Furthermore, evaluating the proposed edge, ridge, and blob measures on a digital image often requires the computation of more than 100 convolutions with digital symmetric molecule filters. While an average execution time of slightly less than 20 seconds %(cf. Table~\ref{tab:synthalgorithms}) 
is still acceptable in most tasks, SymFD is currently far from being applicable in real-time applications. 
\section{Conclusion}
\label{sec:conclusion}
In the present work, we propose a novel framework for the detection and characterization of features such as edges, ridges, and blobs in two-dimensional images. Drawing inspiration from investigations of the phase congruency property \cite{MRBO1986,MoOw1987,Kov1999}, the developed measures exploit the special symmetry properties of directionally sensitive analyzing functions that are based on tensor products of $L^1$-normalized derivatives of the one-dimensional Gaussian and their Hilbert transforms. Differently scaled and oriented two-dimensional analyzing functions are then constructed within the recently introduced $\alpha$-molecule framework \cite{grohs2016alpha}, which provides a maximum degree of flexibility. Combining both old and new ideas from applied harmonic analysis and computational vision, the developed measures are by construction contrast-invariant and furthermore capable of fully characterizing the geometry of features in terms of local tangent directions and local widths.

\rev{By considering pairs of even- and odd-symmetric analyzing functions that are sensitive to scaling and orientation, the measures $\EM(f,y)$, $\RM(f,y)$, and $\BM(f,y)$ are strongly inspired by functional properties that are known to be exhibited by neurons in the early human visual system \cite{daugman1985uncertaintyrelation}. Another property that has widely been observed in neural populations is that the response of single neurons is often normalized in the sense that it is divided by a weighted sum of the responses of neighboring neurons \cite{heeger1992normalization,carandini1997linearity,bonds1989roleofinhibition}. This principle is often termed \emph{divisive normalization} and closely related to the normalization procedure which ensures contrast invariance in the proposed measures (cf. \eqref{eq:emdt}, \eqref{eq:edgemeasure2d1}, \eqref{eq:ridgemeasure2d1}, and \eqref{eq:blobmeasure2d1}).

The implementations of the developed edge, ridge, and blob measures for digital images could also be modeled as artificial neural networks, where the applied even- and odd-symmetric digital filters define a convolutional layer, the absolute value is used as a non-linearity and max-pooling as well as divisive normalization are applied to obtain the final normalized values. It would be interesting to see whether a neural network architecture that is based on similar design principles would yield computations that are related to the formulas proposed in this work when trained for the detection of edges, ridges, or blobs. On one hand, this could lead to new ideas {to improve} the proposed measures {in order to} to better handle difficulties such as corner points, intersections of ridges, or densely grouped blobs. On the other hand, neural network architectures such as \cite{andrade2019extraction}, which have already been trained with great success for similar tasks, could potentially also be further improved by considering the computational principles behind the proposed measures. Ideally, such work would lead to hybrid approaches that combine the best of both worlds in the sense that they yield state-of-the-art performance results for specific applications that are nowadays often only achievable with learning-based methods while also being robust in applications where pure learning is not possible due to a lack of training data.}

We have demonstrated that the proposed measures can yield state-of-the-art detection performances when considering sets of both clean and distorted artificial images that are associated with reliable ground truths. Furthermore, we have given examples of how the developed feature detectors can be applied in tasks like the characterization and detection of blood vessels in retinal images, or the automated counting of cell colonies in a Petri dish. The obtained experimental results are promising and suggest a high applicability of the proposed measures in a wide range of diverse applications. However, they also reveal a number of open issues that leave ample room for future improvements. 

The present work only considers two-dimensional images. However, similar measures and construction principles of even- and odd-symmetric generators could also be applied in a three-dimensional setting to detect and characterize surfaces, planes, \rev{filaments}, or filled three-dimensional objects. The definition as well as the mathematical and experimental analysis of such measures could also be a fruitful topic of future research.
\section{Acknowledgments}
We thank J.-L.\ Starck for suggesting that we extend the framework to the case of blob detection and M.\ Sch\"afer for insightful discussions on the subject of $\alpha$-molecules. We are grateful to Q.\ Geissmann for kindly sharing a dataset of images of cell colonies and the corresponding ground truth counts. We furthermore thank D.\ Labate for sharing the \matlab{} implementation of his shearlet-based edge detector.

R. Reisenhofer acknowledges financial support from the Austrian Science Fund (FWF) under grant number M 2528.
\bibliography{symfdbib}{}
\bibliographystyle{siamplain}
\end{document}